\documentclass[11pt]{article}
\usepackage{bbding}
\usepackage[final]{acl}

\usepackage{times}
\usepackage{latexsym}

\usepackage[T1]{fontenc}
\usepackage{booktabs,multirow}
\usepackage[table]{xcolor}
\usepackage{arydshln}
\usepackage{enumitem}

\definecolor{BackboneBlue}{HTML}{EAF7F3}
\definecolor{BackboneViolet}{HTML}{EEF1F8}
\definecolor{HeaderGray}{HTML}{F4F5F7}

\usepackage[utf8]{inputenc}

\usepackage{microtype}

\usepackage{inconsolata}
\usepackage{amsmath}

\usepackage{graphicx}
\usepackage[most]{tcolorbox}
\title{SubtleMemory: A Benchmark for Fine-Grained Relational Memory Discrimination in Long-Horizon AI Agents}

\title{
  SubtleMemory: A Benchmark for Fine-Grained Relational Memory
  Discrimination in Long-Horizon AI Agents
}

\author{
  \textbf{Wenxuan Wang\textsuperscript{1,2,*}},
  \textbf{Haoyu Sun\textsuperscript{2,3,*}},
  \textbf{Fukuan Hou\textsuperscript{4}},
  \textbf{Mingyang Song\textsuperscript{5}}
\\
  \textbf{Weinan Zhang\textsuperscript{1}},
  \textbf{Yu Cheng\textsuperscript{2,7,\textdagger}},
  \textbf{Yang Yang\textsuperscript{6,\textdagger}}
\\[0.6em]
  \normalfont
  \textsuperscript{1}Harbin Institute of Technology,
  \textsuperscript{2}Shanghai AI Laboratory
\\
  \normalfont
  \textsuperscript{3}Tongji University,
  \textsuperscript{4}Xiamen University,
  \textsuperscript{5}Fudan University
\\
  \normalfont
\textsuperscript{6}Shanghai Jiao Tong University,
\textsuperscript{7}Nanyang Technological University
\\[0.6em]
\normalfont\small
\textsuperscript{*}Equal contribution.
\textsuperscript{\textdagger}Corresponding authors.
}

\begin{document}
\maketitle
\begin{abstract}
Persistent AI assistants, such as OpenClaw, accumulate large collections of related memories over long-term interactions.
As these memories grow, they may reinforce one another, diverge across contexts, or directly conflict, making correct assistance depend on memory relations rather than isolated recall. 
Existing long-term memory benchmarks do not systematically probe how agents
preserve and utilize such relations during downstream tasks.
To address this gap, we introduce \textbf{SubtleMemory}, a benchmark for fine-grained relational memory discrimination in long-running AI agents. SubtleMemory constructs relation-controlled latent semantic artifacts whose variants instantiate \textbf{complementary}, \textbf{nuanced}, or \textbf{contradictory} relations, and embeds them into realistic user-agent histories, requiring agents to recover distributed relational structures during later queries and instructions.
The benchmark contains 1,522 evaluation instances over 10 long histories, grounded in 1,090 relation-controlled memory-variant sets and spanning user-related and non-user-related queries.
Evaluating six standalone memory systems, two Claw-style agents with native memory modules, and three Claw-style agents with plugin memory modules, we find that current systems remain weak on fine-grained relational memory discrimination. 
We further introduce diagnostic protocols that reveal distinct capability profiles across memory preservation, retrieval, and downstream reasoning stages. \href{https://github.com/Yummytanmo/SubtleMemory}{Code} and
\href{https://huggingface.co/datasets/Yummytanmo/SubtleMemory}{data}
are publicly available.
\end{abstract}

\section{Introduction}



\begin{figure}[t!]
  \centering
  \includegraphics[width=0.5\textwidth]{figures/intro.pdf} 

  \caption{As personal memories accumulate, correct assistance depends on using relations among related memories rather than recalling isolated facts.}
  \label{fig:memory_types}
\end{figure}

Large language model agents are increasingly expected to function as persistent assistants rather than stateless conversational interfaces. In this setting, AI agents require memory mechanisms to retain and utilize information from long-term interactions to support continuity, personalization, and informed decision-making (Appendix~\ref{sec:appendix-memory-augmented-llm-agents}). As interactions accumulate over time, agents acquire large collections of highly related memories that may reinforce one another, subtly diverge under different contexts, or directly conflict. AI agents need to, therefore, not merely remember the past, but preserve and distinguish subtle relations among similar memories to behave appropriately.

This challenge is not unique to AI agents. Human memory research has long observed that accumulated experiences can interfere with one another, particularly when memories are highly similar or context-dependent~\citep{Underwood_1957,Johnson_1993,Smith_2001,Yassa_2011,Schlichting_2015,Hupbach_2007}. People may confuse where, when, or under what conditions a memory was formed, merge related experiences that should remain separate, or struggle to reconcile conflicting memories accumulated over time. Similarly, long-horizon AI agents may fail not only because of missing retrieval, but also similar or conflicting memories are incorrectly merged, overgeneralized, or misresolved despite being retrieved.


As discussed in Appendix~\ref{sec:appendix-long-term-memory-benchmarks},
earlier long-term memory benchmarks primarily evaluated the retrieval or
manipulation of individual memories and generally did not explicitly control
relations among multiple memories. More recent benchmarks, such as LongMemEval
and ClawArena, partially or implicitly capture such interactions through
multi-session reasoning, updating, or evolving histories, but do not
systematically control them as an evaluation dimension. To make this dimension
explicit and systematic, we introduce SubtleMemory, a comprehensive benchmark
for fine-grained relational memory discrimination in long-term AI assistant
usage, featuring explicit relation control and stage-specific diagnosis.

In SubtleMemory, we construct \textit{relation-controlled semantic variants} associated with shared resolution targets. These variants instantiate \textbf{complementary, nuanced, or contradictory} relations that determine whether related memories should be aggregated, distinguished, or reconciled during downstream reasoning. Rather than exposing variants as explicit memory entries, we latently embed them into natural multi-turn user--agent interaction sessions distributed across long-horizon conversation histories. We then construct evaluation queries grounded in the underlying resolution targets and semantic variants, requiring agents to recover and correctly reason over semantically related memories scattered throughout the interaction history.

Beyond the benchmark itself, we contribute (1) a unified evaluation framework that supports standalone memory systems, framework-native memory agents, and plugin-based memory agents under a consistent evaluation protocol; (2) a task-level diagnostic framework that decomposes failures arising from memory construction, retrieval, and final response generation, enabling more precise analysis of where long-term memory systems break down; (3) through extensive evaluation across six standalone memory systems, two Claw-style agents with native memory modules, and three Claw-style agents with plugin memory modules, we show that current systems struggle both to preserve fine-grained relational information during memory formation and to retrieve sufficient task-relevant evidence during inference. More strikingly, our relation-wise evaluation and stage-wise diagnostics show
that contradictory-memory instances expose more pronounced limitations across
memory preservation, retrieval, and answer generation than complementary and
nuanced instances, indicating that handling unresolved conflicts remains a key
bottleneck for current long-term memory agents operating over long-horizon
interaction histories.

\begin{figure*}[t!]
  \centering
  \includegraphics[width=1\textwidth]{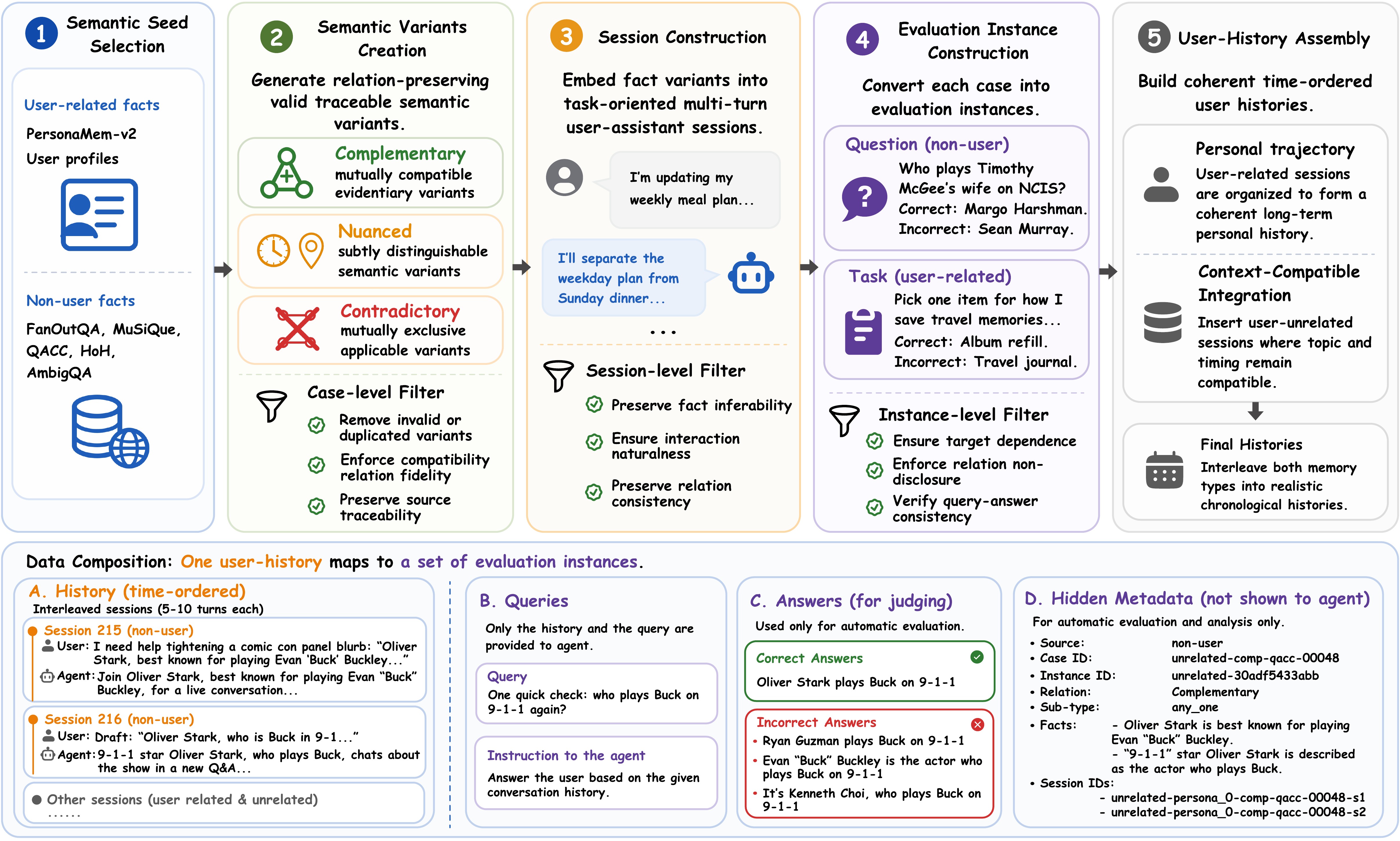} 

  \caption{SubtleMemory builds each split through a five-stage pipeline that turns semantic seeds into relation-preserving variants, task-oriented sessions, evaluation instances, and time-ordered histories with judging metadata.}
  \label{fig:data-construction}
\end{figure*}

\section{Methodology}

\subsection{Preliminary Concepts and Relation Taxonomy}
To characterize how multiple related memories should inform subsequent
decisions, SubtleMemory emphasizes \textit{relational memory reasoning}: the ability to aggregate compatible evidence, distinguish highly similar contexts, and reconcile conflicting memory records during downstream task completion.
We first define the semantic primitives that underlie the benchmark construction process.
\paragraph{Semantic Seeds and Variants.}
We begin with a collection of semantic facts as the seeds to construct the benchmark. A \textbf{semantic seed} may describe either \textbf{user-related facts}, such as preferences, habits, identities, or plans; or \textbf{non-user facts}, such as world knowledge, object attributes, or contextual facts. An example of a user semantic seed is as follows:
\begin{quote}
\textit{Bonita prefers Japanese minimalist interior design.}
\end{quote}
Given a semantic seed $\phi$, we construct a set of \textbf{semantic variants}
$
V(\phi) = \{v_1, v_2, \dots, v_n\},
$
where each variant contextualizes or transforms the original seed through controlled operations such as detail enrichment, partial detail masking, or semantic neighboring search with closely related but non-identical content.
\paragraph{Resolution Target.}
To evaluate memory reasoning behavior, SubtleMemory introduces the notion of a \textbf{resolution target}, defined as an information need whose successful resolution requires reasoning over accumulated memories.  
For example, when a user asks an agent to generate an apartment renovation plan, the resolution target may require \textit{identifying the user's applicable interior-design preference from prior interactions}.
\paragraph{Latent Semantic Artifacts.}
For each target $\tau$, we determine its corresponding fact seed $\phi$ and select  a subset of semantic variants
$
V_\tau \subseteq V(\phi),
$
called the \textbf{target-conditioned semantic variant set}. These variants collectively participate in resolving the target after being instantiated into memories through interaction histories. We define compatibility relation types $r(V_\tau)$ for the items in the target-conditioned semantic variant set as:
\begin{itemize}
    \item \textbf{Complementary}: variants provide mutually compatible evidence should be aggregated to resolve the target. During evaluation, the agent needs to either integrate information from multiple memory items or recognize any single evidence item is sufficient to answer the query. We further divide this relation into two subtypes: \textbf{Multi-evidence} and \textbf{Any-one}.
    \item \textbf{Nuanced}: variants are semantically similar but require fine-grained discrimination under the target. During evaluation, the agent must distinguish subtle differences between related memory items and identify the correct one for the query. Depending on whether the distinction arises from temporal or contextual cues, we further divide this relation into two subtypes: \textbf{Temporal} and \textbf{Contextual}.
    \item \textbf{Contradictory}: variants may appear applicable, but they cannot be jointly satisfied under the same target condition. During evaluation, the agent must recognize the underlying conflict and appropriately handle the resulting uncertainty from past experiences. 
\end{itemize}
The resolution target $\tau$, together with its semantic variant set $V_\tau$ and compatibility relation $r(V_\tau)$, forms a \textbf{latent semantic artifact} ($\tau$, $V_\tau$, $r(V_\tau)$) that is implicitly encoded into the benchmark and used to evaluate fine-grained memory discrimination in long-horizon interactions.

In Appendix~\ref{sec:appendix-relation-taxonomy-examples}, we use an example target to illustrate the target-conditioned variants set under three types of relations. 

\subsection{Evaluation Overview and Taxonomy}
\paragraph{User-history.}
We next operationalize semantic variants into long-horizon user-agent interaction histories.
Given a resolution target $\tau$ and its target-conditioned variant set $V_\tau$, each semantic variant is instantiated into a natural multi-turn user-agent conversation session. Rather than exposing facts directly as isolated memory entries, \textbf{the latent information gradually revealed itself through task-oriented conversations}, such as planning discussions, preference clarifications, or contextual problem-solving interactions (Appendix~\ref{sec:appendix-user-history-examples}). 

The instantiated sessions ${s_i}$ are then distributed across a multi-session conversation history
$
\mathbf{H} = {s_1, s_2, \dots, s_m},
$
where semantically related variants are hidden in different sessions separated by unrelated sessions. This design simulates long-running personal AI assistant usage, in which relevant memories are naturally scattered across long interaction histories.

\paragraph{Memory Injection.}

For each memory system $\alpha$, we replay a user history $\mathbf{H}$ by simulating its original memory formation process. Specifically, we first segment the interaction history into a sequence of history chunks $(H_1, H_2, \ldots, H_l)$ according to the memory formation granularity of the target system. For example, A-Mem~\citep{xu2025amem} constructs memory units at the message level, while Mem0~\citep{chhikara2025mem0} operates over message batches. We then sequentially feed these history chunks into the memory system following the original chronological order, allowing the system to incrementally construct its memory state as
$$
M^\alpha_t = \mathcal{M^\alpha}(M^\alpha_{t-1}, H_t),
$$
where $\mathcal{M}^\alpha$ and $M_t^\alpha$ denote the memory update function and memory state of system $\alpha$, respectively. For simplicity, we use $M=M(\textbf{H})$ to denote the final memory state after replaying the interaction history and omit $\alpha$.

\paragraph{Evaluation Instance.}
Each evaluation instance is associated with a latent semantic artifact $(\tau, V_\tau, r(V_\tau))$ and a query $q_\tau$ that requires agents to recover and correctly reason over the hidden relational structure among semantically related memory items.
When creating an evaluation instance, we use an LLM to generate the \textbf{reference correct answer set} $A^+$ and \textbf{wrong answer set} $A^-$ by conditioning on ($\tau$, $V_\tau$, $r(V_\tau)$), and the subset of target-relevant sessions $\textbf{H}_{\tau}$ from the user history:
\begin{equation}
(A^{+}, A^{-})
=
\mathcal{G}_{\text{LLM}}
\left(
\tau,\,
V_\tau,\,
r(V_\tau),\,
q_\tau,
\textbf{H}_{\tau}
\right)
\label{eq:eq1}
\end{equation}
During evaluation, the agent first retrieves task-relevant memory evidence
$$m_\tau = \mathcal{R}(M, q_\tau),$$
where $\mathcal{R}$ denotes the memory retrieval procedure and $M(H)$ denotes the memory state constructed from the interaction history $H$. Using $\pi$ to denote the response-generation model of the agent, the final answer is then generated as
\begin{equation}
a = \mathcal{\pi}(q_\tau, m_\tau) = \mathcal{\pi}(q_\tau, \mathcal{R}(M(\textbf{H}), q_\tau))
\label{eq:eq2}
\end{equation}
Comparing Eq.(\ref{eq:eq2}) to Eq.(\ref{eq:eq1}), we notice agents \textit{never} observe the latent semantic artifacts directly, but only the entire, raw interaction history. Consequently, SubtleMemory evaluates whether agents can faithfully preserve fine-grained semantic details during memory construction, and subsequently retrieve, distinguish, and reconcile the relevant memories under long-horizon interactions.

\paragraph{Answer Correctness.}
We  adopt an LLM-as-judge protocol to assign a binary correctness label based on whether the generated response correctly resolves the target implied by the query. The judge is provided with the latent semantic artifact $(\tau, V_\tau, r(V_\tau))$, the reference answer sets  ($A^+$, $A^-$), and relation-specific evaluation guidelines. 

\subsection{Construction Pipeline}

As shown in Figure~\ref{fig:data-construction}, we construct the benchmark through a five-stage pipeline that progressively encodes latent semantic artifacts into realistic long-horizon user--agent interaction histories. Each stage includes dedicated verifiers and filters to maintain conversational naturalness while preserving the intended memory relations. Appendix~\ref{sec:appendix-data-construction} provides data construction details.

\paragraph{Stage 1: Semantic Seed Selection.} 
We extract semantic fact seeds from high-quality open-source benchmarks. User-related seeds are derived from PersonaMem-v2~\citep{jiang2025personamemv2} user preference profiles, while non-user seeds are drawn from knowledge-oriented QA benchmarks, including FanOutQA~\citep{zhu2024fanoutqa}, MuSiQue~\citep{trivedi2022musique}, QACC~\citep{liu2025qacc}, HoH~\citep{ouyang2025hoh}, and AmbigQA~\citep{min2020ambigqa}. 

\paragraph{Stage 2: Semantic Variants Creation.}
For user-related seeds, we first use an LLM to determine a candidate relation (complementary, nuanced, or contradictory) and the corresponding resolution target. Conditioned on the selected relation and target, the LLM then generates semantic variants through detail enrichment or selective detail omission.  For non-user seeds, we conduct semantic neighborhood search over the seed corpus, identifying temporally related, contextually specialized, or multi-hop dependent facts that naturally fit the target compatibility relation. Additional variants are generated through controlled detail omission. 


\paragraph{Stage 3: Session Construction.}
For each fact variant, we generate a \textit{task-oriented} multi-turn user--assistant interaction session in which the fact is revealed implicitly and progressively through conversation, better reflecting realistic assistant usage. To promote interaction diversity, we define 10 task categories, each paired with three workflow patterns reflecting different user-agent interaction styles. 

\paragraph{Stage 4: Evaluation Instance Construction.}
For each latent semantic artifact, we generate evaluation queries together with reference correct and incorrect answers. Queries are designed to require relational reasoning over previously instantiated semantic variants. For non-user semantic facts, queries are formulated as knowledge-oriented questions. For user-related facts, queries are designed as either \textit{structured form filling} or \textit{resource arrangement} task, both of which naturally require agents to leverage user preference information accumulated through prior interactions. To verify this target dependence, we evaluate GPT-5.4 on all 815 non-user
instances using only the query and obtain 1.23\% accuracy, compared with
95.46\% when we provide the annotated history evidence, indicating that these
instances depend on interaction-history evidence
(Appendix~\ref{sec:memory-dependence}).


\paragraph{Stage 5: User-history Assembly.}

Finally, we assemble sessions into long-horizon chronological user histories in which semantically related variants are distributed across separated interactions interleaved with unrelated sessions. This process produces realistic long-context histories where relevant evidence remains naturally scattered throughout the interaction stream. 

\begin{table*}[t]
    \centering
    \footnotesize
    \setlength{\tabcolsep}{4.1pt}
    \renewcommand{\arraystretch}{1.14}
    \resizebox{0.98\textwidth}{!}{%
    \begin{tabular}{@{}lcccccccc@{}}
    \toprule
    \multirow{2}{*}{\textbf{Method}} & \multicolumn{3}{c}{\textbf{Complementary}} & \multicolumn{3}{c}{\textbf{Nuanced}} & \multirow{2}{*}{\textbf{Contradictory}} & \multirow{2}{*}{\textbf{All}} \\
    \cmidrule(lr){2-4}\cmidrule(lr){5-7}
    & \textbf{Multi-evidence} & \textbf{Any-one} & \textbf{Overall} & \textbf{Temporal} & \textbf{Contextual} & \textbf{Overall} \\
    \midrule
    \rowcolor{BackboneBlue}
    \multicolumn{9}{c}{\textbf{Base Model: GPT-5.4}} \\
    \midrule
    \quad MemoBase & 26.5\% & 43.3\% & 33.5\% & 33.3\% & 44.6\% & 38.9\% & 16.7\% & 32.1\% \\
    \quad MIRIX & 47.9\% & 52.0\% & 49.6\% & 51.8\% & 69.3\% & 60.5\% & 14.6\% & 46.5\% \\
    \quad MemOS & 51.2\% & 58.0\% & 54.0\% & 70.2\% & 77.8\% & 74.0\% & 23.9\% & 56.8\% \\
    \quad Mem0 & 61.1\% & 61.3\% & 61.2\% & 81.3\% & 85.8\% & 83.5\% & \underline{46.2\%} & 69.0\% \\
    \quad EverMemOS & 63.5\% & \underline{66.7\%} & 64.8\% & 81.8\% & \underline{87.4\%} & 84.6\% & 37.1\% & 68.1\% \\
    \quad A-Mem & 58.3\% & 61.3\% & 59.6\% & 83.1\% & 85.3\% & 84.2\% & \textbf{50.4\%} & \underline{70.0\%} \\
    \noalign{\vskip 0.18em}
    \cdashline{1-9}
    \noalign{\vskip 0.12em}
    \quad MetaClaw & 15.6\% & 20.0\% & 17.5\% & 17.7\% & 26.8\% & 22.2\% & 19.1\% & 20.3\% \\
    \quad OpenClaw & 62.6\% & 55.3\% & 59.6\% & 80.3\% & 83.0\% & 81.6\% & 25.5\% & 62.5\% \\
    \noalign{\vskip 0.18em}
    \cdashline{1-9}
    \noalign{\vskip 0.12em}
    \quad MemOS + OpenClaw & 57.3\% & \underline{66.7\%} & 61.2\% & 67.7\% & 80.2\% & 73.9\% & 15.9\% & 56.5\% \\
    \quad Mem0 + OpenClaw & \textbf{70.1\%} & 61.3\% & \underline{66.5\%} & \textbf{86.4\%} & 85.3\% & \underline{85.8\%} & 45.6\% & \textbf{71.3\%} \\
    \quad EverMemOS + OpenClaw & \underline{66.8\%} & \textbf{67.3\%} & \textbf{67.0\%} & \underline{85.4\%} & \textbf{89.7\%} & \textbf{87.5\%} & 32.6\% & 69.1\% \\
    \noalign{\vskip 0.18em}
    \cdashline{1-9}
    \noalign{\vskip 0.12em}
    \quad Oracle & 79.6\% & 87.3\% & 82.8\% & 95.5\% & 93.8\% & 94.6\% & 68.7\% & 85.4\% \\
    \midrule
    \rowcolor{BackboneViolet}
    \multicolumn{9}{c}{\textbf{Base Model: GPT-OSS-120B}} \\
    \midrule
    \quad MemoBase & 16.1\% & 38.7\% & 25.5\% & 27.8\% & 37.4\% & 32.5\% & 10.1\% & 25.3\% \\
    \quad MIRIX & 38.9\% & 47.3\% & 42.4\% & 42.7\% & 61.3\% & 51.9\% & 12.2\% & 39.8\% \\
    \quad MemOS & 48.8\% & \underline{58.7\%} & 52.9\% & 66.9\% & 69.8\% & 68.4\% & 18.0\% & 52.2\% \\
    \quad Mem0 & 57.8\% & \underline{58.7\%} & \underline{58.2\%} & \underline{80.1\%} & 79.4\% & \underline{79.7\%} & \textbf{40.6\%} & \textbf{64.9\%} \\
    \quad EverMemOS & 53.1\% & \textbf{60.7\%} & 56.2\% & 75.5\% & \underline{81.7\%} & 78.6\% & 26.8\% & 60.4\% \\
    \quad A-Mem & \underline{58.3\%} & 52.0\% & 55.7\% & \textbf{80.3\%} & 80.4\% & \textbf{80.4\%} & 37.7\% & \underline{63.9\%} \\
    \noalign{\vskip 0.18em}
    \cdashline{1-9}
    \noalign{\vskip 0.12em}
    \quad MetaClaw & 10.9\% & 16.7\% & 13.3\% & 13.6\% & 21.1\% & 17.3\% & 14.1\% & 15.6\% \\
    \quad OpenClaw & 46.4\% & 44.0\% & 45.4\% & 71.5\% & 69.8\% & 70.7\% & 20.7\% & 52.3\% \\
    \noalign{\vskip 0.18em}
    \cdashline{1-9}
    \noalign{\vskip 0.12em}
    \quad MemOS + OpenClaw & 40.8\% & 52.7\% & 45.7\% & 57.8\% & 70.6\% & 64.2\% & 15.9\% & 47.8\% \\
    \quad Mem0 + OpenClaw & \textbf{60.2\%} & 57.3\% & \textbf{59.0\%} & 72.2\% & 80.4\% & 76.3\% & \underline{39.0\%} & 62.9\% \\
    \quad EverMemOS + OpenClaw & 55.9\% & 57.3\% & 56.5\% & 74.2\% & \textbf{82.7\%} & 78.4\% & 24.1\% & 59.8\% \\
    \noalign{\vskip 0.18em}
    \cdashline{1-9}
    \noalign{\vskip 0.12em}
    \quad Oracle & 82.0\% & 78.7\% & 80.6\% & 90.9\% & 88.7\% & 89.8\% & 41.6\% & 75.7\% \\
    \bottomrule

    \end{tabular}
    }
    \caption{Main results on SubtleMemory. Answer correctness rates are grouped by answer-generation base model and reported by memory-relation type. Best/second non-oracle results per block bolded, underlined separately.}
    \label{tab:main-results}
    \vspace{-9pt}
    \end{table*}

\subsection{Final Data Composition.}
The final benchmark consists of 10 persona-level splits. Each split contains a long-horizon history of chronological user--agent interactions and a separate evaluation set grounded in latent semantic artifacts. In total, SubtleMemory contains 1,522 evaluation instances derived from 1,090 relation-controlled semantic variant sets, consisting of 361 complementary, 352 nuanced, and 377 contradictory sets. Each history contains an average of 236.4 memory-bearing sessions and 211.6K session tokens, creating long-context environments where target evidence is naturally interleaved with irrelevant or competing information. Queries span 10 domains, including culture, media, competition, world knowledge, society, cuisine, lifestyle, development, STEM, and nature. To assess benchmark-instance validity, we conduct a label-blind human audit of
150 stratified instances, yielding a 95.9\% overall validation rate and 94.5\%
inter-annotator agreement
(Appendix~\ref{sec:appendix-instance-human-validation}).

\begin{table*}[t]
\centering
\small
\setlength{\tabcolsep}{3pt}
\resizebox{0.98\textwidth}{!}{%
\begin{tabular}{llcccccccc}
\toprule
\multirow{2}{*}{\textbf{Model}} & \multirow{2}{*}{\textbf{Prompt}} & \multicolumn{3}{c}{\textbf{Complementary}} & \multicolumn{3}{c}{\textbf{Nuanced}} & \multirow{2}{*}{\textbf{Contradictory}} & \multirow{2}{*}{\textbf{All}} \\
\cmidrule(lr){3-5}\cmidrule(lr){6-8}
& & \textbf{Multi-evidence} & \textbf{Any-one} & \textbf{Overall} & \textbf{Temporal} & \textbf{Contextual} & \textbf{Overall} & & \\
\midrule
\multirow{2}{*}{GPT-4o-mini} & Soft & \textbf{86.4\%} & 54.5\% & 75.8\% & 91.7\% & 94.4\% & 93.1\% & 8.3\% & 67.4\% \\
 & Strong & 59.1\% & 54.5\% & 57.6\% & 69.4\% & 91.7\% & 80.6\% & 33.3\% & 63.1\% \\
\addlinespace[2pt]
\multirow{2}{*}{GPT-OSS-120B} & Soft & 77.3\% & \underline{72.7\%} & 75.8\% & 91.7\% & \underline{97.2\%} & 94.4\% & 11.1\% & 68.8\% \\
 & Strong & \textbf{86.4\%} & 63.6\% & \underline{78.8\%} & \underline{94.4\%} & \textbf{100.0\%} & \underline{97.2\%} & 41.7\% & 78.7\% \\
\addlinespace[2pt]
\multirow{2}{*}{GPT-5.4} & Soft & \underline{81.8\%} & \textbf{81.8\%} & \textbf{81.8\%} & \textbf{97.2\%} & \textbf{100.0\%} & \textbf{98.6\%} & \underline{63.9\%} & \underline{85.8\%} \\
 & Strong & \underline{81.8\%} & \textbf{81.8\%} & \textbf{81.8\%} & \textbf{97.2\%} & \textbf{100.0\%} & \textbf{98.6\%} & \textbf{80.6\%} & \textbf{90.1\%} \\
\bottomrule
\end{tabular}
}
\caption{Oracle-setting answer-generation calibration using one complete user-history split containing 141 evaluation queries. Best values are bolded; second-best distinct values are underlined, with ties marked equally.}
\label{tab:oracle-prompt-model-comparison-summary}
\vspace{-9pt}

\end{table*}

\section{Experiments}

\subsection{Experimental Setup}


\paragraph{Evaluated Systems.} We evaluate three deployment settings. (1) Six standalone memory systems: Mem0, MemOS, EverMemOS, MIRIX, A-Mem, and MemoBase~\citep{chhikara2025mem0,li2025memos,hu2026evermemos,wang2025mirix,xu2025amem,memobase2026docs}. (2) Two Claw-style agents with native memory mechanisms: OpenClaw and MetaClaw~\citep{openclaw2026docs,xia2026metaclaw}. (3) OpenClaw augmented with three plugin-based memory modules: Mem0, MemOS, and EverMemOS. 
Each system uses its default or author-recommended configuration, with detailed
settings in Appendix~\ref{sec:appendix-evaluated-system-settings}.

\paragraph{LLM-as-judge.}
We use Gemini 3.1 Pro Preview Thinking model as the LLM judge for answer evaluation. On a manually annotated benchmark containing 225 candidate answers, the judge achieves strong agreement with human annotations, with a Cohen’s $\kappa$ score of 0.963 (Appendix~\ref{sec:appendix-llm-judge-validation}).

\paragraph{Answer Generation.}
We tested two answer generation models: \texttt{gpt-5.4}
\citep{openai2026gpt54} and \texttt{gpt-oss-120b}
\citep{openai2025gptoss} in our main results. For all evaluated systems, we maintain consistency by using identical answer-generation instruction prompts (Appendix~\ref{sec:appendix-answer-generation-prompts}). Standalone memory systems receive answer-generation instructions directly within the assembled context, whereas OpenClaw-based agents receive the same instructions through preloaded markdown instruction files.

\paragraph{Oracle Setting.}
For each evaluation query $q_\tau$, we identify the user-agent sessions ${\textbf{H}_\tau}$ in which the target semantic variants $V_\tau$ are latently encoded, and directly provide these sessions as retrieval results for answer generation. This setting bypasses the memory system’s information extraction and retrieval processes, thereby approximating the upper-bound performance given perfect memory access.

\paragraph{Perfect Retrieval Setting}
To disentangle retrieval quality from memory construction, we introduce a perfect retrieval setting. After memory system $S$ builds its memory state from the complete user history, we provide the stored objects $\widetilde{m}_{\tau,S}$ written from ${\textbf{H}_\tau}$ instead of raw sessions. This bypasses retrieval while preserving memory-construction effects. Appendix~\ref{sec:appendix-oracle-perfect-retrieval-protocol} details the evidence source used in each setting.

\subsection{Answer Generation Configuration}
\label{sec:answer-calibration}

Under the oracle setting, answer generation performance depends primarily on the underlying model capability and prompting strategy. Using one complete user history from our benchmark (approximately 10\% of the full benchmark), we evaluate three models: \texttt{gpt-4o-mini} \citep{openai2024gpt4omini}, \texttt{gpt-oss-120b} \citep{openai2025gptoss}, and \texttt{gpt-5.4} \citep{openai2026gpt54}, under two prompting settings: a soft prompt with general guidance and a strong prompt with explicit instructions for target identification, conflict recognition, evidence fidelity, and clarification.

Table~\ref{tab:oracle-prompt-model-comparison-summary} shows that answer generation remains imperfect even under oracle memory access, suggesting that response generation itself constitutes a non-trivial limiting factor. Among all settings, \texttt{gpt-5.4} with the strong prompt achieves the best overall performance (90.1\%), substantially outperforming both \texttt{gpt-4o-mini} and \texttt{gpt-oss-120b}, but contradictory-relation cases remain noticeably harder than complementary and nuanced cases.


Based on these results, we adopt \texttt{gpt-5.4} with the strong prompt as the default answer-generation configuration in our main experiments, while additionally reporting results using \texttt{gpt-oss-120b}. This
choice controls answer-configuration variation across systems, allowing
subsequent comparisons to reflect differences among memory mechanisms under
a common answer-generation setting.

\vspace{-2pt}
\begin{table}[t]
\centering
\footnotesize
\providecommand{\deltacell}[2]{\begingroup\setlength{\fboxsep}{1.5pt}\colorbox{#1}{\makebox[3.3em][c]{#2}}\endgroup}
\providecommand{\gainweak}[1]{\deltacell{blue!8}{#1}}
\providecommand{\gainmid}[1]{\deltacell{blue!16}{#1}}
\providecommand{\gainstrong}[1]{\deltacell{blue!28}{#1}}
\providecommand{\lossweak}[1]{\deltacell{red!8}{#1}}
\providecommand{\lossmid}[1]{\deltacell{red!16}{#1}}
\providecommand{\lossstrong}[1]{\deltacell{red!28}{#1}}
\resizebox{\columnwidth}{!}{%
\begin{tabular}{cccccc}
\toprule
\textbf{System} & \textbf{Setting} & \textbf{Comp.} & \textbf{Nuanced} & \textbf{Contr.} & \textbf{All} \\
\midrule
\rowcolor{BackboneBlue}
\multicolumn{6}{c}{\textbf{Base Model: GPT-5.4}} \\
\midrule
\multirow{3}{*}{MemOS} & Base & 54.0 & 74.0 & 23.9 & 56.8 \\
& +OpenClaw & 61.2 & 73.9 & 15.9 & 56.5 \\
& $\Delta$ & \gainstrong{+7.2} & \lossweak{-0.1} & \lossstrong{-8.0} & \lossweak{-0.3} \\
\addlinespace[1pt]
\multirow{3}{*}{Mem0} & Base & 61.2 & 83.5 & 46.2 & 69.0 \\
& +OpenClaw & 66.5 & 85.8 & 45.6 & 71.3 \\
& $\Delta$ & \gainstrong{+5.3} & \gainmid{+2.3} & \lossweak{-0.6} & \gainmid{+2.3} \\
\addlinespace[1pt]
\multirow{3}{*}{EverMemOS} & Base & 64.8 & 84.6 & 37.1 & 68.1 \\
& +OpenClaw & 67.0 & 87.5 & 32.6 & 69.1 \\
& $\Delta$ & \gainmid{+2.2} & \gainmid{+2.9} & \lossmid{-4.5} & \gainweak{+1.0} \\
\midrule
\rowcolor{BackboneViolet}
\multicolumn{6}{c}{\textbf{Base Model: GPT-OSS-120B}} \\
\midrule
\multirow{3}{*}{MemOS} & Base & 52.9 & 68.4 & 18.0 & 52.2 \\
& +OpenClaw & 45.7 & 64.2 & 15.9 & 47.8 \\
& $\Delta$ & \lossstrong{-7.2} & \lossmid{-4.2} & \lossmid{-2.1} & \lossmid{-4.4} \\
\addlinespace[1pt]
\multirow{3}{*}{Mem0} & Base & 58.2 & 79.7 & 40.6 & 64.9 \\
& +OpenClaw & 59.0 & 76.3 & 39.0 & 62.9 \\
& $\Delta$ & \gainweak{+0.8} & \lossmid{-3.4} & \lossweak{-1.6} & \lossmid{-2.0} \\
\addlinespace[1pt]
\multirow{3}{*}{EverMemOS} & Base & 56.2 & 78.6 & 26.8 & 60.4 \\
& +OpenClaw & 56.5 & 78.4 & 24.1 & 59.8 \\
& $\Delta$ & \gainweak{+0.3} & \lossweak{-0.2} & \lossmid{-2.7} & \lossweak{-0.6} \\
\bottomrule
\end{tabular}
}
\caption{Effect of agent-runtime integration. Accuracy and $\Delta$ are
reported in percentage points; darker blue/red cells indicate larger gains/losses.}
\label{tab:openclaw-integration-effect}
\vspace{-10pt}
\end{table}

\subsection{Main Results}

\paragraph{Current memory systems remain substantially below Oracle performance on fine-grained relational memory discrimination.} Across both base model settings, the strongest standalone systems are consistently Mem0, EverMemOS, and A-Mem. Under \texttt{gpt-5.4}, A-Mem achieves the best overall performance (70.0\%), followed by Mem0 (69.0\%) and EverMemOS (68.1\%), yet still falls over 15 points behind the Oracle (85.4\%). The gap remains substantial across all relation types, with the best standalone systems trailing Oracle by 18.0\%, 10.0\%, and 18.3\% on complementary, nuanced, and contradictory relations, respectively. A similar trend holds under \texttt{gpt-oss-120b}.


\paragraph{Memory systems interact with agent context organization.}
Standalone memory systems generally outperform claw-style agents using their own native memory. Under \texttt{gpt-5.4}, native OpenClaw achieves only 62.5\% overall performance, substantially below top standalone systems such as EverMemOS (68.1\%) and Mem0 (69.0\%). Integrating strong memory plugins substantially improves agent performance, raising OpenClaw to 69.1\% with EverMemOS and 71.3\% with Mem0. However, Table~\ref{tab:openclaw-integration-effect} further shows that the interaction between memory systems and agent context organization is not uniformly beneficial, but rather highly task- and model-dependent. Under \texttt{gpt-5.4}, agent-driven context organization improves complementary category for MemOS by 7.2\%, but reduces contradictory category accuracy by 8.0\% relative to the standalone system. Under the weaker \texttt{gpt-oss-120b} base model, adding the agent layer is generally harmful.

\begin{figure*}[htbp]
\centering
\includegraphics[width=0.98\textwidth]{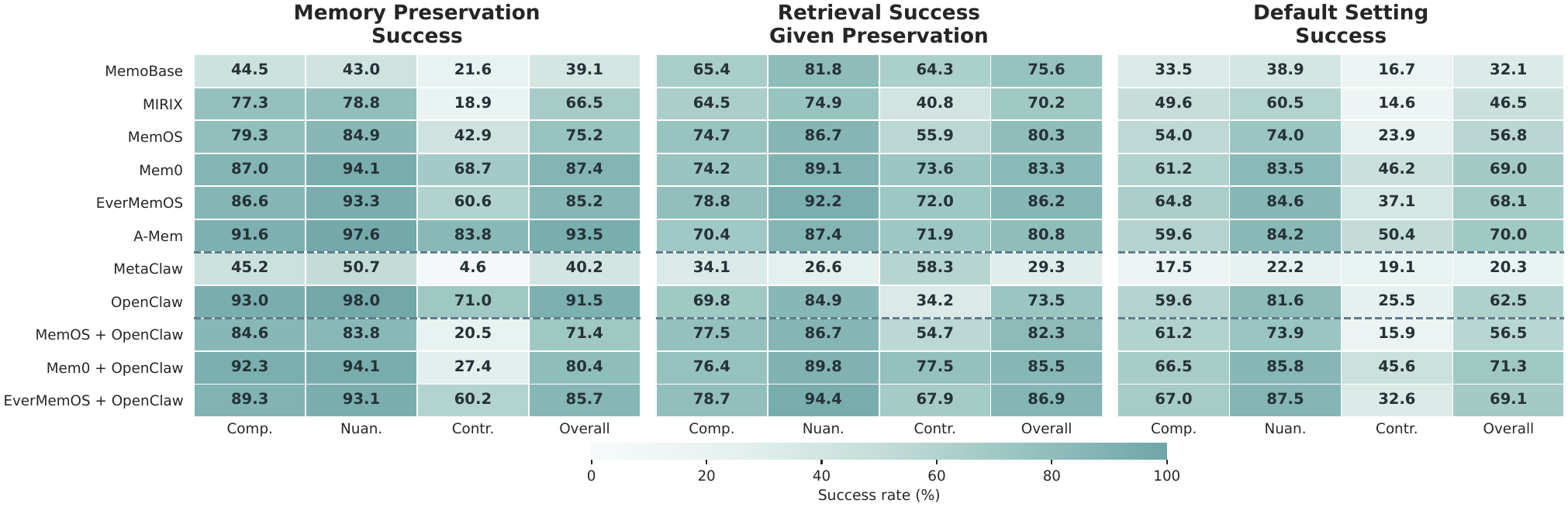}
\caption{Diagnostic waterfall analysis of memory system performance. Overall performance is decomposed into stage-wise memory preservation success rate $P_{\text{preserve}}$ and retrieval success rate conditioned on successful preservation $P_{\text{retrieve}}$. Rows correspond to the evaluated systems using GPT-5.4 as the answer generation model, sorted by overall accuracy.}
\label{fig:conditional-storage-retrieval}

\vspace{-14pt}
\end{figure*}

\paragraph{Contradictory relations constitute a greater challenge for current memory systems.} Unlike nuanced relations, whose performance approaches saturation under Oracle evidence access, contradictory relations exhibit a larger system-to-Oracle gap under gpt-5.4, with GPT-5.4 achieving only 68.7\% and \texttt{gpt-oss-120b} achieving 41.6\%. Large gaps between tested systems and Oracle performance further suggest that current memory mechanisms face greater difficulty handling contradictory evidence. Under \texttt{gpt-5.4}, A-Mem achieves the strongest contradictory performance (50.4\%), but still trails Oracle by 18.3 points.

\paragraph{Existing memory systems remain relatively weak at leveraging temporal information.} Within nuanced relations, under \texttt{gpt-5.4} and \texttt{gpt-oss-120b}, 10 and 9 out of 11 evaluated systems, respectively, achieve higher performance on contextual-detail discrimination than on temporal discrimination, whereas Oracle exhibits the opposite pattern under both base models. This contrast suggests that temporal reasoning remains an important opportunity for improving temporal-aware memory organization.

\section{Discussion}
Because memory preservation and retrieval operate sequentially, end-task accuracy alone cannot localize failure sources. We therefore perform a staged waterfall analysis. Let $S_O$ denote instances answered correctly under the oracle setting, filtering out answer-generation failures. Let $S_P \subseteq S_O$ denote instances that remain correct under perfect retrieval setting, indicating sufficient information is preserved. Therefore, the memory preservation success rate can be defined as 
$
P_{\text{preserve}} = \frac{|S_P|}{|S_O|}.
$
Finally let $S_D \subseteq S_P$ denote instances that remain correct under the default setting. Since instances in $S_P$ already contain sufficient preserved information, failures at this stage primarily reflect retrieval deficiencies. Hence the conditional retrieval success rate can be defined as
$
P_{\text{retrieve}} = \frac{|S_D|}{|S_P|}.
$
\paragraph{Memory preservation and retrieval jointly determine downstream performance.}
As shown in Figure~\ref{fig:conditional-storage-retrieval}, both $P_{\text{preserve}}$ and $P_{\text{retrieve}}$ are generally correlated with final accuracy. 
For example, MemoBase achieves low preservation performance ($P_{\text{preserve}} = 39.1\%$), but relatively strong retrieval performance ($P_{\text{retrieve}} = 75.6\%$) , resulting in only 32.1\% overall accuracy. 
The pattern becomes more evident when broken down by relation type. 
On contradictory-relation category, OpenClaw achieves strong memory preservation ($P_{\text{preserve}} = 71.0\%$), 
but weak retrieval performance ($P_{\text{retrieve}} = 34.2\%$), leading to only 25.5\% final accuracy.

\paragraph{Raw interaction preservation could improve memory fidelity.}
A-Mem and OpenClaw are the strongest memory-preservation systems, achieving overall preservation rates of 93.5\% and 91.5\%, respectively. A unique shared characteristic is that, beyond maintaining structured memory states, both systems also preserve the original interaction sessions, which retain fine-grained cues that compressed memory abstractions may lose and help resolve SubtleMemory queries requiring detail preservation. In contrast, MetaClaw shows weak preservation performance overall (40.2\%) and especially on contradictory cases (4.6\%), together with low conditional retrieval performance (29.3\%). Its memory mechanism emphasizes skill-like, experiential, and run-scoped abstractions, which are effective for reusable procedures but less aligned with factual relation-discrimination tasks that depend on exact details and competing records.



\paragraph{Empirical Observations and Insights.}
From Figure~\ref{fig:conditional-storage-retrieval}, we observe that different relations expose distinct memory bottlenecks. (1) Nuanced relations appear comparatively easier at the retrieval stage, potentially because they mainly require fine-grained discrimination among highly similar memories, where agents only need to identify the memory entry that best matches the target condition rather than retrieve all related evidence.
(2) Complementary and Contradictory relations appear more retrieval-intensive, potentially because successful resolution often requires aggregating and reconciling multiple related memories. (3) Contradictory relations appear particularly challenging for memory preservation, potentially because mutually conflicting facts are more likely to interfere with each other inside the memory state.

\section{Conclusion}
We introduced SubtleMemory, a benchmark for evaluating fine-grained relational memory discrimination in long-horizon AI agents. To make this capability measurable, SubtleMemory implicitly embeds latent relation-controlled semantic artifacts into user histories, requiring agents to recover complementary, nuanced, and contradictory relations from natural interaction traces. Our results show that current systems do not simply fail by missing relevant memories, but also lose relation-critical details, retrieve incomplete evidence, and struggle to use related memories correctly during answer generation. The diagnostic analysis further shows that these failures arise across answer generation, memory formation, and retrieval.
We hope SubtleMemory provides a focused testbed for building memory systems that sustain subtle memory resolution throughout long-horizon interaction.

\section*{Limitations}

\paragraph{Synthetic interaction histories.}
Despite our construction controls for increasing interaction diversity and
reducing structural shortcuts, the inherent limitations of LLM generation mean
that the resulting histories can only approximate naturally occurring
interaction logs. Future work could incorporate naturally collected histories
to better capture linguistic variability, behavioral irregularity, and evolving
interaction patterns.

\paragraph{Controlled downstream interaction.}
To maintain a controllable answer space, SubtleMemory focuses on controlled
single-output memory use and does not evaluate multi-turn downstream
interaction or broader autonomous agent behavior. Future work could extend the
benchmark to interactive task settings while preserving reliable evaluation.

\paragraph{Memory dynamics.}
Although evaluated systems process the fixed histories chronologically and
dynamically update their memory states, the benchmark does not independently
control repetition frequency, reinforcement strength, stability, or update
rate. Future work could vary these factors independently to examine their
effects on relation-sensitive memory use.

\paragraph{Evaluation scope.}
The current benchmark does not cover multimodal memories, multilingual
interactions, or highly domain-specific workflows, and its results depend on
the selected answer-generation and judging models. Future work could broaden
these settings and evaluate a more diverse range of models and judges.

\section*{Ethical Considerations}

\paragraph{Potential Risks.}
SubtleMemory is intended as an evaluation benchmark for long-term memory agents. As with other benchmarks, it may be over-optimized as a leaderboard target, which could limit generalization beyond the evaluated relation types and interaction settings. In addition, our automatic evaluation relies on LLM judges, which may introduce biases or occasional misjudgments.

\paragraph{Personally Identifying Information and Offensive Content.}
The benchmark is constructed from synthetic user-agent histories, LLM-generated annotations, and curated seed data from existing resources. Since seed data and generated interactions may contain sensitive, personally identifying, or inappropriate content, we apply filtering and manual inspection during construction to remove information that could name or uniquely identify individuals, as well as offensive or inappropriate content.

\paragraph{Instructions Given to Participants.}
This work did not involve recruited human participants, crowdworkers, or external annotators. Manual inspection and annotation were conducted by the authors, so no participant instructions, consent forms, or risk disclosures were required.

\section*{Acknowledgements}

This work was supported by the Shanghai Artificial Intelligence Laboratory and Shanghai Jiao Tong University.  This work was also supported by the National Science and Technology Major Project (Grant Nos. 2025ZD1606200 and 2025ZD1606203), the National Natural Science Foundation of China (Grant No. 92470205), and the Suzhou Science and Technology Development Program (SYG2024037).


\bibliography{custom}

\appendix


\definecolor{SubtleBoxBlue}{HTML}{EAF2FA}
\definecolor{SubtleBoxFrame}{HTML}{4D6F95}
\definecolor{SubtleBoxTitle}{HTML}{D8E6F5}
\definecolor{SubtleBoxGray}{HTML}{F6F7F9}
\definecolor{SubtleBoxGrayFrame}{HTML}{7A7F87}
\definecolor{SubtlePromptTitle}{HTML}{6F7680}
\definecolor{SubtleCaseTitle}{HTML}{4D6F95}

\newcommand{\SubtleParaGap}{\par\smallskip\noindent}

\newtcolorbox{subtlememorycasebox}[1]{%
  colback=SubtleBoxBlue,
  colframe=SubtleBoxFrame,
  colbacktitle=SubtleCaseTitle,
  coltitle=white,
  title=#1,
  halign title=left,
  arc=2mm,
  boxrule=0.45pt,
  left=6pt,
  right=6pt,
  top=6pt,
  bottom=6pt,
  fonttitle=\bfseries\normalsize,
  toptitle=3pt,
  bottomtitle=3pt,
  fontupper=\small
}

\newtcolorbox{subtlememoryplaincasebox}{%
  enhanced,
  breakable,
  colback=SubtleBoxBlue,
  colframe=SubtleBoxFrame,
  arc=2mm,
  boxrule=0.45pt,
  left=6pt,
  right=6pt,
  top=6pt,
  bottom=6pt,
  fontupper=\small
}

\newtcolorbox{subtlememorypromptbox}[1]{%
  colback=SubtleBoxGray,
  colframe=SubtleBoxGrayFrame,
  colbacktitle=SubtlePromptTitle,
  coltitle=white,
  title=#1,
  halign title=left,
  arc=2mm,
  boxrule=0.4pt,
  left=5pt,
  right=5pt,
  top=5pt,
  bottom=5pt,
  fonttitle=\bfseries\normalsize,
  toptitle=3pt,
  bottomtitle=3pt,
  fontupper=\footnotesize
}

\newtcolorbox{subtlememoryplainpromptbox}{%
  enhanced,
  breakable,
  colback=SubtleBoxGray,
  colframe=SubtleBoxGrayFrame,
  arc=2mm,
  boxrule=0.4pt,
  left=5pt,
  right=5pt,
  top=5pt,
  bottom=5pt,
  fontupper=\footnotesize
}

\providecommand{\subtleYes}{\Checkmark}
\providecommand{\subtlePart}{\HalfCircleRight}
\providecommand{\subtleNo}{\XSolidBrush}

\section*{Appendix Overview}

This appendix provides supporting material for the benchmark definition, data
construction process, and experimental analysis in the main paper.
\begin{itemize}
\item Appendix~\ref{sec:appendix-related-work} situates SubtleMemory within
prior work on memory-augmented LLM agents
(Appendix~\ref{sec:appendix-memory-augmented-llm-agents}) and long-term memory
benchmarks (Appendix~\ref{sec:appendix-long-term-memory-benchmarks}), then
summarizes the benchmark-positioning comparison
(Appendix~\ref{sec:appendix-benchmark-positioning} and
Table~\ref{tab:appendix-benchmark-comparison}).
\item Appendix~\ref{sec:appendix-methodology} expands the task formulation,
including the preliminary relation taxonomy
(Appendix~\ref{sec:appendix-relation-taxonomy-examples}), the evaluation
overview and taxonomy
(Appendix~\ref{sec:appendix-evaluation-overview-taxonomy}), and examples showing
how latent semantic variants are embedded into user-agent sessions
(Appendix~\ref{sec:appendix-user-history-examples}).
\item Appendix~\ref{sec:appendix-data-construction} documents the five-stage
construction pipeline: semantic seed selection
(Appendix~\ref{sec:appendix-data-stage-semantic-seed-selection}), semantic
variant creation
(Appendix~\ref{sec:appendix-data-stage-semantic-variants-creation}), session
construction (Appendix~\ref{sec:appendix-data-stage-session-construction}),
evaluation-instance construction
(Appendix~\ref{sec:appendix-data-stage-evaluation-instance-construction}), and
chronological user-history assembly
(Appendix~\ref{sec:appendix-data-stage-user-history-assembly}). These sections
also include the construction-model assignment, prompts, examples, and dataset
statistics, together with human validation
(Appendix~\ref{sec:appendix-instance-human-validation}), target-dependence
validation (Appendix~\ref{sec:memory-dependence}), and artifact availability,
licenses, and intended use
(Appendix~\ref{sec:appendix-artifact-license-use}).
\item Appendix~\ref{sec:appendix-experiments} provides additional experimental
details, including evaluated system settings
(Appendix~\ref{sec:appendix-evaluated-system-settings}), OpenClaw context
organization (Appendix~\ref{sec:appendix-openclaw-context-organization}),
LLM-as-judge validation (Appendix~\ref{sec:appendix-llm-judge-validation}),
answer-generation prompts (Appendix~\ref{sec:appendix-answer-generation-prompts}),
oracle and perfect-retrieval protocols
(Appendix~\ref{sec:appendix-oracle-perfect-retrieval-protocol}), answer-model
configuration (Appendix~\ref{sec:appendix-answer-generation-configuration}),
statistical analysis
(Appendix~\ref{sec:appendix-statistical-analysis}), repeated-run robustness
(Appendix~\ref{sec:appendix-repeated-run-robustness}), history-level reliability
(Appendix~\ref{sec:appendix-history-level-reliability}), answer-model and judge
robustness (Appendix~\ref{sec:appendix-answer-model-judge-robustness}), case studies
(Appendix~\ref{sec:appendix-main-experiment-case-studies}), detailed
perfect-retrieval results
(Appendix~\ref{sec:appendix-perfect-retrieval-detailed-results}), and
representative answer examples (Appendix~\ref{sec:appendix-answer-examples}).
\item Appendix~\ref{sec:appendix-ai-assistant-use} describes the roles of AI
assistants in benchmark construction, evaluation, and manuscript editing.
\end{itemize}

\section{Related Work}
\label{sec:appendix-related-work}

\subsection{Memory-augmented LLM agents}
\label{sec:appendix-memory-augmented-llm-agents}
Long-term assistant agents use memory to keep prior interactions available as
reusable agent state rather than transient context. Early agent systems explored
external memory for personalization, reflection, experience reuse, and
open-ended task continuation
\citep{park2023generative,zhong2023memorybank,shinn2023reflexion,zhao2024expel,wang2024voyager,majumder2023clin}.
More recent memory systems make memory management itself an explicit design
problem. They study how memories should be written, consolidated, organized,
updated, compressed, and retrieved through virtual-context memory managers,
production memory services, hierarchical memory, temporal knowledge graphs,
agentic note networks, multi-agent memory routing, memory-OS abstractions, and
lightweight consolidation mechanisms
\citep{packer2023memgpt,chhikara2025mem0,rasmussen2025zep,xu2025amem,sun2026hmem,zhang2026himem,wang2025mirix,li2025memos,fang2026lightmem,hu2026evermemos}.
Claw-style agents introduce another deployment form, where memory can be
maintained by the native agent runtime or supplied by plugin memory modules that
inject recalled information into the runtime context
\citep{openclaw2026docs,xia2026metaclaw}. These designs improve persistence,
organization, and recall, but relation-sensitive behavior is usually left
implicit in retrieval scores, summaries, links, or routing decisions. It
therefore remains unclear whether current memory-augmented agents preserve and
retrieve the distinctions among similar memories when those distinctions
are required during downstream reasoning.

\subsection{Benchmarks for long-term memory systems}
\label{sec:appendix-long-term-memory-benchmarks}
Memory evaluation has moved from long-context use in static inputs to
multi-session and agent-facing memory use. Long-context benchmarks measure
retrieval and reasoning over long inputs
\citep{bai2024longbench,an2024leval,hsieh2024ruler}. Long-term memory
benchmarks then evaluate whether agents can retain conversation history, update
user preferences, apply personalized facts, perform memory operations, or use
stored state during later tasks
\citep{maharana2024locomo,wu2025longmemeval,jiang2025personamemv2,tan2025membench,hu2026memoryagentbench,bian2026realmem}.
More recent benchmarks further stress dynamic memory use in multi-session,
task-oriented, and evolving information environments
\citep{shen2026evolmem,shen2026mem2actbench,he2026memoryarena,ji2026clawarena}, including
ClawArena's emphasis on multi-source conflict, belief revision, and implicit
personalization over long-horizon interactions. 
Together, these benchmarks broaden long-term memory evaluation, with some
partially or implicitly capturing interactions among multiple memories through
multi-session reasoning, updating, or evolving histories. However, they do not
systematically define and control how multiple memories relevant to the same
downstream target should be aggregated, distinguished, or retained as
unresolved conflict.

\subsection{Positioning of SubtleMemory}
\label{sec:appendix-benchmark-positioning}
To fill this gap, we introduce SubtleMemory, a benchmark for evaluating fine-grained relational memory discrimination during downstream reasoning. SubtleMemory organizes each instance around a resolution target and a target-conditioned semantic variant set, and explicitly controls whether the relevant memories are complementary, nuanced, or contradictory. Table~\ref{tab:appendix-benchmark-comparison} compares SubtleMemory with representative benchmarks across different styles of long-term memory evaluation.
The comparison dimensions are:
\begin{itemize}[leftmargin=*, itemsep=2pt, topsep=2pt]
\item \textit{Memory source}, \textit{interaction form}, and \textit{primary
evaluation target} describe the benchmark input, interaction setting, and task.
\item \textit{Controlled dependency} names the dependency explicitly constructed
to determine the answer.
\item The relation rows ask whether the benchmark explicitly tests
target-conditioned related memories, complementary aggregation, nuanced
context/time discrimination, and contradictory conflict preservation.
\item The diagnostic rows compare whether the benchmark separates
preservation, retrieval, and answer-generation behavior.
\end{itemize}


\begin{table*}[!tbp]
\centering
\scriptsize
\setlength{\tabcolsep}{4pt}
\renewcommand{\arraystretch}{1.16}
\resizebox{0.98\textwidth}{!}{%
\begin{tabular}{@{}p{0.23\textwidth}cccccc@{}}
\toprule
\textbf{Comparison dimension} &
\textbf{LoCoMo} &
\textbf{LongMemEval} &
\textbf{PersonaMem-v2} &
\textbf{Mem2ActBench} &
\textbf{ClawArena} &
\textbf{SubtleMemory} \\
\midrule
\textbf{Memory source} &
User dialog &
User history &
Persona prefs. &
Tool history &
Workspace evidence &
\textbf{User + external} \\
\midrule
\textbf{Interaction form} &
Multi-session &
Sessions + distractors &
Long personalization &
Interrupted tools &
Evolving rounds &
\textbf{Task sessions} \\
\midrule
\textbf{Primary evaluation target} &
Recall/summary &
QA/abstention &
Preference inference &
Tool action &
Conflict/revision &
\textbf{Relation-sensitive answer} \\
\midrule
\textbf{Controlled dependency} &
Implicit evidence &
Evidence/update &
Preference evolution &
Memory chain &
Source/update/pref. &
\textbf{Target-conditioned variants} \\
\midrule
\textbf{Target-conditioned related memories} &
\subtlePart &
\subtlePart &
\subtleNo &
\subtlePart &
\subtlePart &
\textbf{\subtleYes} \\
\midrule
\textbf{Complementary aggregation} &
\subtlePart &
\subtlePart &
\subtleNo &
\subtlePart &
\subtlePart &
\textbf{\subtleYes} \\
\midrule
\textbf{Nuanced context/time discrimination} &
\subtlePart &
\subtlePart &
\subtlePart &
\subtleNo &
\subtlePart &
\textbf{\subtleYes} \\
\midrule
\textbf{Contradictory conflict preservation} &
\subtleNo &
\subtlePart &
\subtleNo &
\subtlePart &
\subtleYes &
\textbf{\subtleYes} \\
\midrule
\textbf{Explicit relation taxonomy} &
\subtleNo &
\subtleNo &
\subtleNo &
\subtleNo &
\subtlePart &
\textbf{\subtleYes} \\
\midrule
\textbf{Preservation/retrieval/answer diagnostics} &
\subtleNo &
\subtlePart &
\subtleNo &
\subtlePart &
\subtlePart &
\textbf{\subtleYes} \\
\midrule
\textbf{Main diagnostic granularity} &
Answer &
Retrieval+answer &
Preference type &
Tool+retrieval &
Category reliability &
\textbf{Stage waterfall} \\
\bottomrule
\end{tabular}
}
\caption{Transposed feature comparison between SubtleMemory and representative
long-term memory benchmarks from different evaluation styles: multi-session
conversation (LoCoMo~\citep{maharana2024locomo}), long-term memory QA
(LongMemEval~\citep{wu2025longmemeval}), personalization
(PersonaMem-v2~\citep{jiang2025personamemv2}), memory-grounded tool use
(Mem2ActBench~\citep{shen2026mem2actbench}), and Claw-style agent evaluation
(ClawArena~\citep{ji2026clawarena}). \subtleYes{} denotes an explicit benchmark
feature, \subtlePart{} denotes partial or implicit coverage, and \subtleNo{}
denotes that the feature is not a primary target.}
\label{tab:appendix-benchmark-comparison}
\end{table*}

\section{Methodology}
\label{sec:appendix-methodology}

\subsection{Preliminary Concepts and Relation Taxonomy}
\label{sec:appendix-relation-taxonomy-examples}

Consider the semantic seed:
\begin{quote}
\textit{Bonita prefers Japanese minimalist interior design.}
\end{quote}
For the resolution target of identifying Bonita's applicable interior-design
preference for an apartment renovation plan, SubtleMemory can construct
different target-conditioned semantic variant sets around the same seed
depending on the intended memory relation.

\paragraph{Complementary relation.}
The variant set may include:
\begin{itemize}
    \item \textit{Bonita prefers Japanese minimalist interiors with light wood
    furniture.}
    \item \textit{Bonita prefers minimalist interiors with neutral color
    palettes.}
    \item \textit{Bonita prefers low-clutter rooms with concealed storage.}
\end{itemize}
These memories are mutually compatible, so the agent should aggregate them into
a single renovation brief.

\paragraph{Nuanced relation.}
The variant set may include:
\begin{itemize}
    \item \textit{Bonita prefers Japanese minimalist interiors for her
    apartment.}
    \item \textit{Bonita prefers Scandinavian minimalist layouts for workshop
    spaces.}
    \item \textit{Bonita prefers bold industrial styling for pop-up exhibition
    booths.}
\end{itemize}
These memories are topically similar, but the apartment-renovation target
selects only the home-interior preference.

\paragraph{Contradictory relation.}
The variant set may include:
\begin{itemize}
    \item \textit{Bonita prefers Japanese minimalist interiors for the apartment
    renovation.}
    \item \textit{Bonita no longer wants Japanese minimalist interiors for the
    apartment and now prefers a maximalist vintage style.}
\end{itemize}
These memories cannot be jointly satisfied under the same target, so the agent
should recognize the conflict and ask for clarification instead of silently
choosing one side.

\subsection{Evaluation Overview and Taxonomy}
\label{sec:appendix-evaluation-overview-taxonomy}

\subsubsection{Embedding Semantic Variants into User Sessions}
\label{sec:appendix-user-history-examples}

Semantic variants are not exposed to the evaluated agent as isolated memory
entries. Instead, each variant is embedded into a natural task-oriented
user-agent session, where the relevant information becomes recoverable from the
user's goal, constraints, corrections, and concrete details. The following two
examples show this process for a user-related nuanced-relation case and a
non-user complementary-relation case.
Figure~\ref{fig:appendix-user-related-session-embedding}
shows the user-related case, and
Figure~\ref{fig:appendix-non-user-session-embedding} shows the non-user case.

\begin{figure*}[!tbp]
\begin{subtlememorycasebox}{User-related Task-oriented Session Embedding}
\small
\textbf{Case.} Amara uses different design styles depending on whether she is
shaping her home environment or creating materials and spaces for kids' science
activities.

\textbf{Target-conditioned variants.}
\begin{itemize}
    \item At home, Amara prefers minimalist, Scandinavian-inspired design with
    clean lines, light wood tones, and an uncluttered feel.
    \item When designing STEM workshops or toy car race setups for children,
    Amara intentionally uses bold colors, interactive visual cues, and playful
    layouts to keep the space energetic and engaging.
\end{itemize}

\textbf{Session embedding.} Each variant is assigned to a separate
task-oriented session. The session does not state the variant as a direct memory
sentence; instead, it makes the variant recoverable through the user's task and
interaction details.

\noindent
\begin{minipage}[t]{0.47\linewidth}
\raggedright\sloppy
\textbf{Session 1: \texttt{workflow\_setup}}
\par\vspace{0.2em}\hrule\vspace{0.35em}
\emph{Assigned variant.} Home-design preference.

\SubtleParaGap
\emph{User.} I need a repeatable way to stop overbuying decor for my apartment.
Cute objects keep sneaking in and then the room starts feeling visually noisy.

\SubtleParaGap
\emph{Assistant.} A good fix is a tiny system instead of relying on willpower.
\ldots

\SubtleParaGap
\emph{User.} The things that keep surviving every cleanup are the pieces with
simple lines, lighter wood, soft neutrals, maybe one muted green. When tabletops
get crowded, my brain does too.

\SubtleParaGap
\emph{Assistant.} That gives us a very clear filter. Here's a first pass.
\ldots
\end{minipage}
\hfill
\begin{minipage}[t]{0.47\linewidth}
\raggedright\sloppy
\textbf{Session 2: \texttt{information\_organization}}
\par\vspace{0.2em}\hrule\vspace{0.35em}
\emph{Assigned variant.} Workshop-design preference.

\SubtleParaGap
\emph{User.} My notes for a kids' toy-car physics event look like a lab
notebook went through turbulence. I need help turning the pile into a clean
structure.

\SubtleParaGap
\emph{Assistant.} Let's turn the turbulence into lanes. A good first sort would
be: \ldots

\SubtleParaGap
\emph{User.} Yes, that split is closer. A few non-negotiables: bright lane
colors, big picture-based instructions, arrows or footprints that tell the
cars-and-kids where to move, and separate little track islands instead of one
long teacher table.

\SubtleParaGap
\emph{Assistant.} Absolutely. Based on what you've said, here's the cleaner
sort: \ldots
\end{minipage}
\end{subtlememorycasebox}
\caption{User-related session embedding for a nuanced contextual variant set.
Both sessions are topically about design, but only the home-design session
should guide an apartment-design request.}
\label{fig:appendix-user-related-session-embedding}
\end{figure*}

\begin{figure*}[!tbp]
\begin{subtlememorycasebox}{Non-user Task-oriented Session Embedding}
\small
\textbf{Case.} The answer requires matching several Korean metropolitan cities
with their symbol flowers.

\textbf{Target-conditioned variants.}
\begin{itemize}
    \item Busan's symbol flower is Camellia.
    \item Ulsan's symbol flower is Pear flower.
    \item Incheon's symbol flower is Rose.
\end{itemize}

\textbf{Session embedding.} Each variant is assigned to a task-oriented session.
The session does not state the variants as a benchmark list; instead, it makes
the target facts recoverable through ordinary learning and artifact-production
tasks.

\noindent
\begin{minipage}[t]{0.48\linewidth}
\raggedright\sloppy
\textbf{Session 1: \texttt{learning\_explanation}}
\par\vspace{0.2em}\hrule\vspace{0.35em}
\emph{Assigned variant.} Busan and Ulsan symbol-flower facts.

\SubtleParaGap
\emph{User.} I'm helping my younger cousin study Korean city symbols. Can you
explain in simple terms what a city's symbol flower is, and how to make the
idea easy to remember?

\SubtleParaGap
\emph{Assistant.} A city's symbol flower is an official flower the city uses to
represent itself, a bit like a visual emblem. It usually reflects the image the
city wants to convey, so the easiest way to remember it is: not "favorite
flower," but "flower that stands for the city."

\SubtleParaGap
\emph{User.} That helps, but I learn better with examples. Could you use one
metropolitan city so I can hear how you'd explain it out loud?

\SubtleParaGap
\emph{Assistant.} Sure. If I were explaining it with Busan, I'd say that
Busan's symbol flower is Camellia. Then I'd add that this is just one example
of how a city chooses a flower to act as part of its civic identity, so you're
connecting a place name with a representative image.

\SubtleParaGap
\emph{User.} Nice. Could you give me a second city too? I think comparing two
examples would make the concept stick better.

\SubtleParaGap
\emph{Assistant.} Yes, comparison works well here. You could pair Busan with
Ulsan and point out that Ulsan's symbol flower is Pear flower. That shows the
main idea: each city has its own emblematic flower, so the point isn't to
memorize flowers in general, but to see that the symbol belongs to that specific
city.

\SubtleParaGap
\emph{User.} Good approach. One last thing: can you give me a memory trick for
the Ulsan example? That's the one I'm more likely to forget.

\SubtleParaGap
\emph{Assistant.} A simple trick is to link the word "pear" with the image of a
light spring blossom. If you picture Ulsan paired with a pear blossom image, it
becomes easier to recall that Ulsan's symbol flower is Pear flower.
\end{minipage}
\hfill
\begin{minipage}[t]{0.48\linewidth}
\raggedright\sloppy
\textbf{Session 2: \texttt{artifact\_production}}
\par\vspace{0.2em}\hrule\vspace{0.35em}
\emph{Assigned variant.} Incheon symbol-flower fact.

\SubtleParaGap
\emph{User.} I’m putting together a small cultural display and need a short
caption for an Incheon profile card. Can you help me draft one?

\SubtleParaGap
\emph{Assistant.} Sure — here’s a clean starting option: "Incheon, a dynamic
gateway city, blends modern energy with local pride and tradition."

\SubtleParaGap
\emph{User.} Nice start. I want it to feel a little more specific though, like
something that belongs on a city handout card.

\SubtleParaGap
\emph{Assistant.} Absolutely. Here’s a more specific version: "Incheon is a
vibrant city known for its forward-looking spirit; its symbol flower is the
rose, reflecting beauty and civic pride."

\SubtleParaGap
\ldots

\SubtleParaGap
\emph{User.} I like that. Could you give me two final options: one plain and
one a little more polished? Keep both short.

\SubtleParaGap
\emph{Assistant.} Sure: Plain: "Incheon’s symbol flower is the rose."
Polished: "The rose, Incheon’s symbol flower, reflects the city’s beauty and
pride."

\SubtleParaGap
\emph{User.} Great, I’ll use the polished one. Can you give it to me one last
time with no extra commentary so I can paste it into the layout?

\SubtleParaGap
\emph{Assistant.} The rose, Incheon’s symbol flower, reflects the city’s beauty
and pride.
\end{minipage}
\end{subtlememorycasebox}
\caption{Non-user session embedding for a complementary multi-evidence variant
set. The required facts are distributed across ordinary task-oriented
interactions rather than exposed as a benchmark list.}
\label{fig:appendix-non-user-session-embedding}
\end{figure*}

\section{Data Construction}
\label{sec:appendix-data-construction}
\label{sec:appendix-construction-pipeline-details}

This section records the concrete source criteria, prompt interfaces, filtering
rules, representative examples, and final composition used to instantiate the
construction pipeline.

\paragraph{Construction model assignment.}
Stage 1 selects semantic seeds from source datasets, while Stage 5 assembles
the accepted sessions into chronological histories; neither stage introduces
new LLM-generated content. In Stages 2--4, we use \texttt{gpt-5.4} for relation
planning and all generation steps, including semantic-variant generation,
session construction, and evaluation-query and answer-candidate generation.
We use \texttt{gemini-3.1-pro-preview-thinking} for all LLM-based filters
applied to semantic variants, sessions, queries, and answer candidates. Both
models use temperature 0, and \texttt{gpt-5.4} uses a maximum output length of
16,384 tokens. We assign generation and filtering to different model families
to reduce same-family self-validation. Table~\ref{tab:appendix-answer-model-settings}
reports the model identifiers and settings, while the stage-specific prompts
are provided in the following subsections.

\subsection{Stage 1: Semantic Seed Selection}
\label{sec:appendix-data-stage-semantic-seed-selection}

\paragraph{User-related seed selection.}
\label{sec:appendix-data-stage1-user-related}
User-related seeds come from PersonaMem-v2
\citep{jiang2025personamemv2}. We retain the following source fields and
selection metadata:
\begin{itemize}
\item We select ten personas with distinct backgrounds and application domains
(Table~\ref{tab:appendix-selected-personas}).
\item The retained seed unit is a persona, topic, and source preference.
\item We keep the sanitized profile as user background, remove raw
conversations and update metadata, and use the active preference as the
semantic seed.
\item Figure~\ref{fig:appendix-user-related-source} shows one retained profile
and its preference fields.
\end{itemize}

\begin{table*}[!t]
\centering
\small
\begin{tabular}{@{}p{0.12\textwidth}p{0.25\textwidth}p{0.56\textwidth}@{}}
\toprule
\textbf{Persona ID} & \textbf{Selected Persona} & \textbf{Domain Coverage} \\
\midrule
0 & Amara Nwosu & STEM education, physics, mechanical engineering, youth outreach \\
334 & Dr. Suthida Rattanaporn & Healthcare, nursing education, curriculum design \\
351 & Gareth McAllister & Agriculture, rural work, family business, stress management \\
341 & Malik Thompson & Art, politics, social justice, urban culture \\
92 & Julian Marcus Ortega & NLP, computational social science, social media \\
637 & Lars Henriksen & Professional sports, endurance training, performance optimization \\
45 & Isabel Moraes Santiago & Elderly persona, cultural heritage, language preservation \\
996 & Christopher ``Chris'' Halberg & Law, institutional accountability, professional fallibility \\
135 & Adriana Ionescu & Migration, cross-cultural adaptation, urban planning \\
144 & Dr. Amira Suryani & Neuroscience, speculative technology, ethics, consciousness research \\
\bottomrule
\end{tabular}
\caption{Selected PersonaMem-v2 personas used for user-related semantic seed
selection. The set increases coverage across demographic backgrounds,
occupations, cultures, and application domains.}
\label{tab:appendix-selected-personas}
\end{table*}

\begin{figure*}[!p]
\begin{subtlememorycasebox}{User-related source record and raw preferences}
\scriptsize
\noindent
\begin{minipage}[t]{0.47\linewidth}
\raggedright\sloppy
\textbf{Selected example persona profile fields.}

\SubtleParaGap
\texttt{persona\_id}: 0; \texttt{name}: Amara Nwosu.

\SubtleParaGap
\texttt{short\_persona}: A science whiz who loves to incorporate physics
concepts into unique toy car races.

\SubtleParaGap
\texttt{age}: 28; \texttt{gender}: Female; \texttt{nationality}:
Nigerian-American; \texttt{languages\_spoken}: English, Igbo, French.

\SubtleParaGap
\texttt{personality.traits}: curious, playfully competitive, innovative,
patient, analytical, warm.

\SubtleParaGap
\texttt{personality.hobbies}: custom racetracks, aerodynamics in miniature
models, science fairs, vintage toy cars, astronomy stargazing nights.

\SubtleParaGap
\texttt{personality.quirks}: names toy cars after famous scientists, keeps a
small chalkboard for equations, and uses physics analogies casually.

\SubtleParaGap
\texttt{occupation}: STEM Education Program Designer at the Smithsonian Science
Education Center, focused on interactive STEM workshops, hands-on physics
activities, and museum or school outreach.

\SubtleParaGap
Additional fields include \texttt{education}, \texttt{demographics},
\texttt{speaking\_style}, \texttt{therapy\_background},
\texttt{health\_and\_medical\_conditions}, and
\texttt{sensitive\_information}.
\end{minipage}
\hfill
\begin{minipage}[t]{0.47\linewidth}
\raggedright\sloppy
\textbf{Raw PersonaMem-v2 preference fields.}

\SubtleParaGap
\texttt{stereotypical\_preferences}: ``Values strong ties to extended
family traditions,'' ``Loves spicy jollof rice and suya,'' ``Follows prominent
African diaspora authors and speakers,'' ``Engages in traditional Nigerian
holiday celebrations,'' ``Reads African literature classics,'' ``Participates
in cultural dance workshops,'' ``Enjoys drumming rhythms in community events,''
\ldots

\SubtleParaGap
\texttt{anti\_stereotypical\_preferences}: ``Enjoys minimalist and
Scandinavian-inspired home design,'' ``Prefers quiet coffee shops over loud
social events,'' ``Loves indie folk and ambient music,'' ``Enjoys watching
Nordic noir detective series,'' ``Prefers muted neutral fabrics in clothing,''
``Practices yoga in tranquil environments,'' \ldots

\SubtleParaGap
\texttt{neutral\_preferences}: ``Enjoys reading science fiction novels,''
``Loves homemade pizza nights,'' ``Appreciates well-written mystery novels,''
``Likes learning new languages casually,'' ``Prefers reusable eco-friendly
shopping bags,'' ``Likes solving crosswords and logic puzzles,'' ``Enjoys
planning and organizing events,'' \ldots
\end{minipage}
\end{subtlememorycasebox}
\caption{Profile fields and raw PersonaMem-v2 preference fields used as
user-related semantic seeds for the selected example persona. Additional fields
and preferences are omitted for space.}
\label{fig:appendix-user-related-source}
\end{figure*}

\paragraph{Non-user seed selection.}
\label{sec:appendix-data-stage1-user-unrelated}
Non-user seeds are retained only when their evidence structure can support
relation-controlled variants. The retained source filters are:
\begin{itemize}
\item FanOutQA~\citep{zhu2024fanoutqa}: development samples with at least five
linked evidence or subquestion items.
\item MuSiQue~\citep{trivedi2022musique}: answer-classified development
samples with four linked evidence items, after duplicate-heavy items are
cleaned down to 153 retained samples.
\item QACC~\citep{liu2025qacc}: 796 retained records from roughly 1.7K
candidates, requiring multiple entries with the same answer surface form.
\item AmbigQA~\citep{min2020ambigqa}: records with at least four QA pairs for
context-conditioned candidates after manual filtering, and records with exactly
three QA pairs for contradiction candidates.
\item HoH~\citep{ouyang2025hoh}: records with at least three outdated
information entries for temporal variants.
\end{itemize}
Figure~\ref{fig:appendix-user-unrelated-source} shows one retained non-user
source record and the selected semantic facts.

\begin{figure*}[!tbp]
\begin{subtlememorycasebox}{Non-user source record and selected facts}
\emph{Raw source shape.} External-knowledge inputs contain a source-style
question and evidence items. One retained source question asks, ``What are the
names of six Metropolitan cities in Korea and their respective symbol
flowers?''

\emph{Selected semantic facts after fact selection.}
Busan's symbol flower is Camellia. Incheon's symbol flower is Rose. Ulsan's
symbol flower is Pear flower.
\end{subtlememorycasebox}
\caption{Non-user source question and selected external-knowledge facts used as
semantic seeds for variant creation.}
\label{fig:appendix-user-unrelated-source}
\end{figure*}

\subsection{Stage 2: Semantic Variant Creation}
\label{sec:appendix-data-stage-semantic-variants-creation}

\paragraph{User-related semantic variant creation.}
\label{sec:appendix-data-stage2-user-related}
For user-related variants, the planner first assigns one compatibility relation
type and subtype to each source preference under target count constraints. The
generator then receives the planned relation, source preference, persona,
relation definitions, and boundary rules, and must return the same relation
with a concise case description and one-sentence variants. The filter removes
unsupported, duplicated, mislabeled, internally invalid, unnatural, or
order-dependent variant sets before any session is written.
Figures~\ref{fig:appendix-relation-planning-prompt},
\ref{fig:appendix-case-generation-prompt}, and
\ref{fig:appendix-case-filter-prompt} show the prompts, and
Figure~\ref{fig:appendix-user-related-case-examples} shows accepted examples.

\begin{figure*}[!tbp]
\begin{subtlememorypromptbox}{User-related relation-planning prompt excerpt}
\emph{``Plan the relation type and subtype for each source preference before
semantic variant generation.''}
\emph{``Persona: \ldots''}
\emph{``Persona id: \ldots''}
\emph{``Relation type definitions: \ldots''}
\emph{``Relation subtype definitions: \ldots''}
\emph{``Boundary rules: \ldots''}
\emph{``Target relation type counts: \ldots''}
\emph{``Subtype balancing policy: \ldots''}
\emph{``Preferences: \ldots''}

\emph{``Requirements:''}
\begin{itemize}
\setlength{\itemsep}{2pt}
\setlength{\parsep}{0pt}
\setlength{\topsep}{2pt}
\item Assign exactly one \texttt{relation\_type} and \texttt{relation\_subtype}
to every \texttt{preference\_id}.
\item Match the target relation type counts exactly.
\item Within each \texttt{relation\_type}, split \texttt{relation\_subtype}
assignments as evenly as possible following the fixed subtype order above.
\item Use the fixed subtype order to place any remainder within a
\texttt{relation\_type}.
\item Within these quantity constraints, choose the most natural
preference-subtype pairing.
\item Prefer pairings that can produce faithful, natural cases with the least
unsupported invention.
\item Do not invent synthetic facts, topics, task labels, or unsupported
conflicts.
\end{itemize}

\emph{``Think step by step. Return only JSON:
\{\texttt{planned\_relations}: [\ldots]\}.''}
\end{subtlememorypromptbox}
\caption{User-related relation-planning prompt for assigning compatibility
relation type and subtype before variant generation.}
\label{fig:appendix-relation-planning-prompt}
\end{figure*}

\begin{figure*}[!p]
\begin{subtlememorypromptbox}{User-related variant-generation prompt excerpt}
\emph{``Given the persona and one source preference, write one natural memory
case using the required relation type and subtype.''}

\emph{``Persona: \ldots Topic preference: \ldots Source preference:
\ldots''}

\emph{``Relation type definitions:''}
\begin{itemize}
\item \texttt{complementary}: Multiple semantic variants are related to the same
question or topic and do not conflict, but they may differ in whether they
determine the answer.
\item \texttt{nuanced}: Multiple semantic variants are relevant to the same question
and can all be true, but only under different qualifying conditions such as
time, role, location, task, scope, version, definition, or attribute.
\item \texttt{contradictory}: Multiple semantic variants cannot all be true under
the same interpretation and conditions, and cannot be reconciled by adding
qualifiers.
\end{itemize}

\emph{``Boundary rules: \ldots''}

\emph{``Required relation:
\{\ldots, \texttt{relation\_type\_meaning}: \ldots,
\texttt{relation\_subtype\_meaning}: \ldots\}.''}

\emph{``Requirements:''}
\begin{itemize}
\item Use exactly this relation type and subtype for the case.
\item The case must still be natural, faithful to the persona/source
preference, and correctly labeled.
\item Label \textit{Temporal} or \textit{Contextual} only as nuanced.
\item If multiple facts simply support the same stable answer, choose
complementary \textit{Multi-evidence} instead of nuanced
\textit{Contextual}.
\item For \textit{Multi-evidence} cases, write facts that a plausible single
user task must combine. Each fact should add a different required constraint,
criterion, audience, setting, or scope; avoid facts that are just repeated
examples of the same preference.
\item For \textit{Contextual} cases, write facts so the same practical user task
would require different concrete actions, priorities, or exclusions under each
context. Avoid cases where both facts merely show the same general trait.
\item \emph{[Condensed from original prompt:]} Write one concise case
description and precise one-sentence facts; keep the case faithful to the
persona/source preference and do not mention labels, task framing, or this
instruction. \ldots
\end{itemize}

\emph{``[Condensed from original prompt:] JSON output schema omitted; the
original output returns relation labels, relation-choice reason, case
description, and facts. \ldots''}
\end{subtlememorypromptbox}
\caption{User-related variant-generation prompt for converting one persona
preference into a target-conditioned semantic variant set.}
\label{fig:appendix-case-generation-prompt}
\end{figure*}

\begin{figure*}[!tbp]
\begin{subtlememorypromptbox}{User-related variant-filter prompt excerpt}
\emph{``Review this user semantic variant set for data quality.''}

\emph{``Persona: \ldots''}

\emph{``Relation type definitions: \ldots''}

\emph{``Boundary rules: \ldots''}

\emph{``Case: \ldots''}

\begin{itemize}
\item Accept only if the case is natural, faithful to the persona/source
preference, correctly labeled, non-duplicative, and internally valid.
\item Reject if it is unnatural, mislabeled, factually unsupported, duplicated,
conflicting outside the contradictory subtype, or impossible to resolve under
its stated relation.
\item Do not reject solely because the case shares a broad or background
persona fact with another case when the answer-driving facts, user-facing
intent, and relation behavior are otherwise distinct.
\item Reject \textit{Temporal} or \textit{Contextual} cases if the semantic variants
are only additive evidence for one stable answer and no time/context condition changes the
answer.
\item Reject \textit{Multi-evidence} cases if any required fact can be dropped
without changing the concrete answer, or if the facts merely repeat the same
broad preference in nearby contexts.
\item Reject \textit{Contextual} cases if the listed contexts are different on
the surface but do not produce meaningfully different user-facing answers,
actions, or constraints.
\item Reject \textit{Contextual} cases if both facts mainly express the same
broad preference/value and do not force different concrete actions, priorities,
or exclusions for a plausible user task.
\item Reject \textit{Temporal} cases if the time boundary is not expressed in
the case facts themselves.
\item For contradictory-relation cases, reject if the semantic variants can all be true
together, such as following both African and non-African teams.
\end{itemize}

\emph{``Think step by step.''}

\emph{``\#\#\#Output''}

\emph{``Return only JSON: \{\texttt{final\_decision}: \ldots,
\texttt{accepted}: \ldots, \texttt{reason}: \ldots,
\texttt{reject\_categories}: [\ldots]\}.''}
\end{subtlememorypromptbox}
\caption{User-related variant-filter prompt for checking factual support,
compatibility-relation fidelity, and boundary conditions.}
\label{fig:appendix-case-filter-prompt}
\end{figure*}

\begin{figure*}[!p]
\begin{subtlememorycasebox}{User-related semantic variant sets}
\small
\noindent
\begin{minipage}[t]{0.31\linewidth}
\raggedright\sloppy
\textbf{Complementary, Multi-evidence.}

\emph{Case.} Amara would be especially drawn to a community program that mixes
live drumming with hands-on science learning.

\emph{Semantic variants.}
\begin{itemize}
\item[1.] Amara enjoys drumming rhythms in community events.
\item[2.] She designs interactive STEM workshops for children and teens.
\item[3.] She turns physics concepts into playful hands-on activities.
\end{itemize}
\end{minipage}
\hfill
\begin{minipage}[t]{0.31\linewidth}
\raggedright\sloppy
\textbf{Nuanced, Contextual.}

\emph{Case.} Amara uses different design styles for home spaces and children's
science activities.

\emph{Semantic variants.}
\begin{itemize}
\item[1.] At home, Amara prefers minimalist Scandinavian design with clean
lines and light wood tones.
\item[2.] For children's STEM workshops, she uses bold colors, visual cues, and
playful layouts.
\end{itemize}
\end{minipage}
\hfill
\begin{minipage}[t]{0.31\linewidth}
\raggedright\sloppy
\textbf{Contradictory.}

\emph{Case.} A memory conflict appears around Amara's sustainability habits
during grocery shopping.

\emph{Semantic variants.}
\begin{itemize}
\item[1.] Amara prefers reusable eco-friendly shopping bags for grocery trips.
\item[2.] Amara says she never brings reusable bags and always uses disposable
shopping bags.
\end{itemize}
\end{minipage}
\end{subtlememorycasebox}
\caption{Real generated user-related semantic variant sets from SubtleMemory.
Each block shows the variants that form one target-conditioned set and define
its compatibility relation.}
\label{fig:appendix-user-related-case-examples}
\end{figure*}

\paragraph{Non-user semantic variant creation.}
\label{sec:appendix-data-stage2-user-unrelated}
For non-user variants, relation-specific fact selection follows the source
family:
\begin{itemize}
\item FanOutQA and MuSiQue records become \textit{Multi-evidence} sets whose
target requires combining selected facts.
\item QACC records become \textit{Any-one} sets because repeated surface forms
independently support the same answer.
\item AmbigQA context records become \textit{Contextual} sets, and HoH dated
records become \textit{Temporal} sets.
\item AmbigQA three-answer records become contradictory-relation sets by
selecting two conflicting QA entries and removing qualifiers that would
reconcile them.
\end{itemize}
Figures~\ref{fig:appendix-complementary-fact-selection-prompt},
\ref{fig:appendix-contradictory-fact-selection-prompt}, and
\ref{fig:appendix-nuanced-fact-selection-prompt} show the relation-specific
fact-selection prompts, and Figure~\ref{fig:appendix-user-unrelated-case-examples}
shows accepted non-user variant sets.

\begin{figure*}[!tbp]
\begin{subtlememorycasebox}{Non-user semantic variant sets}
\small
\noindent
\begin{minipage}[t]{0.31\linewidth}
\raggedright\sloppy
\textbf{Complementary, Multi-evidence.}

\emph{Case.} The answer requires matching several Korean metropolitan cities
with their symbol flowers.

\emph{Semantic variants.}
\begin{itemize}
\item[1.] Busan's symbol flower is Camellia.
\item[2.] Incheon's symbol flower is Rose.
\item[3.] Ulsan's symbol flower is Pear flower.
\end{itemize}
\end{minipage}
\hfill
\begin{minipage}[t]{0.31\linewidth}
\raggedright\sloppy
\textbf{Nuanced, Temporal.}

\emph{Case.} The total trophy count for SD Crvena zvezda's clubs changes over
time.

\emph{Semantic variants.}
\begin{itemize}
\item[1.] As of 2024-06-01, the clubs had won 854 trophies.
\item[2.] As of 2024-08-01, the clubs had won 858 trophies.
\item[3.] As of 2024-12-01, the clubs had won 870 trophies.
\end{itemize}
\end{minipage}
\hfill
\begin{minipage}[t]{0.31\linewidth}
\raggedright\sloppy
\textbf{Contradictory.}

\emph{Case.} The same surface question about \textit{The Fault in Our Stars}
can point to different answer scopes.

\emph{Semantic variants.}
\begin{itemize}
\item[1.] Described by intended age group, \textit{The Fault in Our Stars} is
young adult.
\item[2.] Described by general content, it is realistic fiction.
\end{itemize}
\end{minipage}
\end{subtlememorycasebox}
\caption{Real generated non-user semantic variant sets from SubtleMemory. Each
block shows the external-knowledge variants that define the compatibility
relation.}
\label{fig:appendix-user-unrelated-case-examples}
\end{figure*}

\begin{figure*}[!tbp]
\begin{subtlememorypromptbox}{Non-user complementary fact-selection prompt excerpt}
\emph{``Rewrite the complementary source data below into a cleaner
benchmark-ready complementary fact bundle.''}

\emph{``Goal:''}
\begin{itemize}
\item Produce one benchmark-ready complementary question.
\item Produce one canonical answer for that question.
\item Select a manageable fact subset for multi-session generation.
\end{itemize}

\emph{``Hard rules:''}
\begin{itemize}
\item Keep the result complementary: related facts may distract, support, or
redundantly restate the answer, but they must not create contradictory or
context-resolved multi-answer behavior.
\item Keep the selected fact set manageable: choose between 2 and 6 selected
facts total.
\item Each selected fact must reference one source memory\_id.
\item The selected fact roles must obey the subtype definition below.
\item The canonical answer should be natural text, not a debug note.
\end{itemize}

\emph{``Subtype definition: \ldots''}

\emph{``Source question: \ldots''}

\emph{``Source answer for hidden grounding: \ldots''}

\emph{``Available source evidence items: \ldots''}

\emph{``Return JSON in this format: \ldots''}

\emph{``Think step by step.''}
\end{subtlememorypromptbox}
\caption{Non-user complementary fact-selection prompt for converting
multi-evidence source records into benchmark-ready variant sets.}
\label{fig:appendix-complementary-fact-selection-prompt}
\end{figure*}

\begin{figure*}[!tbp]
\begin{subtlememorypromptbox}{Non-user contradictory fact-selection prompt excerpt}
\emph{``Select two source QA entries and rewrite them into one contradictory
setup.''}

\emph{``Goal:''}
\begin{itemize}
\item Create a single canonical conflict question that both selected answers
appear to answer under the same apparent condition.
\item The result must be contradictory memory, not nuanced memory.
\item Remove source qualifiers that would reconcile the answers, such as
explicit year, version, edition, role variant, location, scope, or scenario
markers.
\item The canonical conflict question is an internal benchmark scaffold for
later dialogue and QA generation. It may later be reused directly or lightly
paraphrased.
\end{itemize}

\emph{``Hard rules:''}
\begin{itemize}
\item Select exactly two different source QA entries.
\item Use only answer content grounded in the selected source entries. Do not
invent a new answer.
\item The canonical conflict question must not quietly preserve a qualifier
that makes both answers true.
\item Do not output a question like ``as of 2016 vs 2017'', ``in the 1967 or
2016 version'', ``for the TV series'', or anything else that turns the setup
back into temporal/contextual nuance.
\item Good outcome: two mutually incompatible answers to one plain question.
\item Bad outcome: two answers that are still separated by explicit
time/version/context qualifiers.
\end{itemize}

\emph{``Source question: \ldots''}

\emph{``Available source QA pairs: \ldots''}

\emph{``Return JSON in this format: \ldots''}

\emph{``Think step by step.''}
\end{subtlememorypromptbox}
\caption{Non-user contradictory fact-selection prompt for selecting conflicting
QA entries and removing resolving qualifiers.}
\label{fig:appendix-contradictory-fact-selection-prompt}
\end{figure*}

\begin{figure*}[!p]
\begin{subtlememorypromptbox}{Non-user nuanced fact-selection prompt excerpts}
\emph{``Rewrite the context source data below into a cleaner context fact
bundle.''}

\emph{``Goal:''}
\begin{itemize}
\item Produce one brief ambiguous context question for the topic.
\item Keep the result clearly contextual: the different answers should all
remain true, but under different contexts.
\item \emph{[Condensed from original prompt:]} Preserve every source fact;
rewrite each decisive context cleanly and pair it with the correct answer.
\ldots
\end{itemize}

\emph{``Hard rules:''}
\begin{itemize}
\item Keep every source evidence item exactly once. Do not drop, merge, or invent
facts.
\item Preserve the decisive context for each fact, such as role, location,
task, scope, version, definition, unit, or attribute.
\item \emph{[Condensed from original prompt:]} Do not collapse facts into a
summary fact; keep the broad question ambiguous; make each
\texttt{context\_anchor} short and concrete; write compact fact statements that
bind the answer to its context. \ldots
\end{itemize}

\emph{``Rewrite the HoH temporal source data below into a cleaner temporal fact
bundle.''}

\emph{``Goal:''}
\begin{itemize}
\item Produce one brief umbrella question for the topic. It should be the same
underlying question across all snapshots.
\item Keep the result clearly temporal: the answer to the same question changes
across dated snapshots.
\item \emph{[Condensed from original prompt:]} Preserve every dated snapshot
exactly once; keep the snapshot date and answer paired for each source fact.
\ldots
\end{itemize}

\emph{``Hard rules:''}
\begin{itemize}
\item Keep every source evidence item exactly once. Do not drop, merge, or invent
facts.
\item The time condition must include the exact source snapshot date, e.g. ``as
of 2024-07-01''.
\item \emph{[Condensed from original prompt:]} Do not turn the snapshot date
into the event date; do not collapse snapshots into one summary fact; keep the
umbrella question short, natural, and without a time condition. \ldots
\end{itemize}
\end{subtlememorypromptbox}
\caption{Non-user nuanced fact-selection prompts for converting contextual and
temporal source records into benchmark-ready variant sets.}
\label{fig:appendix-nuanced-fact-selection-prompt}
\end{figure*}

\subsection{Stage 3: Session Construction}
\label{sec:appendix-data-stage-session-construction}

\paragraph{User-related session construction.}
\label{sec:appendix-data-stage3-user-related}
Each generated user-related session may express only its assigned variant, while
other same-set variants are provided only as relation context. The assigned
variant is hidden supervision rather than a memory sentence, so the conversation
must make the target recoverable through concrete signals such as polarity,
behavior, boundary, example, routine, correction, artifact detail, or use
context. The opening user turn cannot carry all target evidence, and task
diversity is sampled over conversation type, workflow rhythm, message count,
persona-signal level, and same-set session history
(Table~\ref{tab:appendix-session-task-styles}). The session filter rejects
unnatural, mechanical, hidden, over-explicit, type-mismatched, label-revealing,
or relation-breaking sessions.
Figures~\ref{fig:appendix-session-generation-prompt} and
\ref{fig:appendix-session-filter-prompt} show the prompts, and
Figure~\ref{fig:appendix-session-example} shows an accepted session set.

\begin{table*}[!p]
\centering
\scriptsize
\setlength{\tabcolsep}{4pt}
\renewcommand{\arraystretch}{1.12}
\begin{tabular}{@{}p{0.20\linewidth}p{0.27\linewidth}p{0.45\linewidth}@{}}
\toprule
\textbf{Conversation type} & \textbf{Primary agent task} & \textbf{Sampled workflow options} \\
\midrule
\texttt{decision\_support} &
Compare options, weigh tradeoffs, clarify criteria, and decide. &
\textit{1.} Start with a practical choice, then let criteria, tradeoffs, and a
final boundary emerge. \newline
\textit{2.} Move from a rough dilemma to narrowed options, adding one
constraint at a time. \newline
\textit{3.} Let the assistant compare options, the user correct weighting, and
the ending settle on a next action. \\
\texttt{planning\_coordination} &
Organize steps, timing, dependencies, preparation, responsibilities, or
contingencies. &
\textit{1.} Start with an upcoming task, then build sequence, dependencies, and
timing. \newline
\textit{2.} Let the assistant sketch a plan, the user reveal constraints, and
the final turn tighten it. \newline
\textit{3.} Move from scattered preparation to an order of operations with
responsibilities or contingencies clarified late. \\
\texttt{troubleshooting} &
Diagnose a practical problem and work toward next steps. &
\textit{1.} Start with a symptom or blocked task, then narrow likely causes
through observations. \newline
\textit{2.} Let the assistant propose a check, the user report what is true,
and the session end with a fix path. \newline
\textit{3.} Move from confusion to diagnosis, with each user turn adding one
clue, failed attempt, or condition. \\
\texttt{learning\_explanation} &
Explain a concept, mechanism, background, example, or applied understanding. &
\textit{1.} Start with a concept or situation, then refine the explanation
through examples and limits. \newline
\textit{2.} Let the assistant explain simply, the user ask for an applied
version, and the final answer connect it to use. \newline
\textit{3.} Move from background explanation to a concrete example, with the
user specifying what kind of understanding helps. \\
\texttt{resource\_selection} &
Choose among tools, places, courses, materials, activities, routes, services,
or other resources. &
\textit{1.} Start with a need for options, then reveal constraints, criteria,
and reasons to rule choices in or out. \newline
\textit{2.} Let the assistant suggest categories, the user narrow the context,
and the ending recommend a short ranked set. \newline
\textit{3.} Move from broad resource search to best fit by adding budget,
setting, time, skill, access, or taste constraints. \\
\texttt{workflow\_setup} &
Build a repeatable routine, checklist, tracking approach, template, or
operating pattern. &
\textit{1.} Start with a recurring friction point, then turn it into a routine,
checklist, or tracking pattern. \newline
\textit{2.} Let the assistant propose a structure, the user identify what
would break it, and the final answer simplify it. \newline
\textit{3.} Move from a messy process to a usable template, with each user turn
adding one real-world constraint. \\
\texttt{information\_organization} &
Sort notes, options, evidence, constraints, ideas, or action items into a
clearer structure. &
\textit{1.} Start with scattered notes, options, or evidence, then sort them
into categories, priorities, or action items. \newline
\textit{2.} Let the assistant impose an initial structure, the user add missing
context, and the final answer reorganize it. \newline
\textit{3.} Move from an unstructured list to a clearer map, with the user
clarifying what distinction matters most. \\
\texttt{personal\_reflection} &
Reason about personal patterns, boundaries, preferences, energy, values, or
decision principles. &
\textit{1.} Start with a noticed pattern, then use concrete examples to name a
boundary or decision principle. \newline
\textit{2.} Let the assistant offer an interpretation, the user correct part of
it, and the session land on a usable framing. \newline
\textit{3.} Move from a vague feeling to a grounded self-read, without turning
the session into generic advice. \\
\texttt{artifact\_production} &
Produce a new text or content artifact, such as an email, message, post,
paragraph, or script. &
\textit{1.} Start with a simple artifact request, then reveal tone, audience,
and constraints before the final text. \newline
\textit{2.} Let the assistant draft first, the user add practical constraints,
and the final version become more specific. \newline
\textit{3.} Move from purpose to usable artifact, with revisions focused rather
than repeatedly reworking the same text. \\
\texttt{artifact\_review\_}\newline\texttt{or\_localization} &
Revise, shorten, translate, localize, or change the tone of an existing
artifact. &
\textit{1.} Start from an existing artifact, then revise purpose, audience,
tone, or language through targeted changes. \newline
\textit{2.} Let the assistant diagnose what is not working, the user clarify
the desired effect, and the final version adjust. \newline
\textit{3.} Move from review to localized or tightened output, with the user
adding context gradually. \\
\bottomrule
\end{tabular}
\caption{Conversation types and sampled workflow options used for session
diversity. Each session samples four candidate conversation types, and each
candidate type is paired with one sampled loose workflow rhythm rather than a
fixed template.}
\label{tab:appendix-session-task-styles}
\end{table*}

\begin{figure*}[!p]
\begin{subtlememorypromptbox}{User-related session-generation prompt excerpt}
\emph{``Generate one realistic user-assistant session for this user.''}

\emph{``Persona: \ldots Fact to embed naturally: \ldots Case context:
\ldots Candidate conversation types: \ldots Current case relation guidance:
\ldots Preferred conversation type: \ldots Fallback conversation types:
\ldots Existing sessions for this same case: \ldots Persona signal level:
\ldots''}

\begin{itemize}
\item Choose exactly one candidate type and write the session in that style.
\item Treat the selected conversation type as the primary agent task, not
merely as an output format.
\item If a candidate type includes a \texttt{suggested\_flow}, use it as a
loose rhythm for the selected type, not a rigid template; if the preferred type
would feel forced, choose the most natural fallback type.
\item Use existing sessions only as same-case relation and style context. Vary
only expression style, such as conversation type, opening shape, request task,
assistant output format, and rhythm. Do not change relation-critical context.
\item The target fact is hidden supervision. Do not state it directly, but make
it recoverable from the user's side of the full session.
\item Before writing, identify the target object/category and any
qualifier/context. User turns should provide that object/category plus at least
one concrete relation signal, such as polarity, behavior, boundary, example,
routine, correction, artifact detail, or use context.
\item Only embed the target fact for this session. Other facts in the same case
are relation context only. Use them to avoid breaking the case design, but do
not express, imply, paraphrase, combine, or leak them in this session.
\item \emph{[Condensed from original prompt:]} Hidden does not mean cryptic:
the fact should be recoverable, but the session should not copy or closely
paraphrase the fact, turn it into a direct self-profile sentence, front-load all
background into the opening user message, or leak other same-case facts.
\ldots
\item Apply \texttt{persona\_signal\_level} only to the user's broader voice;
assistant messages should respond naturally and should not restate the hidden
fact as a conclusion about the user.
\item Use OpenAI message format with exactly the requested number of turns and
messages.
\item Messages must start with the user and alternate user, assistant, user,
assistant until the session ends.
\end{itemize}

\emph{``Think step by step.''}

\emph{``\#\#\#Output''}

\emph{``Return only JSON: \{\texttt{selected\_conversation\_type}: \ldots,
\texttt{messages}: [\ldots]\}.''}
\end{subtlememorypromptbox}
\caption{User-related session-generation prompt for embedding one semantic
variant into an implicit task-oriented dialogue.}
\label{fig:appendix-session-generation-prompt}
\end{figure*}

\begin{figure*}[!p]
\begin{subtlememorypromptbox}{User-related session-filter prompt excerpt}
\emph{``Review a batch of generated user-assistant conversation cases.''}

\emph{``Relation type definitions: \ldots''}

\emph{``Relation subtype definitions: \ldots''}

\emph{``Subtype notes: \ldots''}

\emph{``Boundary rules: \ldots''}

\emph{``Batch: \ldots''}

\begin{itemize}
\item Find only cases that should be rejected.
\item For each conversation, first inspect user turns only and internally
extract the target object/category, any qualifier/context, and the recoverable
signals that make the target fact inferable.
\item Reject as \texttt{too\_hidden} if user turns reveal only a broad
topic/category, or if they lack the target object/category plus at least one
concrete relation signal.
\item Treat hiddenness as acceptable when the conversation does not state the
fact directly, but the target object/category and the user's polarity,
behavior, boundary, or practical use are recoverable from user turns.
\item Reject as \texttt{too\_explicit} if the conversation directly repeats the
fact, closely paraphrases it, or turns it into a profile sentence.
\item Reject a case if any conversation is unnatural, too mechanical, not a
realistic multi-turn session, does not start with the user, does not alternate
user/assistant, does not naturally and implicitly express its target fact,
expresses the target fact only as a vague topic-adjacent aside that a reader
could not recover from the full session, copies or closely paraphrases the fact,
states the fact as a direct self-profile sentence, mismatches the selected
conversation type, conflicts with the persona/case/fact, or exposes benchmark,
memory, label, or task framing.
\item For non-artifact conversation types, reject conversations whose main task
is actually drafting, rewriting, translating, caption writing, or repeated text
revision.
\item Reject a case if its conversations no longer satisfy the current relation
type/subtype design, even if each single conversation looks locally fluent.
\item Reject a non-\textit{Temporal} case if it becomes valid only by treating
a later conversation as the current truth or an earlier conversation as
superseded; reject a \textit{Temporal} case if its time split is only implied by
artifact/session order rather than content.
\item For every rejected case, list the specific problem conversations and
their problems when identifiable.
\item If there are no problem cases in this batch, return an empty
\texttt{problem\_cases} list.
\end{itemize}

\emph{``Think step by step.''}

\emph{``\#\#\#Output''}

\emph{``Return only JSON: \{\texttt{problem\_cases}: [\ldots]\}.''}
\end{subtlememorypromptbox}
\caption{User-related session-filter prompt for checking naturalness, variant
recoverability, and compatibility-relation preservation.}
\label{fig:appendix-session-filter-prompt}
\end{figure*}

\begin{figure*}[!p]
\begin{subtlememorycasebox}{User-related session excerpts for a Multi-evidence variant set}
\small
\textbf{Case.} Amara would be especially drawn to a community program that
mixes live drumming with hands-on science learning.

\textbf{Target facts.} Amara enjoys drumming rhythms in community events, she
designs interactive STEM workshops for children and teens, and she turns physics
concepts into playful hands-on activities.

\noindent
\begin{minipage}[t]{0.31\linewidth}
\raggedright\sloppy
\textbf{Session 1: \texttt{workflow\_setup}.}

\emph{User.} I need a repeatable prep routine for community
events. Something simple I can do the night before and again right before
things start.

\emph{Assistant.} Let's make it short enough to reuse. I would split it into
night-before packing, arrival setup, and a one-minute reset before the first
group joins.

\emph{User.} Mostly I am trying not to forget my small drum, water,
and the first few call-and-response cues before people gather.

\ldots

\emph{User.} My hands feel stiff at first, and once the crowd answers back the
tempo can speed up before I notice.

\emph{Assistant.} The checklist should keep the drum, cue order, warmup, and
tempo anchor together. \ldots
\end{minipage}
\hfill
\begin{minipage}[t]{0.31\linewidth}
\raggedright\sloppy
\textbf{Session 2: \texttt{troubleshooting}.}

\emph{User.} One of my outreach activities keeps falling apart
halfway through, and I am trying to figure out what the actual problem is.

\emph{Assistant.} Tell me what the activity is, who is in the room, and where
the breakdown usually starts.

\emph{User.} It is a ramp-and-toy-car lesson on speed and
angle. I usually have both younger kids and teens together.

\ldots

\emph{User.} I need it reusable for our STEM programs, including school and
museum rooms where groups move at different speeds.

\emph{Assistant.} Split the activity into a quick shared demo, a
younger-kid lane, and a teen extension challenge. \ldots
\end{minipage}
\hfill
\begin{minipage}[t]{0.31\linewidth}
\raggedright\sloppy
\textbf{Session 3: \texttt{resource\_selection}.}

\emph{User.} I am choosing one physics activity for a community
family pop-up, and I only have room for a single setup.

\emph{Assistant.} The best choice depends on whether you want the visitors to
observe a demonstration or change variables themselves.

\emph{User.} I want that little sequence of predict, try,
argue with reality, and try again.

\ldots

\emph{User.} Bonus points if it feels like a game and not a
school lesson, works on a modest budget, and lets participants touch the
system themselves.

\emph{Assistant.} Choose toy-car ramp trials because they make changing
motion visible through repeated hands-on tests. \ldots
\end{minipage}
\end{subtlememorycasebox}
\caption{Real user-related session excerpts from a complementary
Multi-evidence variant set. The compatible facts are distributed across
ordinary task-oriented interactions, and a later community-program query
requires recovering all three sessions rather than using any single session
alone. Ellipses indicate omitted turns from the original sessions.}
\label{fig:appendix-session-example}
\end{figure*}

\paragraph{Non-user session construction.}
\label{sec:appendix-data-stage3-user-unrelated}
A planning step assigns selected external facts to independent sessions, chooses
a conversation type and workflow rhythm for each session, and keeps facts
assigned to other sessions unmentioned. Persona-signal levels are not sampled,
because these facts are external knowledge rather than persona-specific
preferences or states. \textit{Multi-evidence} cases keep partial evidence
separate until query time, \textit{Any-one} cases keep equivalent facts
redundant but natural, \textit{Contextual} and \textit{Temporal} cases preserve
the decisive qualifier, and contradictory-relation cases embed one target claim
per session without announcing or resolving the conflict.
Figure~\ref{fig:appendix-user-unrelated-session-generation-prompt} shows the
planning and generation prompts, and
Figure~\ref{fig:appendix-user-unrelated-session-example} shows real non-user
session excerpts.

\begin{figure*}[!tbp]
\begin{subtlememorypromptbox}{Non-user session-planning and session-generation prompt excerpts}
\textbf{Session planning.}
\emph{``Plan independent user-assistant sessions for a
[complementary-relation / nuanced-relation / contradictory-relation] memory benchmark.''}
The prompt asks the model to produce the required number of session plans,
assign selected semantic facts according to the relation-specific rule, make the
sessions differ in scenario and concrete event, choose one conversation-type
candidate and its paired flow for each session, and keep facts assigned to other
sessions unmentioned. \emph{[Condensed from relation-specific prompt variants:]}
\textit{Multi-evidence} cases distribute required facts for cross-session
integration; \textit{Any-one} cases keep equivalent facts redundant but natural;
\textit{Contextual} and \textit{Temporal} cases preserve the decisive context or
snapshot condition; contradictory-relation cases embed exactly one target claim per
session while staying silent about contradiction. \ldots

\smallskip
\textbf{Session generation.}
\emph{``Generate one standalone user-assistant session for a
[complementary-relation / nuanced-relation / contradictory-relation] benchmark.''}
The prompt asks for a realistic everyday dialogue that follows the approved
plan, uses the chosen conversation type and flow as the interaction rhythm,
embeds only the assigned facts, keeps the event distinct from other hidden
sessions, avoids asking the benchmark question verbatim, and avoids clean
inventories or one-shot recaps. \emph{[Condensed from relation-specific prompt
variants:]} Partial \textit{Multi-evidence} sessions must not state the full
answer; \textit{Contextual} sessions keep qualifiers explicit enough to matter;
\textit{Temporal} sessions treat dates as answer-snapshot conditions; and
contradictory sessions do not mention disagreement, mixed memories, or other
sessions. \ldots

\smallskip
\textbf{Output schema.}
\emph{``Return JSON in this format:''}
\texttt{\{session\_id, chosen\_scenario, conversation, coverage\_notes\}}.
\end{subtlememorypromptbox}
\caption{Non-user session-planning and generation prompts for distributing
selected external-knowledge variants across natural dialogues.}
\label{fig:appendix-user-unrelated-session-generation-prompt}
\end{figure*}

\begin{figure*}[!tbp]
\begin{subtlememorycasebox}{Non-user session excerpts for a Multi-evidence variant set}
\small
\textbf{Case.} The answer requires matching several Korean metropolitan cities
with their symbol flowers.

\textbf{Target facts.} Busan's symbol flower is Camellia; Ulsan's symbol flower
is Pear flower; Incheon's symbol flower is Rose.

\begin{minipage}[t]{0.48\linewidth}
\raggedright\sloppy
\textbf{Session 1: \texttt{learning\_explanation}.}

\emph{User.} I'm helping my younger cousin study Korean city symbols. Can you
explain in simple terms what a city's symbol flower is, and how to make the
idea easy to remember?

\emph{Assistant.} A city's symbol flower is an official flower the city uses to
represent itself, a bit like a visual emblem. The easiest way to remember it
is: not ``favorite flower,'' but ``flower that stands for the city.''

\emph{User.} That helps, but I learn better with examples. Could you use one
metropolitan city so I can hear how you'd explain it out loud?

\emph{Assistant.} If I were explaining it with Busan, I'd say that Busan's
symbol flower is Camellia. Then I'd add that this is one example of how a city
chooses a flower to act as part of its civic identity.

\emph{User.} Nice. Could you give me a second city too? I think comparing two
examples would make the concept stick better.

\emph{Assistant.} You could pair Busan with Ulsan and point out that Ulsan's
symbol flower is Pear flower. That shows the main idea: each city has its own
emblematic flower.

\emph{User.} I like that. When I explain it, should I focus more on the flower
name or on what the symbol is doing?

\emph{Assistant.} Focus first on what the symbol is doing. With Busan, don't
stop at ``Camellia'' by itself; say that Busan uses Camellia as its symbol
flower.

\ldots

\emph{User.} One last thing: can you give me a memory trick for the Ulsan
example?

\emph{Assistant.} Link the word ``pear'' with the image of a light spring
blossom. If you picture Ulsan paired with a pear blossom image, it becomes
easier to recall that Ulsan's symbol flower is Pear flower.
\end{minipage}
\hfill
\begin{minipage}[t]{0.48\linewidth}
\raggedright\sloppy
\textbf{Session 2: \texttt{artifact\_production}.}

\emph{User.} I'm putting together a small cultural display and need a short
caption for an Incheon profile card.

\emph{Assistant.} Here's a clean starting option: ``Incheon, a dynamic gateway
city, blends modern energy with local pride and tradition.''

\emph{User.} Nice start. I want it to feel a little more specific though, like
something that belongs on a city handout card.

\emph{Assistant.} ``Incheon is a vibrant city known for its forward-looking
spirit; its symbol flower is the rose, reflecting beauty and civic pride.''

\emph{User.} That's close, but it sounds a bit formal. Can you make it warmer
and slightly shorter?

\emph{Assistant.} ``Incheon is a lively city whose symbol flower, the rose,
represents its beauty and pride.''

\ldots

\emph{User.} I like that. Could you give me two final options: one plain and
one a little more polished? Keep both short.

\emph{Assistant.} Plain: ``Incheon's symbol flower is the rose.'' Polished:
``The rose, Incheon's symbol flower, reflects the city's beauty and pride.''

\emph{User.} Great, I'll use the polished one. Can you give it to me one last
time with no extra commentary so I can paste it into the layout?

\emph{Assistant.} The rose, Incheon's symbol flower, reflects the city's beauty
and pride.
\end{minipage}
\end{subtlememorycasebox}
\caption{Real non-user session excerpts from a complementary Multi-evidence
variant set. The required facts are distributed across ordinary task-oriented
interactions rather than exposed as a benchmark list.}
\label{fig:appendix-user-unrelated-session-example}
\end{figure*}

\subsection{Stage 4: Evaluation Instance Construction}
\label{sec:appendix-data-stage-evaluation-instance-construction}

\paragraph{User-related evaluation instance construction.}
\label{sec:appendix-data-stage4-user-related}
User-related queries use either \texttt{structured\_form}, where fixed fields
make values and clarification behavior judgeable, or
\texttt{resource\_arrangement}, where candidate resources are provided and the
agent selects, ranks, excludes, assigns, or finalizes them. Each $q_\tau$ must
be self-contained, realistic, and underdetermined without the
target-conditioned variants. Any time, context, role, location, scope, or object
condition must be neutral rather than answer-revealing. Correct answers $A^+$
encode required fact alignment, condition, candidate selection, field value,
exclusion, or clarification, while incorrect answers $A^-$ instantiate
relation-specific failures.
The instance filter removes broad, leaky, unsupported, relation-inconsistent,
or unclearly judgeable instances. Figures~\ref{fig:appendix-query-generation-prompt},
\ref{fig:appendix-answer-generation-prompt}, and
\ref{fig:appendix-instance-filter-prompt} show the prompts, and
Figure~\ref{fig:appendix-instance-example} shows both task forms.

\begin{figure*}[!p]
\begin{subtlememorypromptbox}{User-related query-generation prompt excerpt}
\emph{``Generate candidate user queries for this user semantic variant set.''}

\emph{``Persona: \ldots Case: \ldots Case relation: \ldots Accepted sessions
for this case: \ldots Same-topic sibling cases for interference avoidance:
\ldots Relation-specific task guidance: \ldots''}

\begin{itemize}
\item Generate exactly the requested number of natural first-person user task
requests.
\item Write each request from the user's point of view as a natural
user-to-agent help request, not as a direct question about the user's own
preference, habit, usual choice, or best fit.
\item Each request must be self-contained, realistic, have one primary
decision/output target, and match the current case relation definition.
\item A correct response must require the current case facts and relation;
avoid broad or generic tasks that can be completed well without this case.
\item Prefer task shapes whose correctness can be checked through concrete
selection, exclusion, ranking, checklist items, clarification, or fact-based
criteria.
\item Include only neutral time/context/role/location/task/scope/version/object
conditions needed by the relation; leave the selected preference, habit,
conflict, or state unstated.
\item \emph{[Condensed from original prompt:]} Use accepted sessions only as
background signal; do not refer to prior chats, hidden facts, ground-truth
preferences, memory, labels, benchmark framing, answer direction, or details
that appear only in sessions. \ldots
\end{itemize}

\emph{``Task form menu:''}
\emph{``[Condensed from original prompt:] Choose exactly one
\texttt{task\_form}. Each query must include a visible user goal,
agent-available material, and a concrete output contract.
\texttt{structured\_form} uses fixed fields or slots that make the answer
judgeable; \texttt{resource\_arrangement} uses visible resources and asks for
selection, ranking, exclusion, assignment, scheduling, cart building, or a
similar concrete operation. Use realistic named resources, matched distractors,
and enough visible ambiguity that memory is needed. \ldots''}

\emph{``Hint and interference control:''}
\emph{``[Condensed from original prompt:] The query must be underdetermined
without the current case facts. Visible goals, fields, resources, and neutral
context may make the task natural, but they must not reveal the correct answer.
Avoid option-specific hints, hidden-criterion names, visible answer direction,
and same-topic sibling interference; run a visible-only check and rewrite if
the query alone would make the answer obvious. \ldots''}

\emph{``[Condensed from original prompt:] Task quality calibration examples
and JSON output schema are omitted; the original output stores each generated
query in the \texttt{question} field together with its \texttt{task\_form}.
\ldots''}
\end{subtlememorypromptbox}
\caption{User-related query-generation prompt for producing target queries
$q_\tau$ in assistant-task form.}
\label{fig:appendix-query-generation-prompt}
\end{figure*}

\begin{figure*}[!tbp]
\begin{subtlememorypromptbox}{User-related answer-generation prompt excerpt}
\emph{``Generate short fact-grounded answer candidates for these user
queries.''}

\emph{``Persona: \ldots''}

\emph{``Case: \ldots''}

\emph{``Case relation: \ldots''}

\emph{``Accepted sessions for this case: \ldots''}

\emph{``Queries: \ldots''}

\emph{``Relation-specific answer guidance: \ldots''}

\begin{itemize}
\item For each query identifier, generate exactly the requested number of
correct answers and incorrect answers.
\item Treat each item as a concise, standalone key answer, not as the full task
output.
\item Answers must match the current case relation, respect stated time/context
conditions, and be grounded in the case facts and persona rather than unseen
prior text.
\item Correct answers should capture the essential facts, fact combinations,
conditions, directions, exclusions, or candidate decisions a good response must
reflect.
\item Use \texttt{task\_form} when present: \texttt{structured\_form} keys
identify required field values or slot content, while
\texttt{resource\_arrangement} keys identify selected or excluded resources,
ordering, final set, slot assignment, needed clarification, or
abstention/no-selection when the relation requires it.
\item Incorrect answers should be plausible but wrong: wrong facts, missing
facts, wrong condition/direction, wrong form values, wrong candidate/order/final
set, surface-clue-only choice, stereotype-driven choice, overconfident
contradiction resolution, hedged non-selection, or unsupported criterion.
\item If external named candidates appear, evaluate the in-prompt, case-grounded
candidate decision rather than trivia or outside-world facts.
\item Keep answers natural, specific, similar in length, and limited to the key
answer content.
\item Do not write meta-evaluative statements, full deliverables, unsupported
persona comparisons, or mentions of labels, memory, benchmark framing, hidden
facts, hidden preferences, or task framing.
\end{itemize}

\emph{``[Condensed from original prompt:] JSON output schema omitted; the
original output uses \texttt{question\_id} as the query identifier and returns
\texttt{correct\_answers} and \texttt{incorrect\_answers}. \ldots''}
\end{subtlememorypromptbox}
\caption{User-related answer-candidate generation prompt for producing
reference correct answers $A^+$ and plausible incorrect answers $A^-$.}
\label{fig:appendix-answer-generation-prompt}
\end{figure*}

\begin{figure*}[!p]
\begin{subtlememorypromptbox}{User-related instance-filter prompt excerpt}
\textbf{Task.}
Review generated user queries and short answer candidates.

\textbf{Input.}
\begin{itemize}
\item Batch: \ldots
\item Each case QA item includes \texttt{case\_relation\_guidance}. Use the
current relation type meaning, subtype meaning, and subtype notes when judging
relation quality.
\end{itemize}

\textbf{Reject query.}
\begin{itemize}
\item \emph{[Condensed from original prompt:]} Reject queries with unnatural or
overly broad requests, no realistic completion need, multiple deliverables,
dependency on prior sessions or drafts, direct preference queries,
benchmark/memory leakage, hidden-fact paraphrase, relation mismatch, or tasks
that mainly reward generic writing rather than fact use. \ldots
\item \emph{[Condensed from original prompt:]} Reject task-form failures such
as missing or invalid \texttt{task\_form}, missing visible goal,
agent-available material, or output contract, uncontrolled form fields, missing
resource briefs, requests to invent/search/rewrite resources, over-tuned
candidates, option-specific hints, weak distractors, visible-query
answerability, stereotype answerability, or sibling-case interference. \ldots
\end{itemize}

\textbf{Relation checks.}
\begin{itemize}
\item \emph{[Condensed from original prompt:]} For complementary-relation cases, reject
invented conditions or conflicts; \textit{Multi-evidence} answers must combine
required memories.
\item For nuanced-relation cases, reject missing or exposed conditions, wrong
conditions, global flattening, and tasks without real temporal/contextual
discrimination.
\item For contradictory-relation cases, reject one-side or invented-resolution prompts;
correct answers must acknowledge conflict or uncertainty.
\item \emph{[Condensed from original prompt:]} Rejection calibration examples
cover \texttt{missing\_or\_invalid\_task\_form},
\texttt{missing\_external\_candidate\_framing},
\texttt{over\_tuned\_candidates}, \texttt{case\_keyword\_leakage},
\texttt{option\_specific\_hint}, \texttt{answerable\_without\_memory}, and
weak \textit{Contextual}. \ldots
\end{itemize}

\textbf{Answer checks.}
\begin{itemize}
\item \emph{[Condensed from original prompt:]} Remove correct answers that are
unsupported, too generic, condition-insensitive, contradiction-resolving, or
written as full task outputs.
\item Remove incorrect answers that are actually correct, contain the target
fact, reward the right condition, or remain valid answer targets.
\item If a query has no valid correct answers or no valid incorrect answers,
reject that query.
\end{itemize}

\textbf{Output.}
\emph{[Condensed from original prompt:]} JSON schema omitted; the original
output returns rejected queries and removed answers. \ldots
\end{subtlememorypromptbox}
\caption{User-related instance-filter prompt for validating generated target
queries $q_\tau$ and answer candidates.}
\label{fig:appendix-instance-filter-prompt}
\end{figure*}

\begin{figure*}[!tbp]
\begin{subtlememorycasebox}{User-related evaluation instance excerpts for two task forms}
\small
\textbf{Case.} Amara uses different design styles depending on whether she is
shaping her home environment or creating materials and spaces for children's
science activities.

\textbf{Facts.}
\begin{itemize}
\item[1.] At home, Amara prefers minimalist, Scandinavian-inspired design with
clean lines, light wood tones, and an uncluttered feel.
\item[2.] For children's STEM workshops or toy-car race setups, she uses bold
colors, interactive visual cues, and playful layouts.
\end{itemize}

\begin{minipage}[t]{0.48\linewidth}
\raggedright\sloppy
\textbf{Task form: \texttt{structured\_form}.}

\emph{User request.} Fill a one-card apartment design brief with short phrases:
overall direction, shape/line cue, wood/finish cue, clutter boundary, and room
feeling.

\emph{Reference correct answer.} Minimalist Scandinavian; clean lines; light
wood tones; uncluttered surfaces; calm ease.

\emph{Plausible incorrect behavior.} Use the workshop style instead: bold
graphic forms, vivid finishes, layered visual cues, and energetic engagement.
\end{minipage}
\hfill
\begin{minipage}[t]{0.48\linewidth}
\raggedright\sloppy
\textbf{Task form: \texttt{resource\_arrangement}.}

\emph{User request.} Choose one visual setup package for an elementary
museum toy-car physics activity from a provided resource brief.

\emph{Reference correct answer.} Signal Sprint Kit -- it matches bold colors,
interactive cues, and small activity islands for an energetic kids setup.

\emph{Plausible incorrect behavior.} Choose Vector Grove Kit because its pale
wood and restrained palette match the home-design preference instead.
\end{minipage}
\end{subtlememorycasebox}
\caption{User-related evaluation-instance excerpts showing the two task forms
used in user-related query construction. Both examples come from the same
nuanced Contextual case: the correct response depends on applying the
appropriate remembered context rather than a superficially similar one.}
\label{fig:appendix-instance-example}
\end{figure*}

\paragraph{Non-user evaluation instance construction.}
\label{sec:appendix-data-stage4-user-unrelated}
Non-user evaluation instances add relation-specific query and answer rules:
\begin{itemize}
\item \textit{Multi-evidence} questions preserve the hidden dependency
structure without enumerating every hop.
\item \textit{Contextual} questions specify exactly one decisive context, while
\textit{Temporal} questions cover explicit, event-anchored, and relative
temporal conditions.
\item Contradictory-relation questions target the disputed point without adding
qualifiers that would resolve the conflict.
\item Each query receives three correct and three incorrect answer candidates,
with relation-specific rules for integrated complementary answers,
context-specific answers, dated temporal answers, and contradiction-aware
answers.
\item Separate filters validate session embedding, final query quality, and
answer candidates.
\end{itemize}
Figures~\ref{fig:appendix-user-unrelated-query-generation-prompt} and
\ref{fig:appendix-user-unrelated-answer-generation-prompt} show the query and
answer-generation prompts, Figures~\ref{fig:appendix-user-unrelated-conversation-filter-prompt},
\ref{fig:appendix-user-unrelated-question-filter-prompt}, and
\ref{fig:appendix-user-unrelated-answer-filter-prompt} show the filters, and
Figure~\ref{fig:appendix-user-unrelated-instance-example} shows a real
evaluation instance from the same case.

\begin{figure*}[!tbp]
\begin{subtlememorycasebox}{Non-user evaluation instance excerpt for a Multi-evidence variant set}
\small
\textbf{Target query.} Before I finish my notes, can you remind me which symbol
flower belongs to Busan, Incheon, and Ulsan?

\textbf{Reference correct answer.} Busan's symbol flower is Camellia,
Incheon's is Rose, and Ulsan's is Pear flower.

\textbf{Plausible incorrect behavior.} Busan's symbol flower is Camellia,
Incheon's is Pear flower, and Ulsan's is Rose.

\textbf{Required evidence use.} The answer must combine facts distributed
across the two sessions: Busan and Ulsan are mentioned in the learning
explanation session, while Incheon is mentioned in the artifact-production
session.
\end{subtlememorycasebox}
\caption{Real non-user evaluation-instance excerpt for the same Multi-evidence
variant set shown in
Figure~\ref{fig:appendix-user-unrelated-session-example}. The target query is
answerable only if the system recovers and combines the relevant facts across
sessions.}
\label{fig:appendix-user-unrelated-instance-example}
\end{figure*}

\begin{figure*}[!tbp]
\begin{subtlememorypromptbox}{Non-user query-generation prompt excerpts}
\textbf{Complementary.}
\emph{``Write one new follow-up question based on the sessions below.''}
\begin{itemize}
\item Produce exactly one question that is answerable from the stored sessions
alone, sounds like a natural later follow-up from the same user, is
self-contained, and does not become a meta-question about memory, sessions, or
process.
\item For \textit{Multi-evidence}, preserve the hidden dependency structure
instead of restating every hop in the question. Do not enumerate all
intermediate facts, titles, locations, or aliases; use light anaphora or
natural references when possible.
\item For MuSiQue-derived chains, avoid riddle-like nested clauses and preserve
the benchmark difficulty through recalled session context and the final
reasoning relation rather than through over-engineered surface wording.
\end{itemize}

\textbf{Nuanced Contextual.}
\emph{``Write one or two new follow-up questions based on the sessions
below.''}
\begin{itemize}
\item Every question must specify the decisive context and target exactly one
context fact. The answer should be a single context-specific answer, not a full
list of all contexts.
\item If two questions are produced, they must target different memory
identifiers and ask from meaningfully different angles.
\item Each question must remain understandable after many unrelated sessions
have been stored. Include enough topic, entity, and context clues, such as
role, location, jurisdiction, scope, version, definition, unit, attribute,
object range, or scenario.
\item Do not ask the broad ambiguous question directly, ask for all contexts,
or use vague local references such as ``what is it in that case?''
\end{itemize}

\textbf{Nuanced Temporal.}
\emph{``Write exactly 3 new follow-up questions based on the sessions
below.''}
\begin{itemize}
\item Produce exactly three question objects, each asking for the answer to the
same underlying question under one temporal condition and resolving to exactly
one target memory identifier.
\item Cover exactly one easy, one medium, and one hard question. Easy questions
may name a direct as-of date; medium questions should use a natural event,
task, draft, note, version, or session anchor; hard questions should use
relative temporal reasoning or a latest-before anchor.
\item Include enough topic, entity, fact-property, and temporal or task clues
for a memory system to locate the intended snapshot.
\item Do not ask for all dates, all periods, a full timeline, chronological
order, or the current/latest answer without a temporal anchor.
\end{itemize}

\textbf{Contradictory.}
\emph{``Write one new question based on the independent sessions below.''}
\begin{itemize}
\item Produce exactly one self-contained, natural question that targets the main
disputed point.
\item Do not mention ``contradiction,'' ``conflict,'' or ``different sessions,''
and do not add qualifiers that would resolve the contradiction.
\item Start from the canonical conflict question. It may be reused directly or
lightly paraphrased, but the paraphrase must not choose a side or reintroduce
time, version, edition, role, scope, or context qualifiers that make both
claims compatible.
\end{itemize}

\textbf{Output schema.}
\emph{``Return JSON in this format:''}
\texttt{\{question\}} for complementary-relation and contradictory-relation cases, and
\texttt{\{questions: [\{target\_memory\_id, question,
question\_design\_notes, ...\}]\}} for nuanced-relation cases.
\end{subtlememorypromptbox}
\caption{Non-user query-generation prompts for producing target queries
$q_\tau$ from accepted external-knowledge sessions. The box summarizes the
relation-specific prompt variants for complementary-relation, nuanced-relation
Contextual, nuanced-relation Temporal, and contradictory-relation cases.}
\label{fig:appendix-user-unrelated-query-generation-prompt}
\end{figure*}

\begin{figure*}[!tbp]
\begin{subtlememorypromptbox}{Non-user answer-generation prompt excerpts}
\textbf{Shared instruction.}
\emph{``Generate answer candidates for the question below.''}
\begin{itemize}
\item Produce exactly three correct answers and exactly three incorrect
answers.
\item Each answer must be a dictionary with one key, \texttt{text}.
\item Keep all six answers plausible, natural, concise, and reasonably similar
in style and length without repeating the same wording.
\end{itemize}

\textbf{Complementary.}
\begin{itemize}
\item Every correct answer must mean the same thing as the canonical answer.
\item If the canonical answer is structured, preserve all entities, pairings,
and key details completely.
\item Incorrect answers should prefer distractor-grounded local confusions
instead of arbitrary unrelated alternatives.
\end{itemize}

\textbf{Nuanced Contextual.}
\begin{itemize}
\item Every correct answer must be the single answer for the target context
only.
\item Correct answers may briefly restate the context, but must not expand into
all context-conditioned answers.
\item Incorrect answers should reflect realistic context confusions, such as
choosing the answer from another context for the same broad topic; do not use
random unrelated wrong answers.
\end{itemize}

\textbf{Nuanced Temporal.}
\begin{itemize}
\item Every correct answer must mean the same thing as the target answer for
the requested temporal snapshot.
\item Correct answers should preserve key entities, numbers, and names, and may
briefly mention the temporal anchor, but must not expand into the whole
timeline.
\item Incorrect answers should reflect realistic temporal confusions, such as
choosing an answer from a different snapshot date for the same underlying
question.
\end{itemize}

\textbf{Contradictory.}
\begin{itemize}
\item Every correct answer must explicitly state that the remembered sessions
conflict or remain unresolved.
\item Correct answers must not choose a side, pretend one claim is final, or
quietly repair the contradiction.
\item Good correct answers may name the competing claims and say that
clarification or verification is needed.
\item Incorrect answers should reflect failure modes such as picking one side,
pretending one session overrules the other, inventing a clean resolution, or
ignoring the practical framing of the question.
\end{itemize}

\textbf{Output schema.}
\emph{``Return JSON in this format:''}
\texttt{\{correct\_answers: [\{text\}, \{text\}, \{text\}],
incorrect\_answers: [\{text\}, \{text\}, \{text\}]\}}.
\end{subtlememorypromptbox}
\caption{Non-user answer-candidate generation prompts for producing reference
correct answers $A^+$ and plausible incorrect answers $A^-$ under each
compatibility relation type.}
\label{fig:appendix-user-unrelated-answer-generation-prompt}
\end{figure*}

\begin{figure*}[!p]
\begin{subtlememorypromptbox}{Non-user conversation-filter prompt excerpts}
\emph{``You are judging one [complementary/nuanced/contradictory]-memory
benchmark sample. Return valid JSON only and follow the requested schema
exactly. Judge strictly against the subtype requirements and naturalness
requirements. Think step by step.''}

\emph{``Complementary conversation-stage judging requirements:''}
\begin{itemize}
\item Judge only the conversation quality, not the final benchmark question or
answer candidates.
\item The multi-session conversations should feel like ordinary user-assistant
interactions, not quiz reveals, benchmark templates, or explicit fact dumps.
\item \emph{[Condensed from original prompt:]} Accept varied task scenarios;
reject sessions that fail the complementary subtype, leak the target too
directly, or integrate source facts unnaturally. \ldots
\end{itemize}

\emph{``Nuanced conversation-stage judging requirements:''}
\begin{itemize}
\item Judge only the sessions and how the nuanced facts are embedded.
\item Sessions should feel natural, distribute facts across interactions, and
avoid explicit benchmark recaps.
\item \emph{[Condensed from original prompt:]} For Temporal, sessions must
embed as-of/snapshot/version answers for the same underlying question, not
separate event dates; for Contextual, facts must remain valid under different
conditions. Reject samples that become contradiction, separate-entity facts, or
explicit recaps. \ldots
\end{itemize}

\emph{``Contradictory conversation-stage judging requirements:''}
\begin{itemize}
\item Judge only the sessions and how the contradictory facts are embedded.
\item Sessions should feel independent and ordinary, with the contradiction
emerging across sessions rather than through explicit debate.
\item \emph{[Condensed from original prompt:]} Reject sessions that openly
announce, debate, choose a side, or resolve the contradiction, or that make the
facts separable by time, version, or context. \ldots
\end{itemize}

\emph{``Return JSON in this format: \ldots''}
\end{subtlememorypromptbox}
\caption{Non-user conversation-filter prompts for validating generated sessions
before target-query construction.}
\label{fig:appendix-user-unrelated-conversation-filter-prompt}
\end{figure*}

\begin{figure*}[!p]
\begin{subtlememorypromptbox}{Non-user question-filter prompt excerpts}
\emph{``You are judging the question quality of one
[complementary/nuanced/contradictory]-memory benchmark sample. Return valid
JSON only and follow the requested schema exactly. Judge strictly against the
subtype requirements and question-design requirements. Think step by step.''}

\emph{``Complementary question-stage judging requirements:''}
\begin{itemize}
\item Judge only the final benchmark question, with access to the sessions for
context.
\item Do not judge the answer candidates here.
\item The question should sound like a natural later follow-up and fit the
subtype: \textit{Multi-evidence} requires integrating multiple facts, while
\textit{Any-one} targets a shared answer recoverable from any equivalent
mention.
\item \emph{[Condensed from original prompt:]} Reject questions that are too
direct, unnatural, subtype-mismatched, or answerable in the wrong way. \ldots
\end{itemize}

\emph{``MuSiQue-specific judging adjustment:''}
\emph{``[Condensed from original prompt:] MuSiQue questions may use concrete
names, short labels, or session anchors; long chains are allowed; do not
mechanically enforce the original effective\_k count if the question still
requires meaningful memory-grounded synthesis. \ldots''}

\emph{``Nuanced question-stage judging requirements:''}
\begin{itemize}
\item The question should sound natural and subtype-appropriate.
\item \emph{[Condensed from original prompt:]} For Temporal, generate exactly
three questions with different angles and difficulty levels; each must resolve
to one dated/as-of/version snapshot for the same underlying question and should
not ask for a timeline, all dates, latest/current without an anchor, or a full
recap. \ldots
\item \emph{[Condensed from original prompt:]} For Contextual, specify one
decisive context such as role, location, scope, version, definition, unit,
attribute, jurisdiction, or scenario; reject broad ambiguous questions, full
lists, all-context questions, or questions answerable by more than one selected
context fact. \ldots
\end{itemize}

\emph{``Contradictory question-stage judging requirements:''}
\begin{itemize}
\item The question should directly ask about the disputed fact.
\item \emph{[Condensed from original prompt:]} Reject questions that quietly
add resolving time/version/context qualifiers, turn into meta-questions about
memory or disagreement, or fail to target the disputed point. \ldots
\end{itemize}

\emph{``Return JSON in this format: \ldots''}
\end{subtlememorypromptbox}
\caption{Non-user question-filter prompts for validating target queries under
complementary, nuanced, and contradictory relation requirements.}
\label{fig:appendix-user-unrelated-question-filter-prompt}
\end{figure*}

\begin{figure*}[!p]
\begin{subtlememorypromptbox}{Non-user answer-filter prompt excerpts}
\emph{``You are judging the answer quality of one
[complementary/nuanced/contradictory]-memory benchmark sample. Return valid
JSON only and follow the requested schema exactly. Judge strictly against the
subtype requirements and answer-design requirements. Think step by step.''}

\emph{``Complementary answer-stage judging requirements:''}
\begin{itemize}
\item Judge only the answer candidates.
\item The three correct answers should all be genuinely correct for the
question and semantically aligned with the intended canonical answer.
\item The answers should match the subtype: \textit{Multi-evidence} preserves
the integrated multi-fact result, while \textit{Any-one} targets the shared
answer.
\item \emph{[Condensed from original prompt:]} Incorrect answers are auxiliary
distractor metadata; do not reject merely because they are easy or weak, but
reject actually correct/equivalent distractors, ambiguous answer sets, wrong
correct answers, incomplete correct answers, or subtype-mismatched correct
answers. \ldots
\end{itemize}

\emph{``Nuanced answer-stage judging requirements:''}
\begin{itemize}
\item All three correct answers must be genuinely correct for the subtype and
question.
\item For Temporal, all correct answers should match only the target
dated/as-of/version snapshot. If the question uses an event/version/relative
anchor rather than an explicit date, use the sessions and
\texttt{target\_memory\_id} metadata to judge the intended snapshot.
\item For Contextual, all correct answers should match only the context
explicitly specified by the question.
\item \emph{[Condensed from original prompt:]} Incorrect answers are auxiliary
distractor metadata; reject actually valid/equivalent distractors, ambiguous
answer sets, incomplete or wrong correct answers, all-context answers,
clarification answers, wrong-context answers, or subtype-mismatched correct
answers. \ldots
\end{itemize}

\emph{``Contradictory answer-stage judging requirements:''}
\begin{itemize}
\item All correct answers must explicitly acknowledge the contradiction,
uncertainty, or need for clarification/verification.
\item Correct answers must not choose one side as final.
\item \emph{[Condensed from original prompt:]} Incorrect answers are auxiliary
distractor metadata; do not reject merely because they are easy or repetitive,
but reject valid contradiction-aware distractors, ambiguous answer sets, correct
answers that pick one side, or correct answers that fail to acknowledge
contradiction/uncertainty. \ldots
\end{itemize}

\emph{``Return JSON in this format: \ldots''}
\end{subtlememorypromptbox}
\caption{Non-user answer-filter prompts for validating reference correct
answers $A^+$ and plausible incorrect answers $A^-$.}
\label{fig:appendix-user-unrelated-answer-filter-prompt}
\end{figure*}

\subsection{Stage 5: User-history Assembly}
\label{sec:appendix-data-stage-user-history-assembly}
\label{sec:appendix-merged-histories}

During final assembly, non-user sessions are selected after filtering against
the full pool of user-related variant facts, which removes externally sourced
facts that are too close to persona-specific memories. The remaining non-user
variant sets are assigned to personas without reuse and balanced by
compatibility relation type. Their timestamps are redistributed within each
persona's user-related time span, or within a fixed date range when no span is
available. Each release unit contains a chronological session history and a
benchmark file. The system receives only the history and user request, not
semantic variants, relation labels, or answer candidates.

\subsection{Final data composition.}
Across all construction filters, the final pass rate is 88.86\% for user-related candidate semantic variant sets and 73.23\% for non-user candidate semantic variant sets.
Table~\ref{tab:subtlememory-data-composition} reports the source and
compatibility-relation composition behind the main-text scale summary, and
Table~\ref{tab:subtlememory-domain-distribution} reports the domain
distribution of semantic variant sets.

\begin{table}[!tbp]
\centering
\small
\begin{tabular}{@{}p{0.50\linewidth}rr@{}}
\toprule
\textbf{Measure} & \textbf{Value} & \textbf{Proportion} \\
\midrule
\multicolumn{3}{@{}l}{\textit{Dataset scale}} \\
Persona-level histories & 10 & -- \\
Memory-bearing sessions & 2,364 & -- \\
Total session tokens & $\sim$2.12M & -- \\
Avg. tokens per history & $\sim$211.6K & -- \\
Evaluation instances & 1,522 & -- \\
\hdashline
\multicolumn{3}{@{}l}{\textit{Variant sets by source}} \\
Semantic variant sets & 1,090 & 100.00\% \\
User-related sets & 545 & 50.00\% \\
Non-user sets & 545 & 50.00\% \\
\hdashline
\multicolumn{3}{@{}l}{\textit{Variant sets by relation}} \\
Semantic variant sets & 1,090 & 100.00\% \\
Complementary sets & 361 & 33.12\% \\
Nuanced sets & 352 & 32.29\% \\
Contradictory sets & 377 & 34.59\% \\
\hdashline
\multicolumn{3}{@{}l}{\textit{Instances by source}} \\
Evaluation instances & 1,522 & 100.00\% \\
User-related instances & 707 & 46.45\% \\
Non-user instances & 815 & 53.55\% \\
\hdashline
\multicolumn{3}{@{}l}{\textit{Instances by relation}} \\
Evaluation instances & 1,522 & 100.00\% \\
Complementary instances & 361 & 23.72\% \\
Nuanced instances & 784 & 51.51\% \\
Contradictory instances & 377 & 24.77\% \\
\bottomrule
\end{tabular}
\caption{SubtleMemory data composition. Proportions are computed within each
dashed group. Token counts are approximate session-token counts.}
\label{tab:subtlememory-data-composition}
\end{table}

\begin{table}[!tbp]
\centering
\small
\begin{tabular*}{\linewidth}{@{\extracolsep{\fill}}lrr@{}}
\toprule
\textbf{Domain} & \textbf{Variant sets} & \textbf{Proportion} \\
\midrule
Culture & 191 & 17.52\% \\
Media & 188 & 17.25\% \\
Competition & 147 & 13.49\% \\
World & 143 & 13.12\% \\
Society & 123 & 11.28\% \\
Cuisine & 67 & 6.15\% \\
Lifestyle & 65 & 5.96\% \\
Development & 59 & 5.41\% \\
STEM & 55 & 5.05\% \\
Nature & 52 & 4.77\% \\
\midrule
Total & 1,090 & 100.00\% \\
\bottomrule
\end{tabular*}
\caption{Topic-domain distribution of SubtleMemory semantic variant sets.
Counts and proportions are computed at the set level.}
\label{tab:subtlememory-domain-distribution}
\end{table}

\subsection{Human Validation of Benchmark Instances}
\label{sec:appendix-instance-human-validation}

We conduct a human audit to assess whether the constructed benchmark instances
preserve their intended evidence, relations, and reference answers.

\paragraph{Sampling and annotation.}
We sample 150 evaluation instances stratified by relation type, including 50
Complementary, 50 Nuanced, and 50 Contradictory instances. Two annotators
independently review each instance without access to its predefined relation
label or subtype. For each instance, the annotators examine the interaction
history, target semantic variants, resolution target, evaluation query, and
reference answer.

Each annotator performs the following four steps:
\begin{enumerate}
    \item \textbf{Evidence recovery.}
    Locate each target semantic variant in the interaction history and record
    its supporting session IDs and corresponding evidence.

    \item \textbf{Relation labeling.}
    Infer whether the target variants are Complementary, Nuanced, or
    Contradictory based on the variants and their shared resolution target.

    \item \textbf{Contradiction checking.}
    For instances identified as Contradictory, determine whether the variants
    are mutually incompatible under the resolution target or can be reconciled
    through temporal, contextual, scope, version, or role differences.

    \item \textbf{Answer-key verification.}
    Determine whether the reference answer is supported by the recovered
    evidence and correctly handles the inferred relation.
\end{enumerate}

\paragraph{Validation metrics.}
We derive four metrics from the annotations. \textit{Evidence recoverability}
requires every target semantic variant to be grounded in specific supporting
sessions. \textit{Relation-label correctness} requires the relation inferred by
the annotator to match the predefined benchmark relation.
\textit{Contradictory validity} requires the conflicting variants to remain
irreconcilable under the resolution target. \textit{Answer-key correctness}
requires the reference answer to be supported by the interaction history and
consistent with the inferred relation. An instance is counted as
\textit{overall validated} only when it passes all applicable checks.

For each metric, we report the mean validation rate across the two annotators
over all applicable annotations, with disagreements retained in the
denominator. Inter-annotator agreement is the proportion of jointly applicable
annotations on which the two annotators provide the same judgment.

\begin{table}[t]
\centering
\small
\setlength{\tabcolsep}{4pt}
\begin{tabular}{lrr}
\toprule
\textbf{Validation metric}
& \textbf{Average}
& \textbf{Agreement} \\
\midrule
Evidence recoverability
& 99.7\%
& 99.3\% \\
Relation-label correctness
& 96.6\%
& 95.9\% \\
Contradictory validity
& 93.8\%
& 91.7\% \\
Answer-key correctness
& 99.7\%
& 99.3\% \\
Overall validated
& 95.9\%
& 94.5\% \\
\bottomrule
\end{tabular}
\caption{Human validation results on 150 stratified benchmark instances.
Average validation rates are computed across the two annotators.}
\label{tab:appendix-instance-human-validation}
\end{table}

\paragraph{Results.}
As shown in Table~\ref{tab:appendix-instance-human-validation}, the audit yields
a 95.9\% overall validation rate with 94.5\%
inter-annotator agreement. Evidence recoverability and answer-key correctness
both reach 99.7\%, while relation-label correctness and Contradictory validity
reach 96.6\% and 93.8\%, respectively. These results provide quantitative
support for the internal validity of the audited benchmark instances. The audit
does not measure conversational naturalness or provide an end-to-end
human-performance ceiling.

\subsection{Target-Dependence Validation}
\label{sec:memory-dependence}

To verify that the non-user evaluation instances require information from the
interaction history rather than parametric knowledge alone, we evaluate
GPT-5.4 on all 815 non-user instances using only the query, without providing
the interaction history or retrieved memory. We use the same answer prompt as
in the Oracle setting, where the model receives the annotated target sessions.

\begin{table}[t]
\centering
\small
\begin{tabular}{lrrr}
\toprule
\textbf{Relation} & \textbf{Query only} & \textbf{Oracle}
& \textbf{$\Delta$ (pp)} \\
\midrule
Complementary & 3.37 & 91.57 & 88.20 \\
Nuanced       & 0.88 & 98.46 & 97.59 \\
Contradictory & 0.00 & 91.71 & 91.71 \\
\midrule
Overall       & 1.23 & 95.46 & 94.23 \\
\bottomrule
\end{tabular}
\caption{Accuracy (\%) on the 815 non-user instances using GPT-5.4. Both
settings use the same answer prompt and differ in whether the annotated
interaction-history evidence is provided.}
\label{tab:memory-dependence}
\end{table}

As shown in Table~\ref{tab:memory-dependence}, we obtain 1.23\% overall accuracy
using only the query, compared with 95.46\%
when providing the annotated history evidence. The large gap across all three
relation types indicates that the evaluated non-user instances depend on
interaction-history evidence rather than parametric knowledge alone.

\subsection{Artifact Availability, Licenses, and Intended Use}
\label{sec:appendix-artifact-license-use}

\href{https://github.com/Yummytanmo/SubtleMemory}{Code} and
\href{https://huggingface.co/datasets/Yummytanmo/SubtleMemory}{data}
are publicly available under the Apache License 2.0. The code repository
contains the construction and evaluation pipelines, prompts, and reproduction
configurations, while the data repository contains the generated histories and
evaluation instances.

Existing seed resources remain governed by their original licenses and access
terms. We use them only as seed sources for constructing synthetic benchmark
histories and evaluation instances, cite the original resources, and do not
redistribute the original source artifacts as standalone data unless their terms
permit redistribution. SubtleMemory artifacts are intended for research
evaluation of long-term memory agents, not for profiling real individuals or
deploying personalized assistants from the synthetic histories. The released
artifacts should be used consistently with this research-evaluation purpose and
with any source-specific access conditions.

\section{Experiments}
\label{sec:appendix-experiments}

\subsection{Experimental Setup}
\label{sec:appendix-experimental-setup}

\subsubsection{Evaluation-time Baseline and Model Settings}
\label{sec:appendix-evaluated-system-settings}

In the actual evaluation, each memory system is run through its native write
and retrieval interfaces using the default or author-recommended configuration
reported in Table~\ref{tab:appendix-evaluated-system-settings}. Systems retain
their native chunking, storage, retrieval, and parameter settings; no system is
tuned on a common development set, and storage and inference budgets are not
matched across systems. The results therefore compare the end-to-end memory
mechanisms under their evaluated configurations rather than individual
components or cost-matched efficiency.

\begin{table*}[!tbp]
\centering
\scriptsize
\setlength{\tabcolsep}{3pt}
\renewcommand{\arraystretch}{1.12}
\begin{tabular}{@{}p{0.105\textwidth}p{0.195\textwidth}p{0.320\textwidth}p{0.320\textwidth}@{}}
\toprule
\textbf{Baseline} &
\textbf{System source} &
\textbf{Configurable hyperparameters} &
\textbf{Memory-stage model} \\
\midrule
\textbf{Mem0} &
Platform API. &
\texttt{search.top\_k}=20; \texttt{max\_content\_length}=8000. &
Platform default add model. \\
\midrule
\textbf{MemOS} &
Platform API. &
\texttt{search.top\_k}=20. &
Platform default add model. \\
\midrule
\textbf{EverMemOS} &
Platform API. &
\texttt{search.top\_k}=10; \texttt{retrieve\_method}=hybrid; \texttt{memory\_types}=\{\texttt{episodic\_memory}\}. &
Platform default add model. \\
\midrule
\textbf{A-Mem} &
Local GitHub implementation. &
\texttt{search.top\_k}=20; retriever=\texttt{all-MiniLM-L6-v2}; \texttt{evo\_threshold}=100. &
Add model=\texttt{gpt-4o-mini}. \\
\midrule
\textbf{MIRIX} &
Local GitHub implementation. &
\texttt{search.top\_k}=20; embedding model=\texttt{text-embedding-3-small}. &
Add/search control=\texttt{gpt-4o-mini}; final answer uses configured answer model. \\
\midrule
\textbf{MemoBase} &
Self-hosted/local API service. &
\texttt{search.backend}=context; \texttt{top\_k}=20; max context=3000 tokens; event thresholds=0.2. &
Add model=\texttt{gpt-4o-mini}. \\
\midrule
\textbf{OpenClaw} &
Local OpenClaw runtime. &
\texttt{search.top\_k}=50. &
No LLM write model. \\
\midrule
\textbf{MetaClaw} &
Local MetaClaw runtime. &
\texttt{retrieval\_mode}=hybrid; \texttt{top\_k}=8; max injected memory tokens=2400; max context chars=18000. &
No separate write model configured. \\
\midrule
\textbf{OpenClaw-style plugins} &
Local OpenClaw runtime plus backend plugin. &
\texttt{search\_top\_k}=50; MemOS memory/preference limits=20; EverMemOS retrieval=hybrid episodic. &
Backend write model follows the selected plugin. \\
\midrule
\textbf{Oracle context} &
Local oracle adapter. &
None. &
No write-stage model. \\
\bottomrule
\end{tabular}
\caption{Evaluation-time baseline settings. The table lists the implementation
source, fixed memory hyperparameters, and memory-stage model configuration used
for each evaluated baseline.}
\label{tab:appendix-evaluated-system-settings}
\end{table*}

Answer generation is separated from memory construction for systems that expose
retrieved context to the evaluation pipeline, while native agent settings let
the framework run its own answer loop and prompt injection. We use the
GPT-family models in Table~\ref{tab:appendix-answer-model-settings} for answer
generation, and use Gemini 3.1 Pro Preview Thinking as the LLM judge, which
compares each generated answer against accepted correct references, known
incorrect references, and case metadata. Table~\ref{tab:appendix-answer-model-settings}
summarizes the model sources and key hyperparameters used in the experiments.

\begin{table*}[!tbp]
\centering
\renewcommand{\arraystretch}{1.10}
\resizebox{\textwidth}{!}{%
\begin{tabular}{llll}
\toprule
\textbf{Model} & \textbf{Parameter setting} & \textbf{Source} & \textbf{URL} \\
\midrule
GPT-4o mini & temperature=0.0, max tokens=16,384 & \texttt{gpt-4o-mini} & \url{https://platform.openai.com} \\
GPT-5.4 & temperature=0.0, max tokens=16,384 & \texttt{gpt-5.4} & \url{https://platform.openai.com} \\
GPT-OSS 120B & temperature=0.0, max tokens=16,384 & \texttt{openai/gpt-oss-120b} & \url{https://platform.openai.com} \\
Gemini 3.1 Pro Preview Thinking & temperature=0.0 & \texttt{gemini-3.1-pro-preview-thinking} & \url{https://ai.google.dev} \\
\bottomrule
\end{tabular}
}
\caption{Answer-generation and judge model settings, including decoding
parameters, model identifiers, and provider URLs.}
\label{tab:appendix-answer-model-settings}
\end{table*}

\subsubsection{Context Organization with OpenClaw}
\label{sec:appendix-openclaw-context-organization}

Table~\ref{tab:appendix-evaluated-system-settings} distinguishes standalone
memory systems from OpenClaw-based agent deployments, but this difference is
not only a change in the memory backend. As discussed in the main results
around Table~\ref{tab:openclaw-integration-effect}, adding OpenClaw can change
performance even when the external memory system and the target query remain
comparable. The reason is that OpenClaw~\citep{openclaw2026docs} changes the
answer-time context organization: standalone memory systems serialize retrieved
items directly into the benchmark answer prompt, while OpenClaw-based systems
route the same query through an agent workspace whose instructions, plugin
recall, and current user turn are organized by the runtime.

Figure~\ref{fig:appendix-openclaw-context-organization-example} gives an
artifact-level example using the same Mem0 book-selection query. The example
shows what ``context organization'' means concretely: without OpenClaw, the
answer model receives a flat numbered memory list inside the
\texttt{\# CONTEXT} field of the benchmark prompt; with OpenClaw, the answer
rules are loaded as workspace instructions and the Mem0 plugin recall is
injected as agent runtime context before the target query. This illustrates why
OpenClaw integration should be interpreted as an agent-context intervention,
not simply as another formatting of the same retrieved list.

\begin{figure*}[!tbp]
\begin{subtlememorypromptbox}{Answer-context organization with and without OpenClaw}
\textbf{Example setting.} We compare one \texttt{gpt-5.4} Mem0 item from the
complementary book-selection split. Both runs receive the same target query:
choose two books from a hold shelf containing \textit{So Long a Letter},
\textit{The Concubine}, \textit{Passing}, \textit{Open Water},
\textit{Homegoing}, and \textit{A Month in the Country}. The reference answer
selects \textit{So Long a Letter} and \textit{The Concubine} because both match
the remembered preference for African literary classics.

\SubtleParaGap
\textbf{Without OpenClaw: standalone Mem0.} The evaluation pipeline directly
assembles one answer prompt. The prompt contains the benchmark answer rules,
then a numbered \texttt{\# CONTEXT} block built from Mem0 retrieval, and then
the target query. The saved context for this item begins as follows:
\begin{quote}
\scriptsize
\textit{1. 12:10 pm on 21 April, 2025: Assistant recommended
\textit{So Long a Letter} as the gliding novel and \textit{Nervous Conditions}
as the push-back choice, noting \textit{Efuru} as a softer backup if needed.}

\textit{2. 12:08 pm on 21 April, 2025: Assistant recommended five compact
literary classics: \textit{So Long a Letter}, \textit{Efuru},
\textit{Nervous Conditions}, \textit{The Concubine}, and
\textit{Weep Not, Child}.}

\textit{3. 12:08 pm on 21 April, 2025: User prefers the ``compact literary
classics'' lane, gravitating toward works similar to Chinua Achebe and Buchi
Emecheta, and wants books between 180 and 300 pages.}

\textit{4. 12:10 pm on 21 April, 2025: Assistant ranked the five suggested
compact literary classics: \textit{So Long a Letter} first,
\textit{Efuru} second, \textit{Nervous Conditions} third,
\textit{The Concubine} fourth, and \textit{Weep Not, Child} last.}
\end{quote}
The answer model therefore sees the benchmark instructions, the retrieved
memory list, and the query in one benchmark-formatted prompt.

\SubtleParaGap
\textbf{With OpenClaw: Mem0 + OpenClaw.} The target query is passed to the
OpenClaw local agent as the current user turn. The answer rules are loaded from
the OpenClaw workspace instructions, and the Mem0 plugin supplies recalled
memory as OpenClaw runtime context before the user turn. The corresponding
runtime-injected context begins as follows:
\begin{quote}
\scriptsize
\textit{OpenClaw workspace instruction excerpt:}
\begin{itemize}[leftmargin=*, itemsep=1pt, topsep=1pt]
\item \textit{This is a SubtleMemory memory-system evaluation workspace.}
\item \textit{Answer the current benchmark question using only the current
question and OpenClaw recall-injected memory context.}
\item \textit{Recalled memory/context may be auto-recalled or available through
the configured OpenClaw memory backend or plugin.}
\item \textit{Do not use outside knowledge, browser/web search, or unrelated
workspace files. Keep the final answer concise and grounded in the recalled
memory context.}
\item \textit{SubtleMemory Answer Rules: (1) Use only the recalled
memory/context injected before the question. (2) Your first priority is
evidence fidelity: detect true unresolved conflicts, but do not over-detect
conflicts from compatible evidence. (3) Identify the exact target needed by the
question \ldots}
\end{itemize}

\textit{1. 12:10 pm on 21 April, 2025: Assistant ranked the five compact
literary classics, naming \textit{So Long a Letter} as the best first pick,
\textit{Efuru} as steady and lucid, \textit{Nervous Conditions} as the edgier
option, \textit{The Concubine} as more atmospheric, and advised saving
\textit{Weep Not, Child} for a later weekend.}

\textit{2. 12:08 pm on 21 April, 2025: Assistant recommended five compact
literary classics fitting the 180--300 page range: \textit{So Long a Letter},
\textit{Efuru}, \textit{Nervous Conditions}, \textit{The Concubine}, and
\textit{Weep Not, Child}.}

\textit{3. 12:08 pm on 21 April, 2025: User requests that at least one
suggested novel be authored by a woman, prefers clean prose over stylistic
fireworks, and wants to avoid emotionally heavy reading.}

\textit{4. 12:08 pm on 21 April, 2025: User prefers the ``compact literary
classics'' lane, enjoys works similar to those by Chinua Achebe and Buchi
Emecheta, and wants books roughly 180--300 pages long.}
\end{quote}
The answer model therefore sees an agent-organized input: persistent workspace
instructions, plugin-recalled context injected by OpenClaw, and the current
query as the user turn.

\SubtleParaGap
\textbf{Difference.} The memory backend can be the same, and the target query
is unchanged, but the context organization differs. Standalone Mem0 exposes
retrieval output as a flat benchmark \texttt{\# CONTEXT} list inside the answer
prompt. Mem0 + OpenClaw exposes plugin recall as runtime-injected agent context
under OpenClaw workspace rules. This changes where instructions live, how
retrieved memories are packaged, and how the final answer model encounters the
same evidence.
\end{subtlememorypromptbox}
\caption{Concrete context-organization example for a standalone Mem0 run and a
Mem0 + OpenClaw run on the same target query.}
\label{fig:appendix-openclaw-context-organization-example}
\end{figure*}

\subsubsection{LLM-as-judge Validation}
\label{sec:appendix-llm-judge-validation}

The judge prompt and validation sample are specified as follows:
\begin{itemize}
\item The judge receives the target query, generated answer, accepted correct
answers $A^+$, known incorrect answers $A^-$, supporting facts, case
description, relation metadata, source, and relation-specific grading guidance.
\item The output is a binary correctness label with a short reason. Grading is
semantic rather than surface-form based.
\item For validation, we use 45 fixed-seed queries from one of the ten user-history splits, with 15 from each relation type and five candidate answers per query, yielding 225 human-labeled answers.
\item Human annotators receive the same information as the judge. The automatic
labels reach Cohen's $\kappa=0.963$ against human labels.
\end{itemize}
Figure~\ref{fig:appendix-llm-judge-prompt} summarizes the judge prompt used for
binary answer evaluation.

\begin{figure*}[!tbp]
\begin{subtlememorypromptbox}{LLM-as-judge prompt summary}
\textbf{System role.} The judge acts as a benchmark answer judge and decides
whether the generated answer should be labeled \texttt{CORRECT} or
\texttt{WRONG}. It uses accepted correct answers as positive references, known
incorrect answers as negative references, and relation metadata as auxiliary
context.

\SubtleParaGap
\textbf{Main grading rules.}
\begin{itemize}
\item Mark \texttt{CORRECT} when the generated answer matches the meaning of an
accepted correct answer, including paraphrases or complete longer explanations.
\item Mark \texttt{WRONG} when the answer agrees with a known incorrect answer,
contradicts the accepted answer set, misses the key point, or makes an
unsupported overconfident choice.
\item Use facts, case description, relation type, relation subtype, topic,
source, and additional guidance to resolve grading ambiguity.
\item For contradictory-relation and nuanced-relation cases, do not reward
answers that remove the required distinction or conflict without evidence.
\end{itemize}

\SubtleParaGap
\textbf{Required output.} The prompt requires JSON with exactly two keys,
\texttt{reason} and \texttt{label}.
\end{subtlememorypromptbox}
\caption{Prompt summary for binary LLM-as-judge answer evaluation.}
\label{fig:appendix-llm-judge-prompt}
\end{figure*}

\subsubsection{Answer-generation Prompts}
\label{sec:appendix-answer-generation-prompts}

We use the same answer-generation policy across evaluated systems. For each
target query, the answer model receives the evidence exposed by the evaluated
setting as \texttt{\{context\}} and the target query as
\texttt{\{question\}}. The answer prompt does not reveal the hidden semantic
variant set, relation label, accepted correct answers, or known incorrect
answers. Following the main text, we compare a \textit{soft prompt} with
general guidance and a \textit{strong prompt} with explicit instructions for
target identification, conflict recognition, evidence fidelity, and
clarification. Figures~\ref{fig:appendix-soft-answer-prompt} and
\ref{fig:appendix-strong-answer-prompt} show the two answer-generation prompts.

\begin{figure*}[!p]
\begin{subtlememorypromptbox}{Soft Answer Prompt}
\textbf{Role.} You are a helpful personal assistant. You are very good at
distinguishing detailed conflicts and relationships in memory, then using those
memories to answer questions or complete tasks.

\SubtleParaGap
\textbf{\# CONTEXT}

\texttt{\{context\}}

\SubtleParaGap
\textbf{\# INSTRUCTIONS}
\begin{itemize}
\item Answer the question or complete the user's requested task based on the
provided context.
\item First identify the information and details in the context that are useful
for answering the question.
\item If the useful information contains time-based updates, use the time
mentioned by the user to decide which information applies.
\item Some information may apply only in different situations, such as different
ways of speaking in different professional roles.
\item Some questions require using all useful information to provide a complete
answer. In those cases, consider all relevant information when answering or
completing the task.
\item Do not treat the order of information in the context, or the chronological
order of session timestamps, as proof that one piece of information is an
update. The user may disclose older information in a later conversation, so do
not trust only the latest session or the latest context item.
\item When multiple pieces of information could each answer the question or help
complete the task, but they would lead to different answers or outcomes and
there is no clear way to decide, clarify the conflict.
\item When several unconditioned pieces of information would lead to different
answers or task outcomes, do not choose one directly; ask for clarification.
\end{itemize}

\SubtleParaGap
\textbf{Question:} \texttt{\{question\}}

\SubtleParaGap
\textbf{Answer:}
\end{subtlememorypromptbox}
\caption{Soft answer prompt used for answer generation. The box preserves the
main structure and rules of the prompt while omitting non-essential formatting
lines.}
\label{fig:appendix-soft-answer-prompt}
\end{figure*}

\begin{figure*}[!p]
\begin{subtlememorypromptbox}{Strong Answer Prompt}
\textbf{Role.} You are a helpful personal assistant. You are very good at
distinguishing detailed conflicts and relationships in recalled memories, then
using those memories to answer questions or complete tasks.

\SubtleParaGap
\textbf{\# CONTEXT}

The information below provides only the context relevant to the current
question. It may contain stable preferences, temporary states, factual claims,
summaries, or contradictions.

\texttt{\{context\}}

\SubtleParaGap
\textbf{\# INSTRUCTIONS}
\begin{enumerate}
\item Use only the provided context. Do not use outside knowledge, common
knowledge, or your own guess to resolve conflicts.
\item Your first priority is evidence fidelity: detect true unresolved
conflicts, but do not over-detect conflicts from compatible evidence.
\item Identify the exact target needed by the question: preference, choice,
attribute, state, factual answer, origin, date, name, or category.
\item Treat a conflict as mutually exclusive claims about the same target;
different interests, different sources, background facts, or multiple
constraints are not conflicts unless they directly support incompatible answers
to the question target.
\item Search the context for evidence that supports one answer and evidence that
supports a different or opposing answer for the same target, including semantic
opposites with different wording.
\item Treat preference conflicts, incompatible choice evidence, factual answer
conflicts, and conflicts between user statements and summaries as unresolved
unless the context explicitly resolves them with a time frame, context,
correction, current-state update, condition, or exception.
\item Do not assume the newest statement overrides older evidence unless the
context explicitly says it is an update, correction, or current state, and do
not invent a compromise, exception, or hierarchy to make conflicting evidence
fit together.
\item If unresolved conflict affects a recommendation, reservation, purchase,
registration, list, ranking, yes/no answer, or any other decisive answer, do
not choose a side and do not choose a safer alternative. Start with
\texttt{Unclear --- needs clarification first.} Then briefly state both
conflicting sides.
\item For factual questions with two different final answers for the same
target, say the remembered answers conflict rather than selecting the answer
that seems more correct from world knowledge. If the question is explicitly
time-anchored or context-anchored and that anchor clearly selects one side,
answer that side directly.
\item Treat evidence as compatible, and answer directly, when memories can all
be true, one memory gives background while another answers the question, one
memory adds a satisfiable constraint, different phrasings support the same
answer, or the question requires combining multiple facts.
\item For multi-part or choice questions, combine compatible facts across the
context and follow the option that best matches the question target without
calling unrelated background interests a conflict.
\item Keep the answer concise: one or two sentences, without step-by-step
reasoning.
\end{enumerate}

\SubtleParaGap
\textbf{Question:} \texttt{\{question\}}

\SubtleParaGap
\textbf{Answer:}
\end{subtlememorypromptbox}
\caption{Strong answer prompt used for answer generation. The box preserves the
main structure and core conflict-handling rules of the prompt while omitting
long output-pattern examples.}
\label{fig:appendix-strong-answer-prompt}
\end{figure*}

\subsubsection{Oracle and Perfect-retrieval Protocols}
\label{sec:appendix-oracle-perfect-retrieval-protocol}

The evidence settings differ only in what evidence reaches answer generation:
\begin{itemize}
\item \textbf{Oracle Setting.} The answer model receives the raw annotated
target sessions $\mathbf{H}_\tau$ in chronological order, bypassing both memory
formation and retrieval.
\item \textbf{Perfect Retrieval Setting.} The system first writes the full
history into memory. Retrieval is then replaced by provenance-guided readback of
the stored memory units written from $\mathbf{H}_\tau$, denoted
$m_\tau=\mathcal{M}(\mathbf{H}_\tau)$.
\item \textbf{Default Setting.} The target query is issued normally, and the
system's own retrieval or recall mechanism determines the answer evidence.
\end{itemize}
Thus, the Oracle-to-Perfect gap measures whether answer-usable information
survives memory formation, while the Perfect-to-Default gap measures whether
preserved information is exposed by the default retrieval path.

\subsection{Answer Generation Configuration}
\label{sec:appendix-answer-generation-configuration}

Answer generation is calibrated under oracle evidence before comparing memory
systems. The calibration setting is:
\begin{itemize}
\item The calibration uses one complete user-history split with 141 evaluation
queries.
\item Each answer model receives the same raw target sessions
$\mathbf{H}_\tau$ and target query, but no stored memory and no retrieved memory
context.
\item The same judge scores each answer against $A^+$, $A^-$, and relation
metadata, so differences in
Table~\ref{tab:oracle-prompt-model-comparison-summary} reflect answer-model and
prompt behavior rather than storage or retrieval.
\item The calibration compares \texttt{gpt-4o-mini},
\texttt{gpt-oss-120b}, and \texttt{gpt-5.4} under the soft and strong prompts.
The strong prompt is used in the main evaluation because it reduces unsupported
conflict resolution, especially for contradictory-relation cases.
\end{itemize}

\subsection{Statistical Analysis of Main Comparisons}
\label{sec:appendix-statistical-analysis}

We compute uncertainty and significance tests as post-processing over the final
question-level binary correctness labels. Overall confidence intervals use
10,000 nonparametric bootstrap resamples stratified by relation type. Relation-
and subtype-level intervals use bootstrap resampling within the corresponding
subset. For the main paired claim reported here, we compare Oracle Evidence
against the best non-oracle system under each answer model using two-sided exact
McNemar tests over aligned question IDs, with Holm correction across the two
answer-model families.
The best non-oracle system is selected by the overall point estimate in
Table~\ref{tab:main-results}: Mem0 + OpenClaw for \texttt{gpt-5.4} and Mem0
for \texttt{gpt-oss-120b}. We align Oracle and the selected non-oracle run by
\texttt{question\_id}; each evaluation instance contributes one binary pair.
McNemar tests use the discordant counts, and the reported $\Delta$ confidence
interval is estimated with paired bootstrap resampling over the aligned
questions.
Table~\ref{tab:appendix-main-table-accuracy-ci} reports the overall confidence
intervals, and Table~\ref{tab:appendix-primary-paired-tests} reports the paired
Oracle comparisons.

\begin{table}[!tbp]
\centering
\small
\setlength{\tabcolsep}{6pt}
\renewcommand{\arraystretch}{1.08}
\begin{tabular}{@{}lcc@{}}
\toprule
\textbf{Method} & \textbf{Accuracy} & \textbf{95\% CI} \\
\midrule
\rowcolor{BackboneBlue}
\multicolumn{3}{c}{\textbf{Base Model: GPT-5.4}} \\
\midrule
\quad MemoBase & 32.1\% & [29.8\%, 34.4\%] \\
\quad MIRIX & 46.5\% & [44.2\%, 48.8\%] \\
\quad MemOS & 56.8\% & [54.5\%, 59.1\%] \\
\quad Mem0 & 69.0\% & [66.8\%, 71.2\%] \\
\quad EverMemOS & 68.1\% & [66.0\%, 70.2\%] \\
\quad A-Mem & 70.0\% & [67.8\%, 72.1\%] \\
\quad MetaClaw & 20.3\% & [18.3\%, 22.3\%] \\
\quad OpenClaw & 62.5\% & [60.4\%, 64.6\%] \\
\quad MemOS + OpenClaw & 56.5\% & [54.3\%, 58.6\%] \\
\quad Mem0 + OpenClaw & 71.3\% & [69.2\%, 73.4\%] \\
\quad EverMemOS + OpenClaw & 69.1\% & [67.0\%, 71.1\%] \\
\quad Oracle & 85.4\% & [83.7\%, 87.1\%] \\
\midrule
\rowcolor{BackboneViolet}
\multicolumn{3}{c}{\textbf{Base Model: GPT-OSS-120B}} \\
\midrule
\quad MemoBase & 25.3\% & [23.2\%, 27.5\%] \\
\quad MIRIX & 39.8\% & [37.6\%, 42.2\%] \\
\quad MemOS & 52.2\% & [49.9\%, 54.5\%] \\
\quad Mem0 & 64.9\% & [62.7\%, 67.2\%] \\
\quad EverMemOS & 60.4\% & [58.2\%, 62.7\%] \\
\quad A-Mem & 63.9\% & [61.7\%, 66.2\%] \\
\quad MetaClaw & 15.6\% & [13.8\%, 17.4\%] \\
\quad OpenClaw & 52.3\% & [50.1\%, 54.6\%] \\
\quad MemOS + OpenClaw & 47.8\% & [45.5\%, 50.1\%] \\
\quad Mem0 + OpenClaw & 62.9\% & [60.6\%, 65.2\%] \\
\quad EverMemOS + OpenClaw & 59.8\% & [57.6\%, 62.0\%] \\
\quad Oracle & 75.7\% & [73.8\%, 77.5\%] \\
\bottomrule
\end{tabular}
\caption{Overall main-table accuracy with 95\% bootstrap confidence intervals.
Each interval is computed from aligned question-level binary correctness labels,
with bootstrap resampling stratified by relation type.}
\label{tab:appendix-main-table-accuracy-ci}
\end{table}

\begin{table}[!tbp]
\centering
\small
\setlength{\tabcolsep}{3.5pt}
\begin{tabular}{@{}lccc@{}}
\toprule
\textbf{Answer model} & \textbf{$\Delta$ pp} & \textbf{95\% CI} & \textbf{Holm $p$} \\
\midrule
\texttt{gpt-5.4} & +14.1 & [+12.0, +16.4] & $4.96{\times}10^{-34}$ \\
\texttt{gpt-oss-120b} & +10.8 & [+8.5, +13.0] & $5.41{\times}10^{-20}$ \\
\bottomrule
\end{tabular}
\caption{Oracle Evidence versus the best non-oracle system in the main results.
$\Delta$ is Oracle minus the best non-oracle point estimate in percentage
points. Confidence intervals are paired bootstrap intervals, and $p$-values are
two-sided exact McNemar tests with Holm correction.}
\label{tab:appendix-primary-paired-tests}
\end{table}

\subsection{Repeated-run Robustness}
\label{sec:appendix-repeated-run-robustness}

To assess run-to-run stability, we repeat the complete evaluation pipeline for
all six standalone memory systems on two persona histories using a new model
seed under otherwise unchanged evaluation settings. The rerun includes each
system's native memory formation and retrieval, answer generation, and answer
judging.

For each system, we compare the original and repeated runs using the absolute
change in overall accuracy, instance-level agreement, Cohen's $\kappa$, a
paired bootstrap confidence interval for the accuracy difference, and an exact
McNemar test with Holm correction.
Table~\ref{tab:appendix-repeated-run-summary} summarizes the resulting
run-to-run variation across the six standalone systems.

\begin{table}[!tbp]
\centering
\small
\begin{tabular}{@{}lr@{}}
\toprule
\textbf{Measure} & \textbf{Result} \\
\midrule
Mean absolute score change       & 0.85 pp \\
Maximum absolute score change    & 1.59 pp \\
Instance-level agreement         & 91.43--95.87\% \\
Cohen's $\kappa$                 & 0.826--0.905 \\
Paired bootstrap 95\% CIs        & All include zero \\
Holm-adjusted McNemar $p$-values & All 1.000 \\
Spearman's $\rho$                & 1.000 \\
Kendall's $\tau$                 & 1.000 \\
\bottomrule
\end{tabular}
\caption{Repeated-run robustness across six standalone memory systems on two
persona histories. The repeated runs cover the complete evaluation pipeline,
including memory formation, retrieval, answer generation, and judging.}
\label{tab:appendix-repeated-run-summary}
\end{table}

Across the six systems, the mean absolute score change is 0.85 percentage
points and the maximum change is 1.59 points. Instance-level agreement ranges
from 91.43\% to 95.87\%, with Cohen's $\kappa$ ranging from 0.826 to 0.905.
All paired bootstrap confidence intervals include zero, and all Holm-adjusted
exact McNemar $p$-values are 1.000. The system ordering is unchanged, with
both Spearman's $\rho$ and Kendall's $\tau$ equal to 1.000. These results
indicate small run-to-run variation on the evaluated two-history subset without
changing the system-level conclusions.

\subsection{History-level Reliability}
\label{sec:appendix-history-level-reliability}

We conduct two history-level analyses to examine whether the main conclusions
are driven by individual persona histories and whether within-history
dependencies affect the reported Oracle gaps.

\paragraph{Persona-level consistency.}
We analyze the main-experiment results separately across all 10
persona-specific histories for the 12 evaluated methods under both answer
backbones. Across the resulting 24 system--backbone configurations, the
persona-level standard deviation averages 3.2 percentage points and ranges
from 1.9 to 4.6 points. The overall accuracy differs from the corresponding
unweighted persona mean by at most 0.2 points.

We additionally compare the method ranking within each history with the global
ranking. As shown in
Table~\ref{tab:appendix-persona-ranking-consistency}, the resulting
correlations remain high under both answer models.

\begin{table}[!tbp]
\centering
\small
\begin{tabular}{@{}lcc@{}}
\toprule
\textbf{Answer model}
& \textbf{Spearman's $\rho$}
& \textbf{Kendall's $\tau$} \\
\midrule
\texttt{gpt-5.4}      & 0.955 & 0.887 \\
\texttt{gpt-oss-120b} & 0.970 & 0.920 \\
\bottomrule
\end{tabular}
\caption{Mean correlations between persona-level method rankings and the
global rankings across the 10 histories.}
\label{tab:appendix-persona-ranking-consistency}
\end{table}

These results indicate that the broad system-level performance structure is
consistent across histories and is not driven by a single persona.

\paragraph{History-clustered Oracle gaps.}
To account for dependencies among questions and semantic cases from the same
history, we conduct a paired history-cluster bootstrap with 10,000 resamples,
grouping all instances by complete history. We conduct paired significance
testing by enumerating all $2^{10}$ whole-history label swaps and apply Holm
correction across five claim-aligned comparisons within each answer-model
backbone.

Table~\ref{tab:appendix-history-clustered-oracle-gaps} reports the
history-clustered gaps between Oracle and the strongest non-Oracle system under
each answer model.

\begin{table}[t]
\centering
\small
\begin{tabular}{lccc}
\toprule
Answer model & Gap (pp) & 95\% CI & Holm $p$ \\
\midrule
\texttt{gpt-5.4}
& 14.1 & [12.4, 16.0] & 0.00977 \\
\texttt{gpt-oss-120b}
& 10.8 & [8.5, 13.0] & 0.00977 \\
\bottomrule
\end{tabular}
\caption{History-clustered Oracle gaps over the strongest non-Oracle system.
Both gaps are positive across all 10 histories.}
\label{tab:appendix-history-clustered-oracle-gaps}
\end{table}

The Oracle advantage remains positive across all 10 histories and significant
after history clustering and multiple-comparison correction, supporting the
broad conclusion that current systems remain substantially below Oracle
evidence access.

\subsection{Answer-model and Judge Robustness}
\label{sec:appendix-answer-model-judge-robustness}

\paragraph{Answer-model robustness.}
To examine whether the system-level results depend on a particular answer
model, we repeat the main evaluation protocol for all six standalone memory
systems and Oracle on 315 questions from two persona-specific histories. In
addition to \texttt{gpt-5.4} and \texttt{gpt-oss-120b}, we evaluate
\texttt{Kimi-K2.5} and \texttt{Qwen3.7-Plus} as two additional non-OpenAI
answer models. Table~\ref{tab:persona01-answer-model-main-results-comparison}
reports the complete relation-wise results under the four answer models.

\begin{table*}[t]
    \centering
    \footnotesize
    \setlength{\tabcolsep}{4.1pt}
    \renewcommand{\arraystretch}{1.14}
    \resizebox{0.98\textwidth}{!}{%
    \begin{tabular}{@{}lcccccccc@{}}
    \toprule
    \multirow{2}{*}{\textbf{Method}}
    & \multicolumn{3}{c}{\textbf{Complementary}}
    & \multicolumn{3}{c}{\textbf{Nuanced}}
    & \multirow{2}{*}{\textbf{Contradictory}}
    & \multirow{2}{*}{\textbf{All}} \\
    \cmidrule(lr){2-4}\cmidrule(lr){5-7}
    & \textbf{Multi-evidence}
    & \textbf{Any-one}
    & \textbf{Overall}
    & \textbf{Temporal}
    & \textbf{Contextual}
    & \textbf{Overall}
    & & \\
    \midrule

    \rowcolor{BackboneBlue}
    \multicolumn{9}{c}{\textbf{Answer Model: GPT-5.4}} \\
    \midrule
    \quad MemoBase
    & 20.9\% & 37.9\% & 27.8\%
    & 33.0\% & 43.6\% & 38.0\%
    & 7.8\% & 28.3\% \\
    \quad MIRIX
    & 51.2\% & 48.3\% & 50.0\%
    & 50.0\% & 70.5\% & 59.6\%
    & 10.4\% & 45.4\% \\
    \quad MemOS
    & 48.8\% & \textbf{55.2\%} & 51.4\%
    & 72.7\% & 79.5\% & 75.9\%
    & 20.8\% & 56.8\% \\
    \quad Mem0
    & \textbf{62.8\%} & 48.3\% & \textbf{56.9\%}
    & \textbf{81.8\%} & \textbf{91.0\%} & \textbf{86.1\%}
    & \underline{42.9\%} & \textbf{68.9\%} \\
    \quad EverMemOS
    & \underline{58.1\%} & \underline{51.7\%} & \underline{55.6\%}
    & \underline{80.7\%} & \underline{89.7\%} & \underline{84.9\%}
    & 35.1\% & 66.0\% \\
    \quad A-Mem
    & \underline{58.1\%} & 41.4\% & 51.4\%
    & \underline{80.7\%} & 88.5\% & 84.3\%
    & \textbf{48.1\%} & \underline{67.9\%} \\
    \noalign{\vskip 0.18em}
    \cdashline{1-9}
    \noalign{\vskip 0.12em}
    \quad Oracle
    & 69.8\% & 79.3\% & 73.6\%
    & 95.5\% & 93.6\% & 94.6\%
    & 63.6\% & 82.2\% \\
    \midrule

    \rowcolor{BackboneViolet}
    \multicolumn{9}{c}{\textbf{Answer Model: GPT-OSS-120B}} \\
    \midrule
    \quad MemoBase
    & 16.3\% & 44.8\% & 27.8\%
    & 26.1\% & 35.9\% & 30.7\%
    & 5.2\% & 23.8\% \\
    \quad MIRIX
    & 37.2\% & 48.3\% & 41.7\%
    & 42.0\% & 64.1\% & 52.4\%
    & 5.2\% & 38.4\% \\
    \quad MemOS
    & 51.2\% & \textbf{55.2\%} & \underline{52.8\%}
    & 69.3\% & 73.1\% & 71.1\%
    & 19.5\% & 54.3\% \\
    \quad Mem0
    & \textbf{58.1\%} & 48.3\% & \textbf{54.2\%}
    & \textbf{83.0\%} & 82.1\% & \textbf{82.5\%}
    & \textbf{40.3\%} & \textbf{65.7\%} \\
    \quad EverMemOS
    & \underline{53.5\%} & \underline{51.7\%} & \underline{52.8\%}
    & 75.0\% & \textbf{85.9\%} & 80.1\%
    & 20.8\% & 59.4\% \\
    \quad A-Mem
    & \textbf{58.1\%} & 37.9\% & 50.0\%
    & \underline{79.5\%} & \underline{83.3\%} & \underline{81.3\%}
    & \underline{39.0\%} & \underline{63.8\%} \\
    \noalign{\vskip 0.18em}
    \cdashline{1-9}
    \noalign{\vskip 0.12em}
    \quad Oracle
    & 79.1\% & 82.8\% & 80.6\%
    & 92.0\% & 92.3\% & 92.2\%
    & 39.0\% & 76.5\% \\
    \midrule

    \rowcolor{BackboneBlue}
    \multicolumn{9}{c}{\textbf{Answer Model: Kimi-K2.5}} \\
    \midrule
    \quad MemoBase
    & 18.6\% & 37.9\% & 26.4\%
    & 26.1\% & 43.6\% & 34.3\%
    & 5.2\% & 25.4\% \\
    \quad MIRIX
    & 41.9\% & 34.5\% & 38.9\%
    & 45.5\% & 66.7\% & 55.4\%
    & 6.5\% & 39.7\% \\
    \quad MemOS
    & 44.2\% & \underline{48.3\%} & 45.8\%
    & 51.1\% & 73.1\% & 61.4\%
    & 20.8\% & 47.9\% \\
    \quad Mem0
    & \textbf{67.4\%} & \underline{48.3\%} & \textbf{59.7\%}
    & \underline{83.0\%} & \underline{84.6\%} & \underline{83.7\%}
    & \underline{29.9\%} & \underline{65.1\%} \\
    \quad EverMemOS
    & 55.8\% & 44.8\% & 51.4\%
    & 80.7\% & \textbf{89.7\%} & \textbf{84.9\%}
    & 16.9\% & 60.6\% \\
    \quad A-Mem
    & \underline{60.5\%} & \textbf{51.7\%} & \underline{56.9\%}
    & \textbf{85.2\%} & \underline{84.6\%} & \textbf{84.9\%}
    & \textbf{40.3\%} & \textbf{67.6\%} \\
    \noalign{\vskip 0.18em}
    \cdashline{1-9}
    \noalign{\vskip 0.12em}
    \quad Oracle
    & 79.1\% & 82.8\% & 80.6\%
    & 93.2\% & 93.6\% & 93.4\%
    & 49.4\% & 79.7\% \\
    \midrule

    \rowcolor{BackboneViolet}
    \multicolumn{9}{c}{\textbf{Answer Model: Qwen3.7-Plus}} \\
    \midrule
    \quad MemoBase
    & 9.3\% & 44.8\% & 23.6\%
    & 28.4\% & 41.0\% & 34.3\%
    & 6.5\% & 25.1\% \\
    \quad MIRIX
    & 44.2\% & \underline{55.2\%} & 48.6\%
    & 50.0\% & 64.1\% & 56.6\%
    & 2.6\% & 41.6\% \\
    \quad MemOS
    & 48.8\% & 51.7\% & 50.0\%
    & 63.6\% & 75.6\% & 69.3\%
    & 20.8\% & 53.0\% \\
    \quad Mem0
    & \textbf{65.1\%} & 44.8\% & 56.9\%
    & \underline{85.2\%} & \underline{87.2\%} & \underline{86.1\%}
    & \underline{36.4\%} & \underline{67.3\%} \\
    \quad EverMemOS
    & \underline{62.8\%} & \textbf{58.6\%} & \textbf{61.1\%}
    & \textbf{86.4\%} & \textbf{91.0\%} & \textbf{88.6\%}
    & 27.3\% & \underline{67.3\%} \\
    \quad A-Mem
    & 58.1\% & \textbf{58.6\%} & \underline{58.3\%}
    & 83.0\% & \underline{87.2\%} & 84.9\%
    & \textbf{46.8\%} & \textbf{69.5\%} \\
    \noalign{\vskip 0.18em}
    \cdashline{1-9}
    \noalign{\vskip 0.12em}
    \quad Oracle
    & 81.4\% & 79.3\% & 80.6\%
    & 96.6\% & 97.4\% & 97.0\%
    & 39.0\% & 79.0\% \\
    \bottomrule
    \end{tabular}
    }
    \caption{Answer-model robustness on 315 questions from two persona-specific
    histories. Results cover six standalone memory systems and Oracle under
    four answer models and are reported by relation type. Best and second-best
    non-Oracle point estimates within each answer-model block are bolded and
    underlined, respectively.}
    \label{tab:persona01-answer-model-main-results-comparison}
    \vspace{-9pt}
\end{table*}

Across the four answer models, pairwise system-ranking correlations range from
Spearman's $\rho=0.899$ to $1.000$, with a mean of $0.945$. The same three
standalone systems consistently form the leading group, while Oracle remains
above the strongest non-Oracle system by 9.5--13.3 percentage points. Together,
these results support the consistency of the broad system-level findings across
the evaluated answer models.

\paragraph{Judge robustness.}
We further examine whether answer evaluation depends on the selected LLM judge
by re-annotating 282 Oracle responses generated for a 141-question
persona-specific history under two answer models. We use
\texttt{Gemini 3.1}, \texttt{GLM-5.1}, \texttt{Qwen3.7-Plus}, and
\texttt{Kimi-K2.5} as four independent judges. The four judges unanimously
agree on 92.20\% of the labels, with Fleiss' $\kappa=0.848$ and pairwise
agreement ranging from 93.97\% to 98.58\%. These results support the robustness
of answer evaluation across the four evaluated judges.

\subsection{Main Experiment Case Studies}
\label{sec:appendix-main-experiment-case-studies}

Figures~\ref{fig:appendix-main-experiment-case-studies-a}
and~\ref{fig:appendix-main-experiment-case-studies-b} break representative
main-experiment cases into the underlying facts, relation type, baseline
outputs, and judge decisions, without reproducing the full conversation
sessions.

\begin{figure*}[!p]
\begin{subtlememorycasebox}{Case 1: Complementary Any-one -- coffee profile}
\scriptsize
\begin{itemize}[leftmargin=*, itemsep=1pt, topsep=2pt]
\item \textbf{Facts and target.} Lars drinks black coffee regularly, and when
he has coffee he usually takes it without milk or sugar. The target asks for
three compact staff-profile fields: coffee selection, milk/cream, and
sweetener.
\item \textbf{Relation type.} Complementary / Any-one: either fact is sufficient
to identify the same plain-coffee preference.
\item \textbf{Accepted answer.} Black brewed coffee; no milk or cream; no sugar.
\item \textbf{Baseline outputs.} Mem0 (GPT-5.4) answers ``one large black mug of
coffee,'' with no milk/cream and no sugar, and is judged correct. Oracle
(GPT-5.4) answers straight filter coffee with no milk/cream and no sugar, also
correct. MemOS (GPT-5.4) instead writes ``preferred coffee format for
alertness,'' ``milk-light,'' and no specific sweetener note, so it is judged
wrong.
\item \textbf{Judge decision.} The judge accepts semantically equivalent
plain-black-coffee answers and rejects outputs that miss or contradict the
no-milk/no-sugar fields.
\end{itemize}
\end{subtlememorycasebox}

\vspace{4pt}

\begin{subtlememorycasebox}{Case 2: Nuanced Contextual -- short-story writing}
\scriptsize
\begin{itemize}[leftmargin=*, itemsep=1pt, topsep=2pt]
\item \textbf{Facts and target.} Malik writes politically sharp op-eds for
independent magazines, but he deliberately keeps short fiction unrelated to
politics as a separate creative space. The target asks how to get a short story
moving this week.
\item \textbf{Relation type.} Nuanced / Contextual: the political-writing fact
is related but belongs to a different writing context, while the fiction fact
controls the current task.
\item \textbf{Accepted answer.} Keep the story away from politics; focus on
character, image, scene, and concrete detail rather than slogans or argument.
\item \textbf{Baseline outputs.} Oracle (GPT-5.4) explicitly says the plan
should be personal and concrete, with no debate scenes or slogan energy, and is
correct. MemOS (GPT-5.4) says to use a low-pressure, observation-based routine
that avoids slogans or policy talk, and is also correct.
\item \textbf{Judge decision.} The judge rewards selecting the fiction-specific
memory rather than applying Malik's op-ed persona to the story-writing task.
\end{itemize}
\end{subtlememorycasebox}
\caption{Representative main-experiment cases for complementary and nuanced
relations, showing the facts, relation type, selected baseline outputs, and judge
decisions.}
\label{fig:appendix-main-experiment-case-studies-a}
\end{figure*}

\begin{figure*}[!p]
\begin{subtlememorycasebox}{Case 3: Contradictory -- nature-documentary preference}
\scriptsize
\begin{itemize}[leftmargin=*, itemsep=1pt, topsep=2pt]
\item \textbf{Facts and target.} Julian is remembered both as liking nature
documentaries for unwinding and as finding nature documentaries boring and never
choosing them. The target asks for a hotel media-preference card.
\item \textbf{Relation type.} Contradictory: the target-relevant facts make
opposite claims under the same media-preference target.
\item \textbf{Accepted answer.} Leave the nature-documentary setting
unconfirmed, hold the card, and ask whether nature documentaries should be
enabled or excluded.
\item \textbf{Baseline outputs.} Oracle (GPT-5.4) states that the context points
both ways, sets the field to needs clarification, and is judged correct. MemOS
(GPT-5.4) resolves the conflict into ``leave off recommendation list'' and
``send now,'' so it is wrong. Mem0 (GPT-5.4) says there are no remembered media
preferences and repeats formatting instructions, so it is also wrong.
\item \textbf{Judge decision.} The judge marks a case correct only when the
answer preserves the conflict and asks for confirmation instead of choosing one
side or claiming no evidence exists.
\end{itemize}
\end{subtlememorycasebox}

\vspace{4pt}

\begin{subtlememorycasebox}{Case 4: Complementary Any-one -- book pickup list}
\scriptsize
\begin{itemize}[leftmargin=*, itemsep=1pt, topsep=2pt]
\item \textbf{Facts and target.} Amara enjoys African literature classics and
often gravitates toward classic novels by African authors. The target asks for
two books from an already-approved hold shelf.
\item \textbf{Relation type.} Complementary / Any-one: multiple facts point to
the same selection criterion, so any sufficient remembered variant can support
the answer.
\item \textbf{Accepted answer.} \textit{So Long a Letter} and \textit{The
Concubine}, because they are African literary classics and fit the shelf.
\item \textbf{Baseline outputs.} MemOS (GPT-5.4) selects \textit{So Long a
Letter} and \textit{The Concubine}, and is judged correct. Oracle (GPT-5.4)
selects the same two titles, also correct. Mem0 (GPT-5.4) selects \textit{So
Long a Letter} but replaces \textit{The Concubine} with \textit{A Month in the
Country}, so it is wrong.
\item \textbf{Judge decision.} The judge requires both the right titles and the
right memory relation: the titles must be selected because they match the
African-literary-classics preference, not merely because they are short or easy
weekend reads.
\end{itemize}
\end{subtlememorycasebox}
\caption{Representative main-experiment cases for contradictory and
relation-critical complementary examples, showing the facts, relation type,
selected baseline outputs, and judge decisions.}
\label{fig:appendix-main-experiment-case-studies-b}
\end{figure*}

\subsection{Perfect-retrieval Detailed Results}
\label{sec:appendix-perfect-retrieval-detailed-results}

Table~\ref{tab:gpt54-perfect-retrieval-results} reports the GPT-5.4
perfect-retrieval setting with the same relation and subtype columns as the
main results table. In this setting, answer generation receives the stored
memory units linked to the target evidence sessions, bypassing each system's
default query-time retrieval path.

\begin{table*}[!tbp]
\centering
\footnotesize
\setlength{\tabcolsep}{4.1pt}
\renewcommand{\arraystretch}{1.14}
\resizebox{0.98\textwidth}{!}{%
\begin{tabular}{@{}lcccccccc@{}}
\toprule
\multirow{2}{*}{\textbf{Method}} & \multicolumn{3}{c}{Complementary} & \multicolumn{3}{c}{Nuanced} & \multirow{2}{*}{\textbf{Contradictory}} & \multirow{2}{*}{\textbf{All}} \\
\cmidrule(lr){2-4}\cmidrule(lr){5-7}
& \textbf{Multi-evidence} & \textbf{Any-one} & \textbf{Overall} & \textbf{Temporal} & \textbf{Contextual} & \textbf{Overall} \\
\midrule
\rowcolor{BackboneBlue}
\multicolumn{9}{c}{\textbf{Base Model: GPT-5.4; Setting: Perfect Retrieval}} \\
\midrule
\quad Mem0 & 75.4\% & 82.7\% & 78.4\% & 88.9\% & 91.5\% & 90.2\% & \underline{51.7\%} & 77.9\% \\
\quad MemOS & 64.0\% & 81.3\% & 71.2\% & 79.3\% & 85.6\% & 82.4\% & 31.6\% & 67.1\% \\
\quad EverMemOS & 76.8\% & 80.7\% & 78.4\% & 89.1\% & 92.8\% & 90.9\% & 44.3\% & 76.4\% \\
\quad MIRIX & 66.8\% & 74.7\% & 70.1\% & 68.7\% & 84.5\% & 76.5\% & 15.9\% & 60.0\% \\
\quad A-Mem & 77.7\% & \underline{86.0\%} & 81.2\% & \underline{92.9\%} & \underline{93.0\%} & \underline{93.0\%} & \textbf{61.5\%} & \textbf{82.4\%} \\
\quad MemoBase & 34.6\% & 53.3\% & 42.4\% & 37.4\% & 47.9\% & 42.6\% & 18.8\% & 36.7\% \\
\noalign{\vskip 0.18em}
\cdashline{1-9}
\noalign{\vskip 0.12em}
\quad OpenClaw & 79.1\% & \underline{86.0\%} & \underline{82.0\%} & \textbf{95.2\%} & \textbf{95.4\%} & \textbf{95.3\%} & 51.2\% & \underline{81.2\%} \\
\quad MetaClaw & 18.5\% & 70.7\% & 40.2\% & 47.2\% & 52.3\% & 49.7\% & 5.3\% & 36.5\% \\
\noalign{\vskip 0.18em}
\cdashline{1-9}
\noalign{\vskip 0.12em}
\quad Mem0 + OpenClaw & \underline{79.6\%} & \textbf{88.0\%} & \textbf{83.1\%} & 90.2\% & 92.3\% & 91.2\% & 19.6\% & 71.6\% \\
\quad MemOS + OpenClaw & 69.2\% & 85.3\% & 75.9\% & 76.5\% & 86.6\% & 81.5\% & 16.4\% & 64.1\% \\
\quad EverMemOS + OpenClaw & \textbf{80.1\%} & 83.3\% & 81.4\% & 89.4\% & 91.0\% & 90.2\% & 45.1\% & 76.9\% \\
\bottomrule
\end{tabular}
}
\caption{GPT-5.4 perfect-retrieval results on SubtleMemory. Results are
reported with the same relation and subtype columns as the main results table.
Best and second-best values among non-oracle baselines are bolded and
underlined, respectively.}
\label{tab:gpt54-perfect-retrieval-results}
\end{table*}

\subsection{Representative Answer Examples}
\label{sec:appendix-answer-examples}

Figures~\ref{fig:appendix-correct-answer-examples}
and~\ref{fig:appendix-incorrect-answer-examples} give representative correct
and incorrect answer examples from baseline evaluation results. Each figure
includes both a user-related example and a non-user example, and reports the
original case, facts, reference answer, generated answer, and judge decision.

\begin{figure*}[!tbp]
\begin{subtlememorycasebox}{Representative correct-answer examples}
\scriptsize
\textbf{Example 1: User-related resource arrangement.}
\SubtleParaGap
\emph{System setting.} MemOS with GPT-5.4.
\SubtleParaGap
\emph{Original case.} Amara uses different design styles depending on whether
she is shaping her home environment or creating materials and spaces for kids'
science activities.
\SubtleParaGap
\emph{Original facts.} (1) At home, Amara prefers minimalist,
Scandinavian-inspired design with clean lines, light wood tones, and an
uncluttered feel. (2) For children's STEM workshops or toy-car race setups, she
uses bold colors, interactive visual cues, and playful layouts.
\SubtleParaGap
\emph{Golden answer.} Signal Sprint Kit --- it matches bold colors, interactive
cues, and small activity islands for an energetic kids setup.
\SubtleParaGap
\emph{Observed generated answer.} Signal Sprint Kit --- it best matches the
remembered brief: bright lane colors, big visual cues, and separate activity
islands support intuitive child navigation and fast engagement without relying
on dense text.
\SubtleParaGap
\emph{Judge label/reason.} CORRECT: the generated answer correctly selects the
Signal Sprint Kit and provides a reason that aligns with the accepted correct
answers regarding its suitability for a children's activity.

\SubtleParaGap
\textbf{Example 2: Non-user multi-evidence question.}
\SubtleParaGap
\emph{System setting.} Mem0 with GPT-5.4.
\SubtleParaGap
\emph{Original case.} What are the names of 6 Metropolitan cities in Korea and
their respective symbol flowers?
\SubtleParaGap
\emph{Original facts.} (1) Busan's symbol flower is Camellia. (2) Incheon's
symbol flower is Rose. (3) Ulsan's symbol flower is Pear flower.
\SubtleParaGap
\emph{Golden answer.} Busan's symbol flower is Camellia, Incheon's is Rose, and
Ulsan's is Pear flower.
\SubtleParaGap
\emph{Observed generated answer.} Busan --- Camellia; Incheon --- rose; Ulsan
--- Pear flower.
\SubtleParaGap
\emph{Judge label/reason.} CORRECT: the generated answer correctly identifies
the symbol flowers for Busan, Incheon, and Ulsan, matching the accepted correct
answers.
\end{subtlememorycasebox}
\caption{Representative correct-answer examples from baseline SubtleMemory
evaluation results. The examples cover both user-related and non-user sources.}
\label{fig:appendix-correct-answer-examples}
\end{figure*}

\begin{figure*}[!tbp]
\begin{subtlememorycasebox}{Representative incorrect-answer examples}
\scriptsize
\textbf{Example 1: User-related language archive card.}
\SubtleParaGap
\emph{System setting.} MemOS with GPT-5.4.
\SubtleParaGap
\emph{Original case.} Amara enjoys learning languages casually, but the
language she has focused on in her spare time has shifted over the years.
\SubtleParaGap
\emph{Original facts.} (1) When she first moved to the U.S. for graduate
school, Amara casually practiced French to keep using the basics she already
knew. (2) More recently, Amara has been putting that casual language-learning
energy into strengthening her conversational Igbo to feel more connected to her
family and heritage.
\SubtleParaGap
\emph{Golden answer.} Time period: first move to the U.S. for grad school;
casual language practice: French basics; why I kept it up: to keep using what I
already knew; intensity level: casual.
\SubtleParaGap
\emph{Observed generated answer.} Time period: first move to the U.S. for
graduate school; casual language practice: random podcasts, science-themed
narratives; why I kept it up: made the apartment feel less quiet, more
comfortable; intensity level: light, casual. \ldots
\SubtleParaGap
\emph{Judge label/reason.} WRONG: the generated answer includes incorrect
information for the language practiced and the reason for keeping it up, failing
to mention French basics and maintaining prior knowledge.

\SubtleParaGap
\textbf{Example 2: Non-user contradictory question.}
\SubtleParaGap
\emph{System setting.} Mem0 with GPT-5.4.
\SubtleParaGap
\emph{Original case.} What type of book is \textit{The Fault in Our Stars}?
\SubtleParaGap
\emph{Original facts.} (1) What type of book is \textit{The Fault in Our Stars}
when describing intended age group? Young adult. (2) What type of book is
\textit{The Fault in Our Stars} when describing general content? Realistic
fiction.
\SubtleParaGap
\emph{Golden answer.} The remembered sessions conflict: one says \textit{The
Fault in Our Stars} is young adult, while another calls it realistic fiction, so
the type remains unresolved without clarification.
\SubtleParaGap
\emph{Observed generated answer.} The context supports calling \textit{The
Fault in Our Stars} a young adult novel; on the user's classroom display card,
its genre line is listed as realistic fiction.
\SubtleParaGap
\emph{Judge label/reason.} WRONG: the generated answer invents details about a
classroom display card to resolve the conflict, rather than acknowledging the
unresolved contradiction as required by the accepted answers.
\end{subtlememorycasebox}
\caption{Representative incorrect-answer examples from SubtleMemory evaluation
results. The examples cover both user-related and non-user sources.}
\label{fig:appendix-incorrect-answer-examples}
\end{figure*}

\section{Use of AI Assistants}
\label{sec:appendix-ai-assistant-use}

AI assistance was used in three limited roles. First, LLMs supported benchmark
construction by generating and filtering semantic variants, interaction sessions,
and evaluation instances; the full construction procedure is documented in
Appendix~\ref{sec:appendix-data-construction}. Second, our evaluation protocol
uses an LLM-based automatic judge, with human-agreement validation reported in
Appendix~\ref{sec:appendix-llm-judge-validation}. Third, general-purpose AI
writing tools were used to improve wording and readability. They were not used
to originate the research questions, choose the experimental design, produce the
reported results, or draw the paper's conclusions. The authors reviewed and
approved all benchmark design choices, analyses, results, and manuscript text.

\end{document}